\documentclass{article}
\usepackage{iclr2027_conference,times}
\iclrfinalcopy  

\usepackage{amsmath}
\usepackage{amssymb}
\usepackage{amsthm}
\usepackage{graphicx}
\usepackage{booktabs}
\usepackage{multirow}
\usepackage{longtable}   % only for the 72-row appendix leaderboard
\usepackage{subcaption}
\usepackage{xcolor}
\usepackage{microtype}

\usepackage{wrapfig, blindtext}
\usepackage{comment}
\usepackage{pdflscape}

\usepackage{setspace}
\usepackage{etoolbox}
\usepackage{longtable}
\usepackage{times}
\usepackage{epsfig}
\usepackage{graphicx}
\usepackage{amsmath}
\usepackage{amssymb}

\usepackage{enumerate}
\usepackage{comment}
\usepackage{amsfonts}
\usepackage{gensymb}
\usepackage{float}

\usepackage{makecell}
\usepackage{array}
\usepackage{multirow}
\usepackage{footmisc}
\usepackage{breakurl}
\usepackage{mathtools}
\usepackage{todonotes}
\usepackage{hhline}
\usepackage{footnote}
\usepackage{cases}

\usepackage{listings}
\usepackage{booktabs}

\usepackage{tabularx}
\usepackage{lipsum}
\usepackage{capt-of}
\usepackage{mathrsfs}
\usepackage{epsfig}	
\usepackage{array}
\usepackage[utf8]{inputenc}
\usepackage[T1]{fontenc}

\usepackage{float}
\usepackage{tikz}
\usetikzlibrary{positioning, arrows.meta, fit, backgrounds, calc}

\usepackage{hyperref}
\usepackage{url}
\hypersetup{
	colorlinks = true,
	linkcolor  = blue,
	citecolor  = blue,
	urlcolor   = blue!60!black,
	pdfauthor  = {},
	pdfsubject = {},
	pdfkeywords= {}
}

\usepackage[capitalise,noabbrev]{cleveref}

\graphicspath{{figures/}}

\newcommand{\rdk}[1]{\textsuperscript{\textbf{[#1]}}}

\newcommand{\seanet}{SEA-Net}
\newcommand{\millet}{MILLET}
\newcommand{\webtraffic}{WebTraffic}
\newcommand{\ucr}{UCR}
\newcommand{\ndcg}{NDCG@$n$}
\newcommand{\anonrepo}{\url{https://anonymous.4open.science/r/seanet-ANON}}

\newcommand{\R}{\mathbb{R}}
\newcommand{\bag}{\mathbf{X}}              % the input series (the bag)
\newcommand{\emb}{\mathbf{z}}              % embedding of one time step
\newcommand{\bagpred}{\hat{y}}             % bag logits (the prediction)
\newcommand{\interp}{\mathbf{I}}           % the interpretation
\newcommand{\nsteps}{t}                    % number of time steps
\newcommand{\dmodel}{d}                    % encoder width
\newcommand{\sig}{\sigma}
\newcommand{\softmaxt}{\operatorname{softmax}_j}

\DeclareMathOperator*{\argmax}{arg\,max}

\newtheorem{property}{Property}

\usepackage{enumitem}
\setlist[itemize]{topsep=2pt,itemsep=1pt,parsep=0pt,leftmargin=1.4em}
\setlist[enumerate]{topsep=2pt,itemsep=1pt,parsep=0pt,leftmargin=1.6em}

\usepackage{placeins}
\makeatletter
\AtBeginDocument{%
  \let\sea@section\section
  \def\section{\FloatBarrier\sea@section}%
  \let\sea@subsection\subsection
  \def\subsection{\FloatBarrier\sea@subsection}%
}

\newenvironment{sea@fixedtable}[1][]%
  {\par\addvspace{\intextsep}\def\@captype{table}}%
  {\par\addvspace{\intextsep}}
\newcommand{\stacktables}{%
  \let\table\sea@fixedtable
  \let\endtable\endsea@fixedtable}
\makeatother

\newcommand{\papertitle}{SEA-Net: A Small and Inherently Interpretable Neural
	Network for Time Series Classification}

\title{\papertitle}

\author{Ubaid Ur Rehman \& Tanmoy Mondal\thanks{Corresponding author.} \\
	FOX Team, Centre de Recherche en Informatique, Signal et Automatique de Lille (CRIStAL) \\
	University of Lille, CNRS, Centrale Lille, UMR 9189  \\
	F-59000 Lille, France \\
	\texttt{ubaid-ur.rehman@etu.univ-cotedazur.fr}, \texttt{tanmoy.mondal@univ-lille.fr}
}

\begin{document}

\maketitle
\lhead{Preprint}  

\begin{abstract}
% !TEX root = ../main.tex
% =====================================================================
% ABSTRACT -- four questions, in order: problem / limitation / method /
% result. No implementation detail, no adjectives, no claim that is not
% in Section 5.
% =====================================================================

Traditional time-series classification methods often operate as black boxes, providing a class prediction without indicating which time point or time interval within the series led to that prediction. To improve temporal interpretability, MILLET (Multiple Instance Learning with Temporal Explanation), which combines an InceptionTime encoder with Conjunctive Pooling, provides both classification and temporal explanations. However, MILLET contains approximately $424$\,K parameters, making it unsuitable for deployment on resource-constrained micro-controllers, and it achieves an overall time-series accuracy of $0.888$ and an explanation score of $0.681$ NDCG@$n$. We therefore propose \seanet{}, a lightweight architecture using small multi-scale depthwise-separable convolutional TSC encoders and improved Multiple Instance Learning (MIL) pooling heads. Our \seanet{} bottleneck encoder with Top-$k$ pooling achieves $0.931$ accuracy and $0.765$ NDCG@$n$ with only $41$\,K parameters. Another variant, using a gated encoder with Top-$k$ pooling, achieves $0.955$ accuracy and $0.750$ NDCG@$n$ with $62$\,K parameters. These results demonstrate that lightweight architectures can maintain strong classification performance and temporal explanations while substantially reducing model size. The proposed variants address specific limitations of the pooling strategy used by \millet{}, providing a practical direction for interpretable time-series classification on resource-constrained edge devices.

\end{abstract}

% !TEX root = ../main.tex
% =====================================================================
% 1. INTRODUCTION -- page budget 1.20 (including the teaser figure)
%
% Six paragraphs, in the order a reader needs them:
%   P1 problem      : classification + interpretation
%   P2 deployment   : why size and FLOPs matter on small devices
%   P3 baseline     : what MILLET gives us
%   P4 limitation   : why MILLET is expensive, and where the cost sits
%   P5 our approach : bottleneck encoder + Top-k pooling
%   P6 result       : the trade-off, then the contribution list
% =====================================================================

\section{Introduction}
\label{sec:intro}

% ---------------------------------------------------------------------
% Teaser figure: same input, three models, prediction + explanation.
% ---------------------------------------------------------------------
\begin{figure}[!tbp]
	\centering
	\includegraphics[trim = 0.0cm 0.0cm 0.0cm 9.2cm, clip, width=0.89\linewidth]{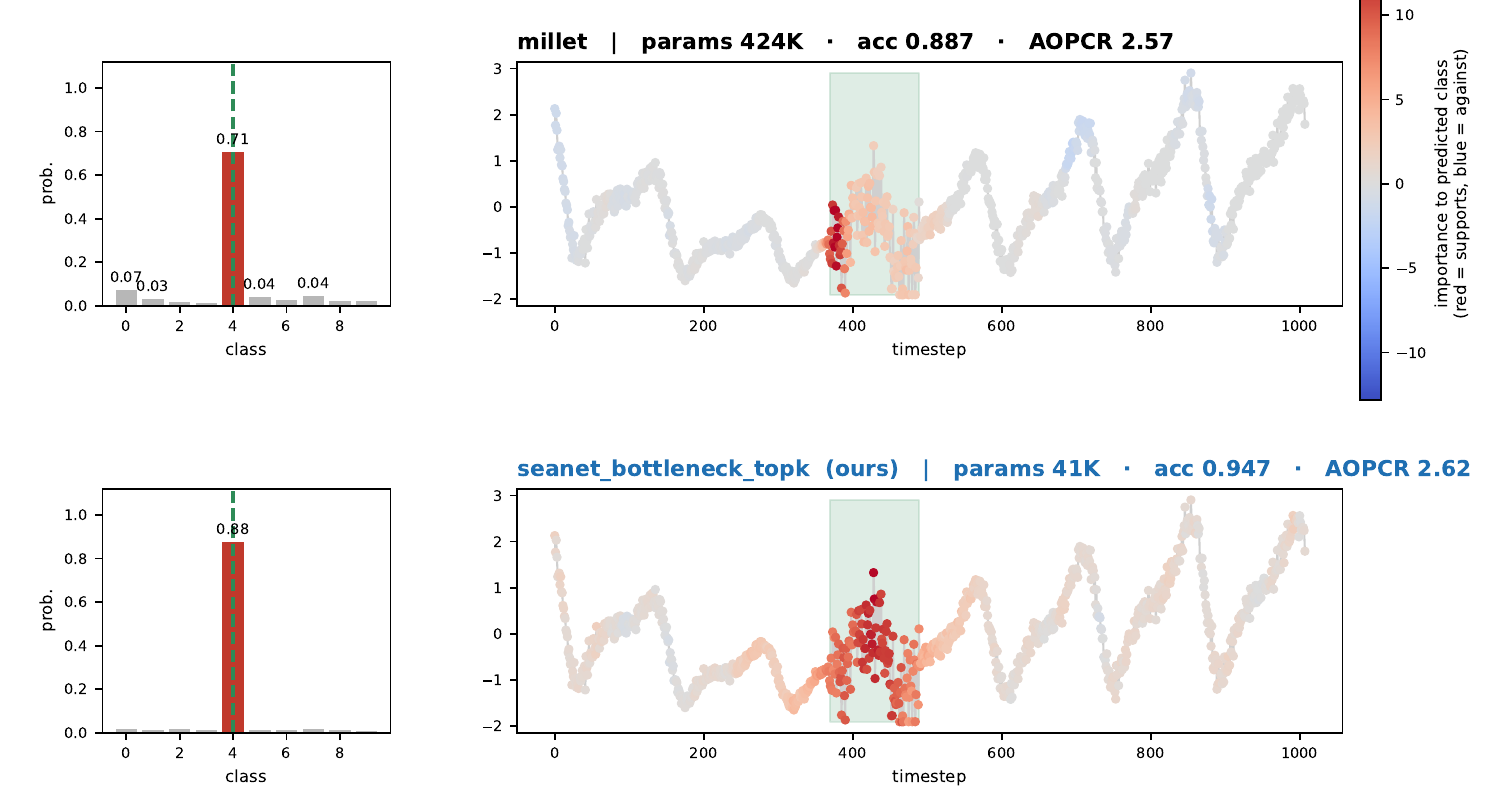}
	\caption{The same \webtraffic{} series classified by three models. Each panel shows the predicted class and, its right side provides the explanation or interpretation of the prediction, given by the model. The red marks provide the evidence, supporting the predicted class at each time step. The ``green shaded window'' is the original ground truth of the time steps and was never used during training.
%	 A standard black-box convolutional model (top) predicts class 4 with 53\,\% confidence and offers no explanation. 
	 \millet{} (top) predicts the same class with 71\,\% confidence and does explain itself, but needs 424\,K parameters. \seanet{} (bottom) is more confident, produces a sharper explanation, and uses 41\,K parameters.}
	\label{fig:teaser}
\end{figure}

%Time Series (TS) represent a large domain of data that cross several disciplines and one of the major applications of time series analysis is \emph{Time Series Classification (TSC)}. A uni-dimensional discrete time series of length $T$ is an ordered list of measurements $X = (x_1, x_2, ..., x_T)$ and TSC assigns to that complete sequence of data points a single label (that means to classify) $y \in \{1, \dots, C\}$. For example, classify a time series as normal or abnormal for a heartbeat recording, healthy or faulty for a machine sensor, one traffic pattern or other for a web server log.
\emph{Time Series Classification (TSC)} assigns a label $y \in \{1, \dots, C\}$ to a univariate time series $X = (x_1, x_2, \dots, x_t)$, for example distinguishing a normal from an abnormal heartbeat recording, or a healthy from a faulty machine.
 Deep learning (DL) models solve this task well, and convolutional architectures in particular are its standard baselines. Accuracy alone is not enough in the applications that motivate this work, which impose three requirements \emph{at the same time}. The first is \textit{accurate classification}: the model must remain competitive with the strong convolutional baselines used in the TSC literature. 
% The second is \textbf{interpretable predictions}: the model must indicate \emph{which time steps} caused its decision, and that indication must quantify the decision was actually computed from, not a reconstruction produced afterwards by a separate tool. 
 The second is \textit{interpretable predictions}: the model must show \emph{which time steps} drove its decision, and this explanation must be part of the computation that produces the prediction itself, not a post-hoc approximation produced afterward by a separate tool. The third is \textit{a small parameter budget}: the model must be small enough to be a realistic candidate for deployment on an edge device or a micro-controller unit (MCU), where memory is measured in hundreds of kilobytes.

We build on \millet{} \citep{early2024millet, dietterich1997mil}, which performs TSC as a \emph{Multiple Instance Learning} problem. In MIL a label is attached to a bag \citep{ilse2018attention} of \emph{instances}, not to each instance. The whole time series and each time step are treated as the bag and the instances respectively, and only the bag-level label $y$ is observed.
%MIL \citep{dietterich1997mil} is a form of weakly supervised learning in which training labels are attached to \emph{groups} of examples rather than to individual ones. A group is a \emph{bag} \citep{dietterich1997mil, ilse2018attention} and its members are \emph{instances}; the bag carries a label, the instances do not. 
%{\color{violet}\rdt{I}  Formally, a bag
%$\bag = (\inst_1, \dots, \inst_{\nsteps})$ has an observed label $y$ while the instance labels $y_1, \dots, y_{\nsteps}$ exist conceptually but are never observed. 
%The problem is to predict $y$ from the bag and, ideally, to recover which instances were responsible.}
This is exactly the structure of time series classification: the series as a whole is labeled, but only a few time steps carry the evidence; the abnormal beat in a long recording, the spike in a traffic trace. Treating the series as the bag and each time step as an instance turns interpretability into a MIL problem, recovering instance-level contributions from bag-level labels and so yielding interpretability alongside classification.
\millet{} provides both the time-series classification and a faithful built-in explanation.
%{\color{violet}\rdt{I}  It treats a time series as a \emph{bag} of time steps and applies multiple instance learning \citep{dietterich1997mil, ilse2018attention}.}
%The encoder block in the model produces one embedding per time step whereas an attention branch decides where to look within the time series. Whereas, a per-step classifier decides what each position argues for, and the product of the two (i.e. attention and prediction) is both summed into the final prediction and considered as the interpretation. 
%{\color{violet}\rdt{II}  The encoder produces one embedding per time step; an attention branch decides where to look in the time series, a per-step classifier decides what each position argues for, and the product of two (i.e. attention and prediction) is both summed into the prediction and considered as the explanation.}
%The \millet{} is accurate and either matching or beating standard deep time series classifiers on the \ucr{} archive \citep{dau2019ucr}.

But its \emph{InceptionTime} \citep{fawaz2020inceptiontime} backbone needs about $423{,}707$ parameters and 848\,MFLOPs for one series of length 1008, so at \mbox{32-bit} precision the weights alone take about 1.6\,MB, which already exceeds the memory budget above, before activations, runtime or application code.
Swapping \millet{}'s interpretable head for ordinary global average pooling on the same backbone removes only 1{,}041 of those parameters (\Cref{sec:method:mil}): the ability to explain itself costs a quarter of a percent of the model, and almost all of the cost is the encoder. This paper builds on that observation.
Edge-oriented models fit micro-controllers \citep{banbury2021mlperftiny, jacob2018quantization} but predict only a class label; to our knowledge no model meets all three requirements.
\begin{comment}
\begin{figure}[!tbp]
	\centering
	\includegraphics[width=0.90\linewidth]{figures/fig02_mil_view.png}
	\caption{The MIL view of a time series, read left to right. 
%		{\color{violet}\rdt{I}  The series is the \emph{bag} and each
%		time step $\inst_j$ is one \emph{instance}.} 
		The same encoder is applied at every time step and
		produces one instance embedding per step; the MIL pooling head weights those embeddings and
		combines them into a class prediction (right). 
%		{\color{violet}\rdt{II}  Keeping the same weighted contributions instead of
%		discarding them yields the interpretation below:} 
		one importance value per time step, drawn on the
		original series (red = pushes towards the predicted class, blue = pushes away). Both gray boxes
		are interchangeable, which is what allows any encoder of \Cref{sec:method:enc} to be paired
		with any pooling head of \Cref{sec:methods:pooling}.}
	\label{fig:mil_view}
\end{figure}
\end{comment}
Hence, we propose our \emph{\textbf{S}elective \textbf{E}vidence \textbf{A}ggregation \textbf{Net}work (\seanet{})}, an inherently interpretable architecture for time series
classification. 
%{\color{violet}\rdt{III}  which combines multi-scale temporal feature extraction with selective, class-specific aggregation of the evidence (i.e. interpretability), found at each time step.} 
%{\color{violet}\rdt{IV}  We therefore keep the MIL setup and change the two parts that decide its cost and how selective it is. First, the encoder: a stack of multi-scale depth wise separable dilated convolutions, which keeps the length unchanged so that one embedding per time step reaches the head, and in which the point-wise $1\times1$ mixing that holds most of the parameters once the convolutions are separable is split into two smaller matrices through a narrow bottleneck of width $r = d/4$, halving that term.
%Second, the pooling head: instead of averaging the evidence over all $T$ time steps behind one shared attention gate, we score each class separately and average only the $k = \lceil \kappa T \rceil$ largest evidence values for that class.}
Both parts already exist --- separable convolutions \citep{chollet2017xception, howard2017mobilenets}, squeeze-and-expand bottlenecks \citep{he2016resnet, iandola2016squeezenet, sandler2018mobilenetv2} and $k$-max selection in MIL \citep{durand2016weldon}, and we claim none as new; what is new is their combination inside an interpretable MIL classifier.%
%In that manner, the resulting model uses 41{,}324 parameters and
%76.7\,MFLOPs per time series, $10.3\times$ and $11.1\times$ below \millet{}. 
%On \webtraffic{}, the one dataset in which the ground truth information has time steps that actually carry the class label, our technique beats the
% \millet{}'s baseline at both jobs at once. We obtained the accuracy of $0.947$ against $0.887$, \ndcg{} $0.772$
%against $0.677$ (\Cref{fig:teaser}) on  \webtraffic{} dataset. 
Our model has 41{,}324 parameters and 76.7\,MFLOPs, $10.3\times$ and $11.1\times$ below \millet{}. On \webtraffic{}, the one dataset whose ground truth marks the time steps carrying the class label, it beats \millet{} on both jobs: accuracy $0.931$ against $0.888$ and \ndcg{} $0.765$ against $0.681$ (\Cref{fig:teaser}).
%Whereas, in the case of 128 datasets of UCR repository; our technique slightly lags behind on the 85 \ucr{} datasets with an accuracy of $0.810$ against $0.827$ by our evaluation of \millet{} \footnote{\millet{}'s published result is $0.843$, but when we evaluated by using their saved model and their provided code, we obtained $0.827$ accuracy. We contacted the authors to know this difference from their published result but we didn't get any reply}.
Across the $85$ \ucr{} datasets, our method stays within two accuracy points of \millet{} (\Cref{sec:res:ucr}).\footnote{We reach $0.810$ on the $85$ \ucr{} datasets against $0.827$ for our re-run of \millet{} using the authors' released model and code, whose published figure is $0.845$; the authors did not reply to our query about the gap.} The following are our key contributions:
%{\color{violet}\rdt{VI}  We show where that loss comes from
%instead of hiding it in an average: a wide version of the same encoder, at 269\,K parameters, reaches
%$0.829$ and ranks above the equally-trained baseline, so the price is paid for the smaller size and
%not for a flaw in the design.}
%\paragraph{Contributions:}
%\begin{itemize}
   \textbf{i) A MIL encoder small enough to be a candidate for small devices}
        (\Cref{sec:method:enc}): multi-scale depthwise separable dilated convolutions with a
        bottlenecked pointwise mix, keeping the length unchanged so that the per-time-step
        interpretation is not lost.
%        {\color{violet}\rdt{VII}  at $10.3\times$ fewer parameters and $11.1\times$ fewer FLOPs
%        than \millet{}}.
   \textbf{ii) Per-class Top-$k$ conjunctive pooling} (\Cref{sec:method:head}), which adds no
        parameters and becomes \emph{exactly} the baseline head at $\kappa = 1$
        (\Cref{prop:topk}). This property is what makes the $\kappa$ study of \Cref{sec:res:ablation}
        a controlled experiment rather than a comparison of two different models.
%  \item \textbf{An ablation} separating the two changes (\Cref{sec:res:ablation}): the bottleneck
%        buys the size reduction with no accuracy cost we can measure, and \mbox{Top-$k$} buys the
%        interpretation quality, measured against per-time-step ground truth.
   \textbf{iii) A warning about the AOPCR measure} (\Cref{sec:res:aopcr}): AOPCR has no fixed scale, and with the architecture and the code held fixed it changes by $4.9\times$ when only the training budget changes, so we compare only against baselines we retrained ourselves.

\section{Related Work}
\label{sec:related}

\textbf{Deep learning for time series classification: }
%Convolutional models have been the strongest in this research area and 
In \citet{wang2017fcn}, the ``Fully Convolutional Network (FCN)'' and the ``Residual Network (ResNet)'' are established as strong baselines for TSC. The InceptionTime \citep{fawaz2020inceptiontime} architecture extends this baseline by applying filters of several kernel sizes in parallel, capturing patterns of different durations.
%{\color{violet}\rdt{X}  The ``Temporal Convolutional Networks'' \citep{bai2018tcn} has been highly useful for time series analysis and also been used within our proposed model, \emph{dilated} convolutions, which widen the region a filter sees without adding weights.}
Temporal Convolutional Networks (TCNs) \citep{bai2018tcn} have demonstrated their effectiveness for time-series analysis, and we take inspiration from them for our proposed model. TCNs use \emph{dilated convolutions}, which enable the network to capture information over a wider temporal context. This enlarges the receptive field and allows the model to capture both short-term and long-term temporal dependencies without significantly increasing the number of model parameters.
%The property of dilated convolutions to effectively enlarge the receptive field that allows the model to capture both short-term and long-term temporal dependencies while maintaining computational efficiency, is used in our proposed model.
The comprehensive review by \citet{fawaz2019review} shows that convolutional architectures are an effective family of approaches for TSC. Despite their architectural differences, these models typically conclude with a \emph{Global Average Pooling (GAP)} layer, which removes the explicit temporal resolution of the learned representations.
% {\color{violet}\rdt{XI}  This operation aggregates the feature representations across all time steps, transforming the temporal feature map into a single fixed-dimensional vector. Although this aggregation is beneficial for classification, it removes the explicit temporal resolution of the learned representations. Consequently, it becomes difficult to determine which specific time steps or temporal regions contributed to a particular prediction, limiting the model's ability to provide explanations at the temporal level.}

\textbf{Interpretability and multiple instance learning:}
Interpretability methods can be broadly divided into post-hoc approaches and models that incorporate explanations directly into the prediction process. Post-hoc attribution methods, such as LIME \citep{ribeiro2016lime}, SHAP \citep{lundberg2017shap}, and Integrated Gradients \citep{sundararajan2017ig}, explain a trained model after prediction, whereas \citet{rudin2019stop} advocates incorporating interpretability directly into the model architecture.
For time series, two main directions follow this principle. The first direction uses \emph{shapelets}, i.e., short representative subsequences whose presence can explain a prediction. VQShape \citep{wen2024vqshape} learns a codebook of representative shapes, while InterpGN \citep{wen2025interpgn} combines a shapelet expert with a deep expert through a confidence gate. The second direction, which we follow, is based on MIL \citep{dietterich1997mil}. Attention-based MIL \citep{ilse2018attention} introduced learnable instance-level pooling, later extended by models such as DSMIL \citep{li2021dsmil} for histopathology. \millet{} \citep{early2024millet} adapted this idea to TSC.
%by treating each time step as an instance and using \emph{conjunctive pooling} to score and aggregate them. These instance scores directly contribute to the prediction and therefore also provide a natural temporal explanation. 
%Finally, selecting only the most informative instances rather than averaging all of them is also related to earlier approaches such as WELDON \citep{durand2016weldon}, which introduced $k$-max region selection for weakly supervised recognition.

\textbf{Small architectures and deployment budgets:}
Depthwise separable convolutions \citep{chollet2017xception, howard2017mobilenets} and squeeze-and-expand bottlenecks \citep{he2016resnet, iandola2016squeezenet, sandler2018mobilenetv2} are standard techniques for reducing the parameter count and computational cost. For time series, ATCN \citep{baharani2022atcn} combines residual connections with separable convolutions for efficient inference, while the method in \citet{risso2023tcn} optimizes the dilation and channel structure of TCNs under edge-device constraints. However, these approaches focus on efficiency rather than interpretability: they produce a prediction without explicitly indicating which temporal regions contributed to it.

For deployment on resource-constrained devices, MCUNet \citep{lin2020mcunet}, MicroNets \citep{banbury2021micronets}, and MLPerf Tiny \citep{banbury2021mlperftiny} provide benchmarks and design principles based on constraints such as flash memory, SRAM, and computational cost. 
%Post-training integer quantization \citep{jacob2018quantization} can further reduce memory and computational requirements by representing weights and activations with lower-precision integers. 
These works demonstrate how neural networks can be made suitable for micro-controller-class hardware, but they do not address the additional challenge of maintaining interpretability under such resource constraints.

\section{Method}
\label{sec:method}
%The proposed \emph{\textbf{S}elective \textbf{E}vidence \textbf{A}ggregation \textbf{Net}work (\seanet{})} model is a MIL based technique for TSC.
%{\color{violet}\rdt{XIII}  Like \millet, our model is also based on MIL a natural framework that provides inherent interpretability without any requirement of additional labeling beyond what is already provided by existing TSC datasets.} 
%In the following section, we describe the \seanet~framework in detail. 

\subsection{Problem Setup and Notation} 
\label{sec:character_seanet} 
%The proposed \seanet{} model combines multi-scale temporal feature extraction with selective, class-specific aggregation of the evidence found at each time step.
%\subsubsection{The Characteristics of \seanet{}}  
We propose the \emph{\textbf{S}elective \textbf{E}vidence \textbf{A}ggregation \textbf{Net}work} (\seanet{}), a MIL model for TSC that keeps \millet{}'s two-part structure but replaces its \emph{InceptionTime} backbone with a far smaller encoder (\Cref{sec:method:enc}) and its \texttt{Conjunctive} head with three selective pooling variants (\Cref{sec:method:head}).
%The \seanet{} model consists of two main components: an \emph{encoder \& instance level embedding} and a \emph{MIL pooling head}. We first introduce the notation used throughout both components.
%Before describing these two components in detail, we first introduce the general terminology and concepts required to understand their respective architectures and operations. 
\seanet{} has two components: an encoder $\psi_{enc-embed}$, which produces one embedding per time point, and a MIL pooling head $\psi_{pool}$, which turns those embeddings into a class prediction. A dataset holds $n$ univariate series $X_i=(x_i^1,\dots,x_i^t)$, $i \in \{1,\dots,n\}$, of common length $t$, each carrying a single label $Y_i\in\{1,\dots,C\}$, where $C$ is the number of classes. We treat each series as a MIL \emph{bag} and each time point as an \emph{instance}, so the encoder maps $X_i$ to $Z_i=[z_i^1,\dots,z_i^t]\in\mathbb{R}^{t\times d}$, one $d$-dimensional embedding per instance. The instance labels $\{y_i^1,\dots,y_i^t\}$ are never annotated in TSC datasets and are never used for training; recovering them from the bag label alone is precisely what MIL techniques are capable of. Unlike classical MIL \citep{hashimoto2020msdamil,li2021dsmil}, our instances are ordered rather than i.i.d., and every layer of \seanet{} preserves that order.

Whether that per-instance evidence survives depends on $\psi_{pool}$. Classical \emph{embedded pooling} \citep{wang2018revisiting} applies global average pooling, $P_i=\frac{1}{t}\sum_{j=1}^{t}z_i^j$, then a classifier $\hat{Y}_i=\psi_{clf}(P_i)$ to classify: the time axis is collapsed \emph{before} classification, so no per-time-point prediction exists and 
the model cannot report which time points drove the prediction for the series. Our pooling heads (\Cref{sec:method:head}) preserve this order: they score each time point first and aggregate afterward, which makes the interpretation $\hat{I}_i\in\mathbb{R}^{t\times C}$ available.
We keep \millet{}'s two-part structure but replace \emph{InceptionTime} with a smaller, cheaper encoder, and its \texttt{Conjunctive} pooling head with three pooling variants of our own: \texttt{Class-wise Conjunctive}, \texttt{Soft-Max Conjunctive with Learnable Sharpness} and \texttt{Top-k Conjunctive} pooling.
%{\color{violet}\rdt{XIV}  Our pooling techniques are designed in a \emph{plug-and-play} mode, that means any TSC model's architecture which uses the \texttt{GAP}, \texttt{Conjunctive} or any other similar pooling method can be easily upgraded to one of these pooling techniques and will be able to meet aforementioned characteristics of \seanet{} (mentioned in Section~\ref{sec:character_seanet}).}

\subsection{SEA-Net Encoder}
\label{sec:method:enc}
\begin{figure}[!tbp]
	\centering
	\makebox[\linewidth][c]{\includegraphics[width=1.0\linewidth]{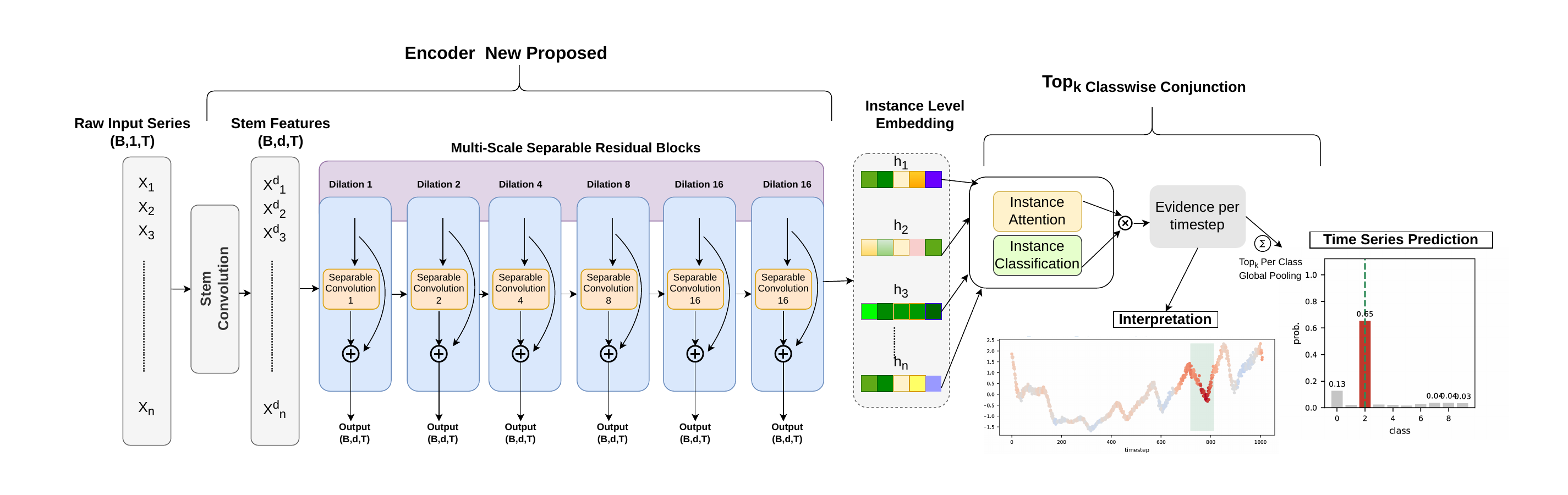}}
	\caption{Overview of a complete \seanet{} model, read left to right, with the tensor shape under each stage. \textbf{Left, the \seanet{} encoder:} the series $(B,1,\nsteps)$ passes through a stem convolution to $(B,\dmodel,\nsteps)$ and then through six multi-scale separable residual blocks, whose dilation grows $1,2,4,8$ and is then capped at $16$; the curved arrow inside each block is the residual connection, and $\nsteps$ is unchanged at every stage. \textbf{Middle:} one instance-level embedding $\emb_j$ per time step. \textbf{Right, the MIL pooling head (\Cref{sec:methods:pooling}):} 
		%	{\color{violet}\rdt{II}  an attention branch scores how much each time step counts, a classifier branch scores what it argues for, and their product is the evidence at that time step. Aggregating the evidence over time gives the prediction (bar chart); retaining the same per-time-step evidence gives the interpretation drawn on the series. The explanation and the prediction are the same quantities.}
	}
	\label{fig:arch_overall}
\end{figure}

\Cref{fig:arch_overall} gives an overview of the encoder and its interface with the MIL pooling head.
The encoder module begins with a \emph{stem convolution layer}, that extracts local temporal features from the input time series while simultaneously increasing its representation from a single input channel of a time point to $d$ feature channels. The stem convolution is followed by a stack of \emph{multi-scale separable residual blocks} (\Cref{sec:arch:block}). The output ($Z_i \in \mathbb{R}^{t \times d}$) of the encoder module is given as the input to the MIL pooling module, which in turn provides the prediction as the posterior probabilities $\hat{Y}_i \in \mathbb{R}^{1 \times C} $ of $C$ output classes and the interpretation score of each time step $\hat{I}_i \in \mathbb{R}^{t \times C} $. 

\paragraph{Multi-scale separable residual blocks:}
\label{sec:arch:block}
The encoder body is a stack of six identical blocks built from two established concepts. \emph{Dilated convolutions} \citep{bai2018tcn} insert gaps between the filter taps, widening the temporal window a filter sees without adding weights, and \emph{depthwise separable convolutions} \citep{chollet2017xception, howard2017mobilenets} split a convolution into a depthwise part
$D_k^{(\delta)}$, filtering each channel independently along time, and a pointwise $1\times1$ part $P$ mixing channels at each time point.
The smallest element is a \emph{unit}. Three depthwise convolutions read the same input in parallel at kernel sizes  $\mathcal{K}=\{5,11,23\}$, so short, medium and long patterns are examined at once and no single scale has to be chosen in advance; their outputs are summed and mixed by $P$:
\begin{equation}
	\mathcal{U}(\mathbf{h}) = \operatorname{Drop}\Big(\operatorname{ReLU}\Big(\operatorname{BN}\Big(
	P\Big(\textstyle\sum_{k \in \mathcal{K}} D_k^{(\delta)}(\mathbf{h})\Big)\Big)\Big)\Big),
	\qquad \mathcal{K} = \{5, 11, 23\},
	\label{eq:unit}
\end{equation}
where $\mathbf{h}\in\mathbb{R}^{t \times d}$ is the input and $\delta$ the dilation. A block is two units in sequence with a residual connection, so it learns a correction to its input rather than replacing one representation with another:
\begin{equation}
	\operatorname{Block}(\mathbf{h}) = \operatorname{ReLU}\big(\mathbf{h} +
	(\mathcal{U} \circ \mathcal{U})(\mathbf{h})\big).
	\label{eq:block}
\end{equation}

Six blocks are stacked, each feeding the next, with the dilation growing geometrically and then held: $\delta_i = \min(2^i, \delta_{\max})$, $\delta_{\max}=16$. This operation has two properties. First, the receptive field widens quickly but stays bounded: the five taps of the $k=5$ filter span five consecutive time points at $\delta=1$, whereas they span up to $65$ time points at $\delta=16$, so each embedding summarizes a local, finite region at several scales. Second, every convolution uses ``same'' padding with a stride of one, and no pooling or temporal downsampling is applied; the length $t$ is therefore unchanged at every
stage, which is what delivers one embedding per time point to the pooling head.

\begin{comment}
%\todo{it is not understandable}
\paragraph{\seanet{} Bottleneck point-wise mixing:}

In this variant, we introduce a $1\times1$ convolution inside the dilation block \Cref{fig:arch_block} to further reduce the number of parameters and create a stronger bottleneck.

The point-wise step dominates the parameter count, costing $\dmodel^2$ weights per unit (\Cref{sec:arch:block}).

\begin{equation}
  P = P_{\text{expand}} P_{\text{squeeze}}, \qquad
  P_{\text{squeeze}} \in \R^{r \times \dmodel}, \quad
  P_{\text{expand}} \in \R^{\dmodel \times r},
  \qquad \text{cost } \dmodel^2 \rightarrow 2\dmodel r.
  \label{eq:bottleneck}
\end{equation}

With $r = \dmodel/4$, the point-wise cost becomes $\dmodel^2/2$, exactly half, independently of $\dmodel$. At the setting used in the experiments ($\dmodel = 64$, 4 blocks), this reduces the complete model from 57{,}580 to 41{,}324 parameters -- a 28\,\% saving on the whole encoder, not only on the point-wise step. The block still consumes and returns the same $(B, \dmodel, \nsteps)$ shape, so the pooling head is unchanged.
\end{comment}

\paragraph{\seanet{} Bottleneck pointwise mixing:}
In the separable unit, the depthwise filters are cheap because each channel has one small kernel. Most of the parameters are in the pointwise mixing $P$ of \Cref{eq:unit}, which costs $\dmodel^{2}$ weights per unit. We therefore factorize $P$ into two smaller $1\times1$ convolutions: the first squeezes the $\dmodel$ channels to a width $r<\dmodel$, and the second expands them back: 

\begin{equation}
	P(\mathbf{h}) \;=\; P_{\text{exp}}\big(P_{\text{sq}}(\mathbf{h})\big),
	\qquad
	P_{\text{sq}}:\mathbb{R}^{\dmodel}\!\to\!\mathbb{R}^{r},\quad
	P_{\text{exp}}:\mathbb{R}^{r}\!\to\!\mathbb{R}^{\dmodel},
	\qquad
	\frac{2\dmodel r}{\dmodel^{2}}=\frac{2r}{\dmodel}.
	\label{eq:bottleneck}
\end{equation}
Because there is no non-linearity between the two convolutions, the pair is equivalent to a single mixing matrix of rank $r$. This leaves the depth of the block unchanged and adds no extra layer. With $r=\dmodel/4$, the pointwise cost drops from $\dmodel^{2}$ to $2\dmodel r=\dmodel^{2}/2$, a halving at any width. At $\dmodel=64$ this reduces the whole model from $57{,}580$ to $41{,}324$ parameters ($-28\%$). The block still outputs a matrix of shape $\mathbb{R}^{t \times d}$, so the pooling head needs no change.

\paragraph{\seanet{} Channel re-weighting variants:}
Both variants introduce a channel re-weighting layer after the last \emph{multi-scale separable residual} block. The purpose of this layer is to adaptively adjust the contribution of the $d$ feature channels according to their relevance to the input time series, while leaving the encoder blocks and the MIL pooling head unchanged. In both variants, the output of the sixth block, $Z_i\in\mathbb{R}^{t\times d}$, is first passed through the channel re-weighting layer. Each time-step embedding $z_i^j$, $j=1,\ldots,t$, is then re-weighted according to:

\begin{equation}
	\tilde{z}_i^{\,j}=z_i^j\odot(1+g_i^{\,j}),
	\qquad
	\underbrace{g_i^{\,j}=\sigma(W_g\,s_i)}_{\text{Gated (mean)}},\;\;
	s_i=\tfrac{1}{t}\textstyle\sum_{j} z_i^j
	\qquad
	\underbrace{g_i^{\,j}=\sigma(W_g * X_i)^j}_{\text{Input Gated}},
	\qquad g_i^{\,j}\in[0,1]^d .
	\label{eq:generic_gate}
\end{equation}
where $g_i^j\in[0,1]^d$ contains one weight for each feature channel. The use of $(1+g_i^j)\in[1,2]^d$, rather than $g_i^j$ alone, preserves the original features while allowing informative channels to be amplified. This also preserves the differences between time-step embeddings, which are used to compute the interpretation score $\hat{I}_i$. Since the re-weighting is applied only along the channel dimension, the temporal structure of $Z_i$ remains unchanged. Consequently, both the following variants produce an output $\tilde{Z}_i\in\mathbb{R}^{t\times d}$ that can be directly passed to the same MIL pooling head.

\subparagraph{\seanet{} Gated (mean):}
\label{eq:gate_mean}
In the first variant, a single set of channel weights is computed for the entire time series. The output $Z_i$ of the sixth encoder block is first averaged over its $t$ time points to obtain a series-level representation.
%\begin{equation}
%	s_i=\frac{1}{t}\sum_{j=1}^{t}z_i^j\in\mathbb{R}^{d}.
%\end{equation}
This representation is passed through a linear transformation $W_g\in\mathbb{R}^{d\times d}$ followed by a sigmoid function to obtain one weight for each channel:
%\begin{equation}
%	g_i=\sigma(W_gs_i)\in[0,1]^d
%	\label{eq:gate_mean}
%\end{equation}
%{\color{violet}\rdt{XVIII}  The same channel weights are then applied to every time-step embedding:
	%\begin{equation}
	%	\color{violet}
	%	\tilde{z}_i^{\,j}=z_i^j\odot(1+g_i)
	%	\qquad j=1,\ldots,t
	%\end{equation}}
These weights are applied to every time-step embedding as in \Cref{eq:generic_gate}.
Thus, this variant determines which feature channels are generally more informative for the entire time series. However, the same weights are applied at every time point and therefore cannot account for changes in the importance of a channel over time. 
%The mean operation is one of three possible strategies for obtaining the series-level representation; maximum and last-time-point operations are also evaluated and reported in the supplementary material. 
The mean-based version is used in the main experiments (others in \Cref{app:gates}) 
	
\subparagraph{\seanet{} Input Gated:}

The second variant performs channel re-weighting at each time point. Instead of computing the weights from the deep representation $Z_i$, they are obtained directly from the raw input time series $X_i$. A 1D convolution with kernel size $7$ transforms the single input channel into $d$ channels, followed by a sigmoid function to produce a separate set of channel weights $g_i^j\in[0,1]^d$ for each time point:
%\begin{equation}
%	g_i^j=\sigma\big(W_g*X_i\big)^j\in[0,1]^d
%	\qquad
%	\tilde{z}_i^{\,j}=z_i^j\odot(1+g_i^j)
%	\quad j=1,\ldots,t
%	\label{eq:input_gate}
%\end{equation}
where $W_g*X_i$ denotes the 1D convolution.
% with ``same'' zero padding, ensuring that the resulting weights contain exactly $t$ time points. 
Unlike the previous variant, the channel weights can therefore vary both across channels and across time points. Because they are computed directly from the raw input, they can respond to local events, such as a sudden peak, before these patterns are potentially smoothed by the encoder's receptive field.
%Overall, the two variants differ in the granularity and source of the channel weights. \seanet{} Gated (mean) computes a single set of channel weights for the entire time series from the deep features $Z_i$, whereas \seanet{} Input Gated computes a separate set of channel weights for every time point directly from the raw input $X_i$. 
These two variants allow us to study the effect of coarse, series-level channel re-weighting versus fine-grained, time-dependent channel re-weighting while keeping the remainder of the architecture unchanged (\Cref{app:encoders}).

\paragraph{\seanet{} Spike/Trend:}

A stack of convolutions followed by batch normalization is biased toward smooth and sustained patterns, whereas in some datasets the evidence of a class is a single sharp discontinuity (i.e., a spike) in the time series. In this variant, we therefore run a second, high-pass-filter-like branch in parallel with the encoder blocks and merge the two branches before the pooling head. The output of the encoder is called the \emph{trend} features, and the output of this new high-pass-filter-like branch is called the \emph{spike} features.
In \Cref{fig:arch_overall}, the encoder path (the stem convolution and the six blocks) is untouched and gives the trend features $Z_i \in \mathbb{R}^{t \times d}$. 
The new spike branch takes the original time series $X_i$ as input, applies a moving average $\operatorname{avg}_w(X_i)$ with window $w = 5$, and then subtracts this smoothed signal from the original time series.
This subtraction removes the smooth, slowly varying component of the signal, leaving a residual that is close to zero in smooth regions and becomes large around sudden changes or discontinuities. 
Then a convolution with kernel size $3$, followed by BN and ReLU layers, is applied, which turns the output into $d_{\text{spike}} = 16$ channels.
Finally, at each time point, the trend features and spike features are concatenated and projected back to $d$ channels using the pointwise convolution $P$, as defined in \Cref{eq:unit}. The resulting representation is then aggregated using the mean operation described in \Cref{eq:generic_gate}.
\vspace{-1mm}
\begin{align}
	\text{spike}_i^{\,j} &= \operatorname{ReLU}\Big(\operatorname{BN}\big(\operatorname{Conv}_3(X_i - \operatorname{avg}_w(X_i))\big)\Big)^j \in \mathbb{R}^{d_{\text{spike}}}
	\\\tilde{z}_i^{\,j} &= P\big([\,z_i^j \,;\, \text{spike}_i^{\,j}\,]\big) \odot (1 + g_i)
	\label{eq:spike}
\end{align}

where $[\,\cdot\,;\,\cdot\,]$ is the concatenation along the channel axis, giving $d + d_{\text{spike}}$ channels, and $g_i$ is computed as in \Cref{eq:generic_gate} on the mixed features.
With this branch, the pooling head directly receives features which mark the discontinuities of the series. Hence, we expect an advantage on datasets where the class evidence has impulsive patterns. 

\subsection{MIL Pooling Head}
\label{sec:methods:pooling}
\label{sec:method:head}

With the encoder complete, we now describe the second component of \seanet{}, the MIL pooling head $\psi_{pool}$, shown on the right side of \Cref{fig:arch_overall}. Given the instance-level embeddings
$Z_i\in\mathbb{R}^{t\times d}=[z_i^1,\ldots,z_i^t]$, the pooling head has two branches: an attention branch that determines the importance of each time point, and a classifier branch that determines the class evidence provided by each time point. Their product defines the evidence $e_i^{j,k}$ of time point $j$ for class $k$. Aggregating this evidence over time produces the class prediction $\hat{Y}_i\in\mathbb{R}^{1\times C}$, while retaining the evidence at each time point gives the interpretation score $\hat{I}_i\in\mathbb{R}^{t\times C}$. 
%Thus, the interpretation is directly derived from the prediction: both are computed from exactly the same time-point-level evidence.
Our pooling head is based on the \texttt{Conjunctive} pooling of \millet{} (\Cref{eq:conj_pool}) \citep{early2024millet}, which uses a single attention weight $a_i^j$ for each time point and a class-wise classifier output $\hat{y}_i^{\,j}$:

\vspace{-2.8mm}
\noindent
\begin{minipage}[t]{0.48\linewidth}
	\begin{equation}
		\begin{aligned}
			a_i^j &= \sigma\big(\mathbf{w}_a^\top \tanh(W_a z_i^j)\big) \in [0,1]\\
			\hat{y}_i^{\,j} &= W_c z_i^j + \mathbf{b}_c \in \mathbb{R}^{C}\\
			\hat{Y}_i^{\,k} &= \tfrac{1}{t}\textstyle\sum_{j=1}^{t} a_i^j \hat{y}_i^{\,j,k},
			\quad \hat{I}_i^{\,j,k} = a_i^j \hat{y}_i^{\,j,k}
		\end{aligned}
		\label{eq:conj_pool}
	\end{equation}
\end{minipage}\hfill
\begin{minipage}[t]{0.48\linewidth}
	\begin{equation}
		\begin{aligned}
			a_i^j &= \sigma\big(W_2 \tanh(W_1 z_i^j)\big) \in [0,1]^C\\
			e_i^{j,k} &= a_i^{j,k}\hat{y}_i^{\,j,k},
			\quad \hat{I}_i^{\,j,k} = e_i^{j,k}\\
			\hat{Y}_i^{\,k} &= \tfrac{1}{t}\textstyle\sum_{j=1}^{t} e_i^{j,k}
		\end{aligned}
		\label{eq:classwise_pool}
	\end{equation}
\end{minipage}
%\begin{equation}
%	\begin{split}
%		a_i^j &= \sigma\big(\mathbf{w}_a^\top \tanh(W_a z_i^j)\big) \in [0,1]
%		\qquad
%		\hat{y}_i^{\,j} = W_c z_i^j + \mathbf{b}_c \in \mathbb{R}^{C} \\
%		\hat{Y}_i^{\,k} &= \frac{1}{t}\sum_{j=1}^{t}
%		a_i^j \hat{y}_i^{\,j,k}
%		\qquad
%		\hat{I}_i^{\,j,k} = a_i^j \hat{y}_i^{\,j,k}
%	\end{split}
%	\label{eq:conj_pool}
%\end{equation}
\vspace{2.8mm}

where $W_a\in\mathbb{R}^{d_a\times d}$ and $\mathbf{w}_a\in\mathbb{R}^{d_a}$ define the attention branch, while $W_c\in\mathbb{R}^{C\times d}$ and $\mathbf{b}_c\in\mathbb{R}^{C}$ define the classifier branch; $W_1\in\mathbb{R}^{d_a\times d}$ and $W_2\in\mathbb{R}^{C\times d_a}$ are its class-wise counterparts below.
%\paragraph{Limitations of conjunctive pooling.}

Conjunctive pooling has three limitations that motivate our pooling variants. First, the scalar attention weight $a_i^j$ is shared across all $C$ classes and therefore cannot represent class-specific importance. Second, the sigmoid weights are not normalized over time: although $a_i^j\in[0,1]$, their sum depends on the series length $t$, while the aggregation divides by $t$. Consequently, the magnitude of $\hat{Y}_i^{\,k}$ can depend on the series length, which is particularly relevant for the \ucr{} archive, where series lengths range from 15 to 2{,}844. Third, averaging all $t$ evidence values can dilute a small number of highly informative time points when most of the series contains background information.
%\paragraph{\seanet{} Pooling Variants}
We developed seven pooling variants; here we describe the three that directly address the limitations above, while the remaining four are presented in \Cref{app:heads}. All three variants retain the two-branch structure and the classifier output $\hat{y}_i^{\,j}$ of \Cref{eq:conj_pool}. They differ only in how the attention weights are computed and/or how the resulting evidence is aggregated. In every case, the interpretation $\hat{I}_i$ is obtained from the same evidence used to compute $\hat{Y}_i$, ensuring consistency between prediction and interpretation.

\paragraph{Class-wise / Soft-Max / Top-$k$ conjunctive pooling:}
The \texttt{Class-wise} variant (\Cref{eq:classwise_pool}) addresses the first limitation by replacing the single scalar attention weight with a class-specific vector:
%\begin{equation}
%	\begin{split}
%		a_i^j &= \sigma\big(W_2\tanh(W_1z_i^j)\big)\in[0,1]^C
%		\qquad
%		W_1\in\mathbb{R}^{d_a\times d},\quad
%		W_2\in\mathbb{R}^{C\times d_a}, \\
%		e_i^{j,k} &= a_i^{j,k}\hat{y}_i^{\,j,k}
%		\qquad
%		\hat{Y}_i^{\,k}
%		= \frac{1}{t}\sum_{j=1}^{t}e_i^{j,k}
%		\qquad
%		\hat{I}_i^{\,j,k}=e_i^{j,k}
%	\end{split}
%	\label{eq:classwise_pool}
%\end{equation}
Each time point can therefore have a different importance for different classes. The modification only widens the final attention layer, adding $(C-1)(d_a+1)$ parameters compared with the original attention branch. If $a_i^{j,k}=a_i^j$ for all $k$, the formulation reduces exactly to the original \texttt{Conjunctive} pooling.
%\subparagraph{Soft-Max Conjunctive Pooling with Learnable Sharpness:}
The \texttt{Soft-Max} variant addresses the second limitation by normalizing the attention weights across time. For each class, the attention scores are converted into a probability distribution using a Soft-Max with learnable temperature $\tau>0$ (\Cref{eq:softmax_pool}):

\vspace{-2.8mm}
\noindent
\begin{minipage}[t]{0.49\linewidth}
	\begin{equation}
		\begin{aligned}
			s_i^{j,k} &= \big(W_2 \tanh(W_1 z_i^j)\big)_k\\
			\alpha_i^{j,k} &= \frac{\exp(s_i^{j,k}/\tau)}{\sum_{l=1}^{t}\exp(s_i^{l,k}/\tau)}\\
			\hat{Y}_i^{\,k} &= \textstyle\sum_{j=1}^{t}\alpha_i^{j,k}\hat{y}_i^{\,j,k},
			\quad \hat{I}_i^{\,j,k} = \alpha_i^{j,k}\hat{y}_i^{\,j,k}
		\end{aligned}
		\label{eq:softmax_pool}
	\end{equation}
\end{minipage}\hfill
\begin{minipage}[t]{0.49\linewidth}
	\begin{equation}
		\begin{aligned}
			m &= \max\big(1,\lceil\kappa t\rceil\big)\\
			\mathcal{S}_i^k &= \operatorname{top}_m\big(e_i^{1,k},\ldots,e_i^{t,k}\big)\\
			\hat{Y}_i^{\,k} &= \tfrac{1}{m}\textstyle\sum_{j\in\mathcal{S}_i^k} e_i^{j,k},
			\quad \hat{I}_i^{\,j,k} = e_i^{j,k}
		\end{aligned}
		\label{eq:topk_pool}
	\end{equation}
\end{minipage}
\vspace{2.8mm}

%\begin{equation}
%	\begin{split}
%		s_i^{j,k} &= \big(W_2\tanh(W_1z_i^j)\big)_k
%		\qquad
%		\alpha_i^{j,k}
%		=
%		\frac{\exp(s_i^{j,k}/\tau)}
%		{\sum_{l=1}^{t}\exp(s_i^{l,k}/\tau)} \\
%		\hat{Y}_i^{\,k}
%		&= \sum_{j=1}^{t}\alpha_i^{j,k}\hat{y}_i^{\,j,k}
%		\qquad
%		\hat{I}_i^{\,j,k}
%		= \alpha_i^{j,k}\hat{y}_i^{\,j,k}
%	\end{split}
%	\label{eq:softmax_pool}
%\end{equation}

Because $\sum_j\alpha_i^{j,k}=1$, the prediction is no longer directly affected by the series length, making series of different lengths more comparable. The temperature controls the concentration of the attention: large $\tau$ produces a more uniform distribution approaching $1/t$, whereas small $\tau$ concentrates the weight on the highest-scoring time points. We parameterize $\tau=\exp(\rho)$ with learnable $\rho$, ensuring $\tau>0$ while introducing only one additional parameter.

%\subparagraph{Top-$k$ Conjunctive Pooling:}
The \texttt{Top-$k$ conjunctive} variant addresses the third limitation by aggregating only the most informative time points rather than averaging the evidence over the entire series. It uses the class-wise attention of \Cref{eq:classwise_pool} and retains only the $m$ largest evidence values for each class (\Cref{eq:topk_pool}):
%\begin{equation}
%	\begin{split}
%		m &= \max(1,\lceil\kappa t\rceil), \\
%		\mathcal{S}_i^k
%		&=
%		\left\{
%		j:\, e_i^{j,k}\text{ is among the }m\text{ largest values of }
%		(e_i^{1,k},\ldots,e_i^{t,k})
%		\right\} \\
%		\hat{Y}_i^{\,k}
%		&= \frac{1}{m}\sum_{j\in\mathcal{S}_i^k}e_i^{j,k}
%		\qquad
%		\hat{I}_i^{\,j,k}=e_i^{j,k}
%	\end{split}
%	\label{eq:topk_pool}
%\end{equation}
Here, $\kappa\in[0,1]$ controls the fraction of time points retained and introduces no additional learnable parameters. When $\kappa=1$, this formulation reduces to the class-wise pooling of \Cref{eq:classwise_pool}; for $\kappa<1$, the aggregation focuses on the most informative time points, reducing the influence of background regions. Consequently, the prediction is based on an explicit subset of time points, which also makes the resulting interpretation more focused. This is the pooling head used in our best-performing models.

Every encoder and head above shares the same input and output shapes, so any pair combines from configuration alone. \Cref{sec:setup_new} evaluates these combinations, taking the bottleneck encoder with Top-$k$ pooling as our main model.

% !TEX root = ../main.tex
% =====================================================================
% 4. EXPERIMENTAL SETUP -- page budget 0.60
%
% Short, and it carries the two warnings that a reviewer would otherwise
% raise: WebTraffic was also the set we chose the design on, and the seed
% counts are uneven. Both are stated here, once, so the results section
% can be read as it is.
% =====================================================================

%\section{Experimental Setup}
%\label{sec:setup}

%\todo{Rewrite the experiment setup}
\section{Experimental Setup and Results}
\label{sec:setup_new}
\label{sec:setup}
%Before going to the results, we first describe the research questions which we answered, datasets we used, the metrics used for the evaluation in the following section.

\textbf{Research questions and datasets:}
We ask whether the small model retains classification accuracy, retains a useful
interpretation, and how much cost it actually removes; two further questions, \textbf{RQ4} what the
bottleneck contributes and \textbf{RQ5} what \mbox{Top-$k$} pooling contributes, are answered by the
ablation of \Cref{sec:res:ablation}. We answer them on two datasets (\Cref{tab:protocol}).
%\webtraffic{} is the synthetic benchmark released with \millet{} \citep{early2024millet}: 1{,}000
%series (500 train / 500 test) of length 1{,}008 over 10 classes, in which a generator puts the class
%signal inside a known window. It is the only dataset here with \emph{per-time-step ground truth}, so
%it is the only place where we can score an interpretation instead of just looking at it, and RQ2 is
%answered there. For wider coverage we use the \ucr{} archive \citep{dau2019ucr}: we report the
%85-dataset subset used by \millet{}, so that the numbers can be compared, and the full 128-dataset
%archive in \Cref{app:ucr}.

We use the same \webtraffic{} dataset as \millet{} \citep{early2024millet}: a synthetic set of $1{,}000$ series of length $1{,}008$ over $10$ classes, split $500/500$, in which each class signature is injected inside a known window (\Cref{fig:web_classes}). It is the only dataset with a ground truth per time point, hence the interpretation can be scored rather than merely inspected.
%The properties of this dataset, and of all the other datasets used in this work, are listed in \Cref{tab:datasets_all}
%\textcolor{red}{\textbf{The \ucr{} archive.}}
For wider coverage we use the \ucr{} archive \citep{dau2019ucr}, reporting the $85$-dataset subset used by \millet{} so that the numbers are comparable; we also report the performance on the full $128$-dataset archive in \Cref{app:ucr}. Class signatures are not annotated there, so on \ucr{} the interpretation can only be evaluated with AOPCR.

%\paragraph{Datasets}
%
%\subparagraph{\webtraffic{} Dataset}
%We use exactly the same \webtraffic{} dataset as the one used in the original paper of \millet{} \citep{early2024millet}. It is a synthetic dataset with $10$ classes, and it contains $1{,}000$ time series of length $1{,}008$, out of which $500$ are used for the training and $500$ for the testing. In each time series, the signature of its class is injected inside a known window, as shown in \Cref{fig:web_classes}. This is the only dataset of this work which provides a ground truth for each time point, so it is also the only place where we can score the interpretation of the model instead of only looking at it. The properties of this dataset, and of all the other datasets used in this work, are listed in \Cref{tab:datasets_all}.

%\begin{figure}[!htbp]
%  \centering
%  \includegraphics[width=0.95\linewidth]{selected_figures/01_main/fig11_webtraffic_classes/fig11_webtraffic_classes.png}
%  \caption{One example of each class of the \webtraffic{} dataset. Each series is a daily seasonal
%  signal with a weekly trend and noise, and the signature that defines the class is injected in a
%  single window, shown in yellow; the only exception is the class 0 (None), which has no signature.
%  These windows are the ground truth of each time point (\Cref{tab:datasets_all}) against which
%  \ndcg{} is scored.}
%  \label{fig:web_classes}
%\end{figure}

%\paragraph{Metrics}
We use the test accuracy for the classification, and \emph{AOPCR} \citep{samek2017aopc} together with \emph{\ndcg{}} \citep{jarvelin2002ndcg} for the interpretation; both metrics are defined in \Cref{app:metrics}. \emph{AOPCR} progressively removes time points according to their estimated importance and accumulates the resulting decrease in the score of the predicted class. Computing \emph{AOPCR} does not require ground truth, so we can report this metric on both datasets.
% {\color{violet}\rdt{XX}  but it has no fixed scale, and \Cref{sec:res:aopcr} shows what this costs.} 
 \ndcg{} ranks the time points by the importance the model assigns them and compares that ranking with the ground truth.

\subsection{Classification and Interpretation Results}
\label{sec:res:main}
\label{sec:res:ucr}
\label{sec:res:aopcr}
\label{sec:res:ablation}
%\paragraph{Classification and Interpretation on \webtraffic{}}
\textbf{\webtraffic{}:} \Cref{tab:web_main} gives one complete model per row, written as \emph{encoder $+$ pooling head}.
%The results are given in \Cref{tab:web_main}, where one row is one complete model, written as \emph{encoder $+$ pooling head}, and the columns are the accuracy, the AUROC and the NLL for the classification, the AOPCR and the \ndcg{} for the interpretation, and the number of parameters.
%\Cref{fig:web_bars} shows the same models across every metric we measured, with the quality panels and the cost panels side by side. 
Our \seanet{} model exceeds \millet{}'s accuracy and achieves a better \ndcg{}, while being about ten times smaller: \textbf{0.931} $\pm$ 0.032 accuracy and \textbf{0.765} $\pm$ 0.013 \ndcg{} with $41{,}324$ parameters, against $0.888 \pm 0.007$ and $0.681 \pm 0.013$ with $423{,}707$ parameters for \millet{} under the same budget (\Cref{tab:web_main}). The gated (mean) $+$ Top-$k$ variant is more accurate and far more stable ($0.955 \pm 0.002$), but needs $61{,}740$ parameters and a lower \ndcg{} ($0.750 \pm 0.018$), so the bottleneck model stays our headline. Two further columns point the same way. Our negative log-likelihood is $0.312 \pm 0.063$ against $0.513 \pm 0.019$ for the matched-budget baseline, a $39\,\%$ reduction, and it stays below \millet{}'s own $1{,}500$-epoch recipe ($0.392 \pm 0.021$) and its published figure ($0.379$); AUROC moves the same way, $0.996 \pm 0.002$ against $0.988 \pm 0.002$. The $41$\,K model is therefore better calibrated than the baseline even when the baseline is given nearly four times the training budget.
%\ucr{} was never used to choose anything, and it is where the trade-off shows.
\textbf{\ucr{} archive:}
On \ucr{}, the comparison depends on the size budget. Our $41$\,K model reaches
$0.810$ mean accuracy, $1.8$ points below the matched-budget \millet{} baseline
($0.827$) and $3.5$ points below \millet{}'s published accuracy ($0.845$), at a tenth
of the parameters and with mean rank $15.40$ against $12.60$. Counted per dataset
against \millet{}'s published accuracies, it wins on $20$, ties on $15$ and loses on
$50$ of the $85$, where the ten-times-larger matched-budget baseline scores
$13/25/46$: more outright wins than the baseline, but also more losses
(\Cref{tab:ucr85_rank}).
The gap comes from size, not from design. Re-running the \emph{same} architecture,
with the same blocks and head, in a \emph{wide} configuration ($\dmodel$ from 64 to 128,
blocks from 4 to 6) reaches $0.829$ accuracy and mean rank $11.53$, the best rank in
\Cref{tab:ucr85_rank}: ahead of the matched-budget \millet{} baseline on accuracy
($0.827$), on rank ($12.60$) and on wins ($26$ against $13$), with $36\,\%$ fewer
parameters. At a comparable size and the same training budget, replacing the \millet{}
backbone by ours is worth about $1.1$ rank places. The \seanet{} design is therefore not
what loses accuracy on \ucr{}; the smaller parameter budget is.
%
%Two comparisons isolate the two factors, each measured while the other is held fixed.
%At a comparable model size and a fixed training budget, replacing the \millet{}
%backbone by ours improves the mean rank from $12.60$ to $11.53$, about $1.1$ places.
%For a fixed architecture, extending the training budget from our 400 epochs to
%\millet{}'s 1{,}500 epochs improves it from $12.60$ to $9.41$, about $3.2$ places,
%three times as much. Any backbone comparison run at unequal budgets is therefore
%largely measuring the budget rather than the architecture.

% tab01_webtraffic_main - corrected layout (Params column removed, all data columns centred).
\begin{table}[tbp]
\caption{Classification and explanation quality on WebTraffic, each model written as \emph{encoder $+$ pooling head}. Rows marked $^\ast$ are mean $\pm$ s.d. over the \emph{Runs} column. Higher is better except for NLL. AOPCR is unnormalised, so it compares only models trained under the same recipe; rows marked $^\dagger$ use MILLET's recipe (1500 epochs) and are never marked best. Run conventions in Appendix~\ref{app:hparams}, parameter counts in Table~\ref{tab:cost}.}
\label{tab:web_main}
\begin{center}
\footnotesize
\setlength{\tabcolsep}{4.5pt}
\renewcommand{\arraystretch}{1.15}
\ifdefined\seatablebox\else\newsavebox\seatablebox\fi
\sbox\seatablebox{\begin{tabular}{lcccccc}
\toprule
\textbf{Model (encoder $+$ pooling head)} & \textbf{Runs} & \textbf{Accuracy} & \textbf{AUROC} & \textbf{NLL} & \textbf{AOPCR} & \textbf{NDCG@$n$} \\
\midrule
\multicolumn{7}{l}{\emph{SEA-Net (ours): encoder $+$ Top-$k$ conjunctive head}} \\
gated (mean) + Top-$k$ conj.$^\ast$ & 3 & $\mathbf{0.955 } \pm \mathbf{ 0.002}$ & $\mathbf{0.998 } \pm \mathbf{ 0.000}$ & $\mathbf{0.254 } \pm \mathbf{ 0.010}$ & $2.225 \pm 0.107$ & $0.750 \pm 0.018$ \\
bottleneck + Top-$k$ conj. ($\lambda_{\text{focus}}=0$) & 1 & 0.950 & \textbf{0.998} & 0.263 & 2.303 & 0.765 \\
spike/trend + Top-$k$ conj. & 1 & 0.950 & 0.997 & 0.288 & 2.316 & 0.756 \\
input-gated + Top-$k$ conj. & 1 & 0.946 & 0.997 & 0.261 & 2.803 & 0.748 \\
bottleneck + Top-$k$ conj. & 1 & 0.942 & 0.996 & 0.294 & 2.951 & \textbf{0.786} \\
bottleneck + Top-$k$ conj.$^\ast$ & 3 & $0.931 \pm 0.032$ & $0.996 \pm 0.002$ & $0.312 \pm 0.063$ & $2.498 \pm 0.149$ & $0.765 \pm 0.013$ \\
\midrule
\multicolumn{7}{l}{\emph{Baselines re-trained here under the same recipe}} \\
MILLET (InceptionTime + conjunctive)$^\ast$ & 4 & $0.888 \pm 0.007$ & $0.988 \pm 0.002$ & $0.513 \pm 0.019$ & $2.457 \pm 0.655$ & $0.681 \pm 0.013$ \\
ResNet + conjunctive & 1 & 0.772 & 0.957 & 0.824 & 2.952 & 0.554 \\
FCN + conjunctive & 1 & 0.742 & 0.955 & 0.910 & \textbf{3.826} & 0.533 \\
\midrule
\multicolumn{7}{l}{\emph{MILLET under its own recipe}} \\
MILLET, re-trained here with their recipe$^\ast$ & 2 & $0.918 \pm 0.002$ & $0.994 \pm 0.001$ & $0.392 \pm 0.021$ & $12.060 \pm 1.208$$^\dagger$ & $0.643 \pm 0.019$ \\
MILLET, published (mean of 5 seeds) & 5 & 0.924 & 0.994 & 0.379 & 12.768$^\dagger$ & 0.674 \\
MILLET, published 5-model ensemble & 1 & 0.940 & 0.996 & 0.277 & 11.100$^\dagger$ & 0.694 \\
\bottomrule
\end{tabular}}
\ifdim\wd\seatablebox>\linewidth
  \resizebox{\linewidth}{!}{\usebox\seatablebox}
\else
  \usebox\seatablebox
\fi
\end{center}
\end{table}

% tab02_ucr85_summary - corrected layout (Params and "vs. MILLET" columns removed, all data columns centred).
\begin{table}[tbp]
\caption{Performance on the 84 UCR datasets shared by every row, as \emph{mean / average rank / number of wins}. Higher is better except for NLL. Our models and the re-trained baselines share one training recipe; the published MILLET rows use their own (1500 epochs) and are ranked with the rest. Ranking rules, the full protocol and parameter counts are in Appendix~\ref{app:hparams} and Table~\ref{tab:cost}.}
\label{tab:ucr_accuracy}
\label{tab:main}
\label{tab:ablation}
\begin{center}
\footnotesize
\setlength{\tabcolsep}{6pt}
\renewcommand{\arraystretch}{1.15}
\ifdefined\seatablebox\else\newsavebox\seatablebox\fi
\sbox\seatablebox{\begin{tabular}{lcccc}
\toprule
\textbf{Model} & \textbf{Accuracy} & \textbf{Bal.\ accuracy} & \textbf{AUROC} & \textbf{NLL} \\
\midrule
\multicolumn{5}{l}{\emph{SEA-Net (ours), our recipe}} \\
gated (mean) + Top-$k$ conj. & 0.806 / 9.17 / 11 & 0.778 / 9.38 / 11 & 0.916 / 9.39 / 11 & 0.561 / 7.28 / 3 \\
base (wide, $d=128$) + (MILLET) & 0.822 / 7.75 / 13 & 0.797 / 7.64 / 13 & 0.927 / 7.60 / 15 & 0.522 / 6.35 / 7 \\
gated (max) + Top-$k$ conj. & 0.809 / 8.73 / 11 & 0.783 / 8.61 / 10 & 0.916 / 8.38 / 13 & 0.551 / 6.92 / 6 \\
spike/trend + Top-$k$ conj. & 0.807 / 8.44 / 9 & 0.781 / 8.32 / 9 & 0.913 / 8.57 / 12 & 0.551 / 7.06 / 6 \\
reconstruction-residual + adaptive class-wise & 0.812 / 8.96 / 11 & 0.784 / 8.84 / 10 & 0.922 / 8.74 / 12 & 0.555 / 7.95 / 4 \\
base (wide, $d=128$) + class-wise & \textbf{0.827 / 7.32 / 16} & 0.802 / 7.10 / 15 & 0.925 / 7.96 / 16 & \textbf{0.516 / 6.60 / 8} \\
input-gated + Top-$k$ conj. & 0.811 / 8.29 / 12 & 0.786 / 8.27 / 11 & 0.922 / 8.02 / 16 & 0.541 / 6.29 / 4 \\
bottleneck + Top-$k$ conj. & 0.808 / 9.10 / 7 & 0.781 / 9.31 / 7 & 0.917 / 9.47 / 7 & 0.556 / 7.29 / 2 \\
\midrule
\multicolumn{5}{l}{\emph{Baselines re-trained here}} \\
MILLET (InceptionTime + conjunctive), our recipe & 0.827 / 8.08 / 5 & 0.800 / 8.14 / 5 & 0.927 / 7.96 / 9 & 0.530 / 7.45 / 4 \\
%MILLET (InceptionTime + conjunctive), their recipe & \textbf{0.843 / 5.82 / 21} & 0.821 / 5.59 / 21 & \textbf{0.936 / 5.49 / 22} & 1.170 / 9.43 / 15 \\
ResNet + conjunctive & 0.813 / 8.23 / 15 & 0.782 / 8.76 / 12 & 0.923 / 8.03 / 14 & 0.575 / 8.44 / 3 \\
FCN + conjunctive & 0.812 / 8.92 / 10 & 0.782 / 9.43 / 11 & 0.921 / 8.90 / 10 & 0.578 / 8.73 / 6 \\
\midrule
\multicolumn{5}{l}{\emph{MILLET, published \citep{early2024millet}}} \\
Conjunctive FCN, published (mean of 5 seeds) & 0.828 / 8.12 / 10 & 0.803 / 8.02 / 9 & 0.928 / 7.70 / 13 & 1.059 / 9.31 / 9 \\
Conjunctive ResNet, published (mean of 5 seeds) & 0.834 / 7.29 / 10 & 0.811 / 7.15 / 9 & 0.930 / 7.35 / 9 & 1.186 / 10.79 / 2 \\
Conjunctive InceptionTime, published (mean of 5 seeds) & \textbf{0.843 / 5.79 / 12} & \textbf{0.822 / 5.45 / 13} & 0.932 / 6.44 / 13 & 1.232 / 10.12 / 5 \\
\bottomrule
\end{tabular}}
\ifdim\wd\seatablebox>\linewidth
  \resizebox{\linewidth}{!}{\usebox\seatablebox}
\else
  \usebox\seatablebox
\fi
\end{center}
\end{table}

\textbf{RQ4, the bottleneck:} \Cref{tab:leaderboard_sweep}, one seed per row,
	isolates the encoder: the two rows of each pair share the same pooling head and
	differ only in whether the bottleneck is used. Replacing the plain encoder by the
	bottleneck encoder removes $16{,}256$ parameters, that is $28\,\%$ of the $57{,}580$
	of the plain model. Accuracy then changes by $-0.014$ with the class-wise head and by
	$-0.001$ with the \mbox{Top-$k$} head, and \ndcg{} by $+0.014$ and $-0.013$
	respectively. The accuracy changes are well inside the three-run spread of
	$\pm 0.032$ and the \ndcg{} changes sit at its $\pm 0.013$ spread
	(\Cref{tab:web_main}), so none can be resolved from these single-seed runs. We therefore claim
	only that the bottleneck delivers the size reduction \emph{at no measurable cost in
		accuracy or interpretation}.
%		; we do not claim that it improves either.

\textbf{RQ5, \mbox{Top-$k$} pooling:} Read the other way, the same
	table isolates the head: the two rows of each pair share the same encoder and differ
	only in the pooling head. Replacing the class-wise head by \mbox{Top-$k$} raises
	\ndcg{} from $0.719$ to $0.778$ on the plain encoder ($+0.059$) and from $0.733$ to
	$0.765$ on the bottleneck encoder ($+0.032$), two to five times the $\pm 0.013$
	seed spread. This gain in interpretation ranking is the claim we make for the head.
	Accuracy moves the same way, by $+0.016$ and $+0.029$, but both sit inside the
	$\pm 0.032$ accuracy spread, so we do not present \mbox{Top-$k$} as an accuracy gain.
%	but every cell of
%	this table is a single seed, which is not enough to separate a genuine accuracy gain
%	from seed noise; we therefore do not present \mbox{Top-$k$} as an accuracy
%	improvement.

\Cref{tab:ablation_topk} varies only the selection fraction $\kappa$, with the
	architecture and the training recipe unchanged. By \Cref{prop:topk}, the $\kappa = 1$
	row \emph{is} the class-wise baseline head, so the difference between its \ndcg{} of
	$0.722$ and the $0.755$ reached at $\kappa = 0.10$ comes from the selection alone and
	not from a change of model. Two caveats follow. First, $\kappa$ has a best value
	rather than a monotone trend: $\kappa = 0.05$ is worse than $\kappa = 0.10$ on AOPCR
	and \ndcg{} and no better on accuracy, so keeping too few time points hurts as much
	as keeping too many. Second, AOPCR does not track $\kappa$ the way \ndcg{} does,
	peaking at $\kappa = 0.50$ ($3.114$) where \ndcg{} is among its lowest ($0.732$),
	which is one more reason to read AOPCR with caution (\Cref{sec:res:aopcr}).
	\Cref{tab:ablation_focus} switches off the attention-entropy penalty
	$\lambda_{\text{focus}}$ (\Cref{sec:setup_new}), a control independent of $\kappa$:
	removing it raises accuracy from $0.888$ to $0.950$ and \ndcg{} from $0.755$ to
	$0.765$, while AOPCR falls from $2.409$ to $2.303$. On this single seed the penalty
	does not pay for itself. As every cell here is one seed, we read these rows as a
	pattern rather than as exact values.
%\subsection{Sensitivity of AOPCR to the Training Budget}
%\label{sec:res:aopcr}
%One number in \Cref{tab:main} needs an explanation, and the explanation applies more widely.
Re-trained under \millet{}'s own 1{,}500-epoch recipe, \millet{} scores AOPCR
	$12.060 \pm 1.208$ on \webtraffic{}, against $2.457 \pm 0.655$ for the same model under
	our matched budget (\Cref{tab:web_main}). The architecture, code, metric and data are
	identical; only the recipe differs. A $4.9\times$ higher AOPCR
	would normally mean a far better explanation, but it does not here. AOPCR averages the
	drop in the predicted-class \emph{logit} as the top-ranked time points are removed
	(\Cref{app:metrics}), and logits have no fixed scale: a model trained longer is more
	confident, so every removal drops the logit further, whether or not its ranking improved.
	\ndcg{}, which scores the ranking itself and is normalized to $[0,1]$, confirms this:
	over the same pair it \emph{falls} from $0.681 \pm 0.013$ to $0.643 \pm 0.019$, a drop of
	twice its own spread. AOPCR is therefore comparable only within a
	fixed training budget, hence not across papers: we retrain every baseline under our own
	budget rather than copying published numbers, never compare against \millet{}'s published
	AOPCR, and base every interpretation claim on \ndcg{}.

\section{Discussion and Conclusion}
\label{sec:disc}
%\paragraph{The trade-off.}

Our trade-off depends on where the model runs: saving
$10.3\times$ in parameters for $1.8$ accuracy points on \ucr{} buys nothing on a
workstation, but for the micro-controller class of MCUNet \citep{lin2020mcunet} and
MicroNets \citep{banbury2021micronets} a $41$\,K model is a candidate where a $424$\,K
one is not, and the per-time-step interpretation is kept. The price is paid for the
size and not the design, since widening the same encoder recovers the \ucr{} gap
(\Cref{sec:res:ucr}).
Our costs are parameter counts, analytic FLOPs and GPU timings;
nothing was quantized or run on a micro-controller (\Cref{tab:cost}), so deployment fitness is argued
from size rather than shown. Activation memory does not shrink, because the encoder
preserves length: peak memory exceeds the baseline's, and where SRAM rather than flash
binds, streaming or strided variants are the next step. Interpretation is scored only
on \webtraffic{}, which is also our design set, and its backup metric AOPCR is
budget-sensitive (\Cref{sec:res:aopcr}). Seeds are uneven: three models use three
seeds, each ablation one, read against those spreads (\Cref{sec:setup_new}). Training
is slower, since GPU kernels favor dense over depthwise convolutions; the saving is
at inference only.
In summary, the cost of MIL lies in the encoder and not in the interpretability: our
model keeps the per-time-step explanation and beats the equally-trained \millet{}
baseline on \webtraffic{} at a tenth of the parameters. Built-in interpretability is
therefore not what makes these models large, and need not be given up to make them
small.
%\input{sections/07_conclusion}

% Statements go after the main text and before the references. None of
% them counts towards the 9-page limit.
\clearpage
% !TEX root = ../main.tex
% =====================================================================
% STATEMENTS -- after the main text, before the references. None of
% these counts towards the 9-page limit.
%
% The AI use statement is REQUIRED by ICLR 2027 and follows the wording
% pattern of the official template (iclr2027_conference.tex, "AI use
% statement"). It is filled in honestly -- an inaccurate disclosure is
% itself a Code of Ethics problem.
% =====================================================================

\section*{Reproducibility Statement}

Everything needed to reproduce the results is either in the paper or in the anonymous code release.
The models are defined in \Cref{sec:method} and, for the encoder and pooling variants not used in the
main text, in \Cref{app:encoders,app:heads}. The single training setting used for every model and
every dataset is listed in \Cref{sec:setup} and repeated with all the remaining hyperparameters in
\Cref{app:hparams}. The two evaluation metrics are defined in \Cref{sec:setup}, and the perturbation
procedure behind AOPCR, which \Cref{sec:res:aopcr} shows depends on the training recipe, is given in
\Cref{app:metrics}. \webtraffic{} is produced by the released generator of \citet{early2024millet} at
the settings in \Cref{app:hparams}; the \ucr{} archive \citep{dau2019ucr} is public and used with its
published train/test splits and no preprocessing other than standardizing each series. Seed counts
are given per row in \Cref{tab:main} and the per-seed values are in \Cref{app:seeds}. Cost figures
come from the measurement procedure in \Cref{sec:setup}; the script that produces them is in the code
release. All 72 model combinations are listed row by row in \Cref{app:leaderboard}. Anonymous code,
configurations and result tables: \anonrepo{}.

\section*{Ethics Statement}

This work collects no new data and involves no human subjects. It uses the public \ucr{} archive
\citep{dau2019ucr} under its original terms and a synthetic benchmark produced by released code; some
\ucr{} datasets come from human physiological recordings, but they are distributed in anonymized form
by their maintainers and we did not try to re-identify them or add to them. The one risk worth naming
is specific to this topic: a model that shows an explanation makes people trust it more than one that
does not, and \Cref{sec:res:aopcr} shows that a common faithfulness score can be raised by training
longer rather than by explaining better. Interpretations from models of this kind should therefore be
checked on the target domain before they are used to support decisions about people. The code we
build on is released under the Apache 2.0 license and its attribution notice is kept in our release.

 \section*{AI Use Statement}

 In this work, we used generative AI tools for writing and editing the text of this paper, for
 literature search assistance while preparing \Cref{sec:related}, and for producing supporting
 experiment-management and plotting code. We have not used generative AI tools to generate,
 select, filter or interpret experimental results, to design the models or the experimental protocol,
 or to produce any of the numbers reported in this paper. We have reviewed all AI-assisted work. Every
 numeric value in the paper was traced by the authors to the specific result file and column it came
 from before it was written down, and the tables were checked against those files a second time after
 the paper was complete; AI-assisted code was read and tested by the authors, and all reported
 measurements come from runs the authors executed themselves. Every reference was checked against the
 original publication, and claims of novelty were checked against a manual literature survey rather
 than against tool output. We take responsibility for the final content of this work, including text,
 claims or artifacts produced with the aid of generative AI.

\bibliography{refs}
\bibliographystyle{iclr2027_conference}

\appendix
\clearpage
\begin{center}
	{\LARGE\bfseries Appendix of\par}
	\vspace{0.45em}
	{\large\bfseries \papertitle\par}
\end{center}
\vspace{1.2em}

\makeatletter
\let\subsection\sea@subsection   % keep the \section barrier only
\makeatother
% a page that ends before a float or a table is not stretched to the
% bottom margin, so no large gaps open between the paragraphs
\raggedbottom
% float pages (a page holding only tables or figures) start at the top
% of the page instead of being centred vertically
\makeatletter
\setlength{\@fptop}{0pt}\setlength{\@fpsep}{12pt plus 1fil}\setlength{\@fpbot}{0pt plus 1fil}
\makeatother
% p-columns that wrap long cells: left, centred and right aligned
\newcolumntype{L}[1]{>{\raggedright\arraybackslash}p{#1}}
\newcolumntype{C}[1]{>{\centering\arraybackslash}p{#1}}
\newcolumntype{R}[1]{>{\raggedleft\arraybackslash}p{#1}}
% one table style for the whole appendix: the \small size of the main
% tables and a moderate amount of space between the rows
\newcommand{\seatab}{%
  \small
  \setlength{\tabcolsep}{3.5pt}%
  \renewcommand{\arraystretch}{1.2}%
  \setlength{\LTcapwidth}{\linewidth}%
  \setlength{\LTpre}{6pt}\setlength{\LTpost}{2pt}}

% =====================================================================
\section{Why MIL Makes \seanet{} Interpretable}
\label{app:milview}

\Cref{sec:intro} and \Cref{sec:character_seanet} treat time series classification as a \emph{Multiple Instance Learning} (MIL) problem \citep{dietterich1997mil, early2024millet}: the series $X_i$ is the bag, each time point $x_i^j$ is an instance, and only the bag label $Y_i$ is observed. MIL has one main role in \seanet{}. The instance labels $y_i^j$ are never annotated, yet the model must still say which time points carry the evidence for its prediction. MIL recovers this from the bag label alone: the pooling head first gives every time point its own class evidence and only then aggregates it into one prediction, so the same per-time-point evidence that forms the prediction is also the interpretation $\hat{I}_i \in \mathbb{R}^{t \times C}$. The interpretation is therefore part of the forward pass and not a post-hoc approximation.

\begin{figure}[!htbp]
	\centering
	\includegraphics[width=0.88\linewidth]{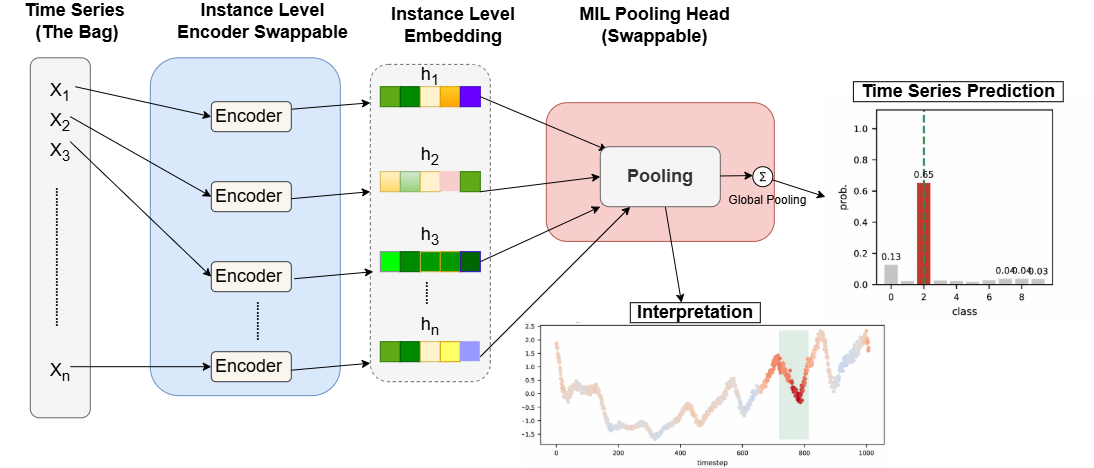}
	\caption{The MIL view of a time series, read from left to right. The series $X_i$ is the bag and each time point is one instance. The encoder gives one embedding $z_i^j$ per instance; the pooling head weights those embeddings and combines them into the prediction $\hat{Y}_i$ (right). The same weighted contributions, kept instead of discarded, are the interpretation $\hat{I}_i$ drawn on the series below, where red pushes towards the predicted class and blue pushes away from it. The two grey boxes are interchangeable, which is what lets any encoder of \Cref{app:encoders} be paired with any head of \Cref{app:heads}.}
	\label{fig:mil_view}
\end{figure}

The main text states this view in words. This appendix adds two details: \Cref{fig:mil_view} draws the full pipeline from the bag to the prediction and the interpretation, and \Cref{eq:GAP_explanation} writes out the embedded pooling that this view replaces, so that the order of its two steps can be compared with the order of our heads.

The alternative is worth writing down once, because the rest of the design follows from it. \emph{Embedded pooling} \citep{wang2018revisiting} first averages the embeddings and then classifies the average:
\begin{equation}
	P_i = \frac{1}{t}\sum_{j=1}^{t} z_i^j \in \mathbb{R}^{d},
	\qquad
	\hat{Y}_i = \psi_{clf}(P_i) .
	\label{eq:GAP_explanation}
\end{equation}
Once $P_i$ is formed, the time index is gone, so there is no per-time-point prediction left to report. Every head of \Cref{app:heads}, starting with the conjunctive pooling of \millet{} (\Cref{eq:conj_pool}), runs the two steps in the opposite order: it classifies each time point first and aggregates afterwards, which is why it can report the interpretation.

% =====================================================================
\section{How the Encoder Stays Small}
\label{app:encoders}

\Cref{sec:method:enc} describes the \seanet{} encoder: a stem convolution followed by a stack of multi-scale separable residual blocks (\Cref{eq:unit} and \Cref{eq:block}), which keeps the length $t$ unchanged, so that one embedding per time point reaches the pooling head, as the MIL view of \Cref{app:milview} requires. It also gives four variants of this encoder: the bottleneck pointwise mixing (\Cref{eq:bottleneck}), the two channel re-weighting variants, \emph{Gated (mean)} and \emph{Input Gated} (\Cref{eq:generic_gate}), and the Spike/Trend encoder (\Cref{eq:spike}). Every variant returns the same output $Z_i \in \mathbb{R}^{t \times d}$, so any encoder can be paired with any pooling head.

This appendix gives the details that the main text leaves out. \Cref{app:block} draws one block and the effect of the dilation, \Cref{app:bottleneck} works out where the saving of the bottleneck comes from, and \Cref{app:gates} and \Cref{app:spiketrend} write out the gated and the Spike/Trend variants step by step. \Cref{app:recon} to \Cref{app:pyramid} then describe one more encoder and two wrappers that the main text does not cover, and \Cref{app:encoder_tables} lists the two main encoders layer by layer.

% ---------------------------------------------------------------------
\subsection{Multi-scale Separable Blocks}
\label{app:block}

\begin{figure}[!htbp]
	\centering
% The data path of one block, panel (a) of
% figures/selected_figures/01_main/fig04_separable_block/fig04_block_full_tikz.tex.
% Only that panel is used here; panel (b) of that file is the dilation
% drawing, which is its own figure below.
% ---- panel (a): the data path ---------------------------------------
\resizebox{\linewidth}{!}{%
\begin{tikzpicture}[
  font=\small,
  >={Stealth[length=2mm]},
  % a convolution branch: the three of them differ only in kernel size
  conv/.style   = {draw=orange!65!black, fill=orange!18, rounded corners=2pt,
                   minimum height=8mm, inner xsep=4pt, align=center},
  % the cheap "mix and clean up" steps after the sum
  stage/.style   = {draw=black!55, fill=black!7, rounded corners=2pt,
                   minimum height=8mm, minimum width=11mm, align=center},
  % the tensors going in and out
  tensor/.style = {draw=black!45, fill=black!3, rounded corners=2pt,
                   minimum height=11mm, minimum width=13mm, align=center},
  op/.style     = {draw=black!65, circle, inner sep=1pt, minimum size=5.5mm},
]

% --- input ------------------------------------------------------------
\node[tensor] (h) at (0,0) {$\mathbf{h}$\\[1pt]\scriptsize$(B,d,\nsteps)$};

% --- the three depthwise convolutions, one per kernel size ------------
% drawn with increasing width, so "wider box = looks further in time"
\node[conv, minimum width=20mm] (k5)  at (3.7, 1.7) {depthwise $k{=}5$\\[-1pt]\scriptsize dilation $\delta$, groups $=d$};
\node[conv, minimum width=26mm] (k11) at (3.7, 0.0) {depthwise $k{=}11$\\[-1pt]\scriptsize dilation $\delta$, groups $=d$};
\node[conv, minimum width=32mm] (k23) at (3.7,-1.7) {depthwise $k{=}23$\\[-1pt]\scriptsize dilation $\delta$, groups $=d$};

\draw[->] (h.east) -- ++(0.4,0) |- (k5.west);
\draw[->] (h.east) -- ++(0.4,0) |- (k11.west);
\draw[->] (h.east) -- ++(0.4,0) |- (k23.west);

% --- they are simply added together -----------------------------------
\node[op] (sum) at (6.9,0) {$+$};
\draw[->] (k5.east)  -| (sum.north);
\draw[->] (k11.east) -- (sum.west);
\draw[->] (k23.east) -| (sum.south);

% --- mix the channels, normalise, activate, regularise ----------------
\node[stage] (pw)   at ( 8.9,0) {$1{\times}1$\\[-1pt]pointwise};
\node[stage] (bn)   at (10.7,0) {BatchNorm};
\node[stage] (relu) at (12.4,0) {ReLU};
\node[stage] (drop) at (14.0,0) {Dropout};

\draw[->] (sum.east)  -- (pw.west);
\draw[->] (pw.east)   -- (bn.west);
\draw[->] (bn.east)   -- (relu.west);
\draw[->] (relu.east) -- (drop.west);

% --- the residual add that closes the block ---------------------------
\node[op] (add) at (16.0,0) {$+$};
\node[stage, minimum width=9mm] (outrelu) at (17.6,0) {ReLU};
\node[tensor] (out) at (19.6,0) {output\\[1pt]\scriptsize$(B,d,\nsteps)$};

\draw[->] (drop.east)     -- (add.west);
\draw[->] (add.east)      -- (outrelu.west);
\draw[->] (outrelu.east) -- (out.west);

% the skip connection: the BLOCK's own input, carried over the top
\draw[->, thick, black!60] (h.north) |- (0,3.0) -| node[pos=0.25, above, black!60]
      {\small skip connection: the block's own input $\mathbf{h}$} (add.north);

% --- the dashed box marking what gets repeated twice ------------------
\begin{scope}[on background layer]
  \node[draw=black!55, dashed, rounded corners=3pt, inner sep=7pt,
        fit=(k5)(k23)(drop)] (unitbox) {};
\end{scope}
% the label sits UNDER the dashed box, where nothing else is drawn
\node[anchor=north west, black!70] at ($(unitbox.south west)+(2pt,-1pt)$)
      {\small \textbf{one unit} -- run twice, back to back ($\times 2$)};

\end{tikzpicture}}
	\caption{One multi-scale separable block. The three depth wise convolutions read the same input at the kernel sizes $5$, $11$ and $23$, their outputs are summed, and the point-wise $1\times1$ convolution $P$ mixes the channels; BatchNorm, ReLU and dropout close the unit $\mathcal{U}$ of \Cref{eq:unit}. The dashed box is what is applied twice, and the arrow over the top is the residual connection of \Cref{eq:block}.}
	\label{fig:arch_block}
\end{figure}
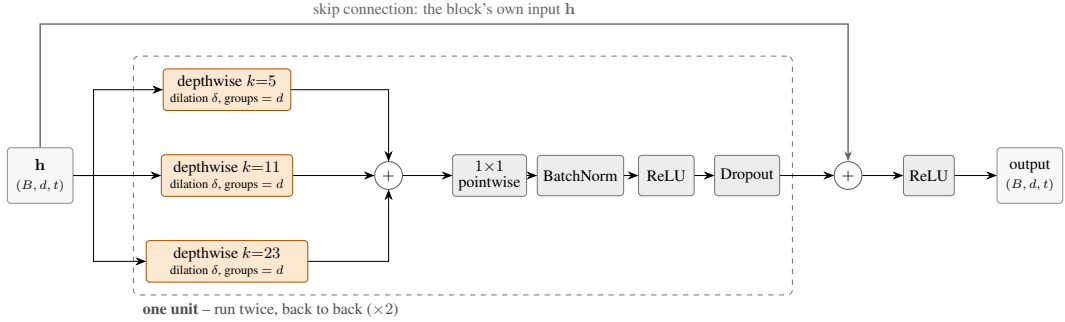

\Cref{fig:arch_block} draws one block of \Cref{sec:arch:block} as a data path, i.e. \Cref{eq:unit} and \Cref{eq:block}. Two terms of \Cref{eq:unit} need a short explanation. In a depthwise convolution $D_k^{(\delta)}$, each input channel is filtered with its own kernel, and the channels are not mixed. It is a grouped convolution with as many groups as channels, so it needs $d$ times fewer parameters and FLOPs than a full convolution; the same idea is the basis of efficient architectures such as those of \citet{howard2017mobilenets} and \citet{xie2017aggregated}. The pointwise $1\times1$ convolution $P$ is therefore the only place where the channels are mixed, and it also holds most of the parameters; this is why the bottleneck of \Cref{eq:bottleneck} factorizes $P$ and nothing else. Batch normalization \citep{ioffe2015batchnorm} stabilizes the activations and dropout regularizes the model; neither of them changes the shape.

\begin{figure}[!htbp]
	\centering
	\begin{subfigure}[t]{\linewidth}
		\centering
		% !TEX root = ../../../../internsip_ubaid.tex

% Figure 4.1 -- the k=5 filter at dilation 1: five taps read five neighbouring time steps.
% Split out of panel (b) of fig_block_tikz.tex so it can be placed on its own.
\resizebox{0.92\linewidth}{!}{%
\begin{tikzpicture}[x=0.185cm, y=1cm, font=\small, >={Stealth[length=1.6mm]}]
  % 65 input positions, the filter takes 5 neighbouring points
  \foreach \i in {0,...,64} { \fill[black!22] (\i,0) circle (0.5mm); }
  \foreach \i in {30,31,32,33,34} {
    \draw[orange!65!black, thin] (\i,0) -- (32,0.85);
    \fill[orange!75!black] (\i,0) circle (0.9mm);
  }
  \fill[black!80] (32,0.85) circle (1.1mm);
  \node[anchor=west, black!80] at (34,0.85) {\ one output value};
  \node[anchor=east] at (-1.5,0) {$\delta = 1$\ };
  \node[anchor=north, black!60] at (32,-0.30) {\small sees 5 time steps in a row};
\end{tikzpicture}}
		\caption{$\delta = 1$: the five taps read five neighbouring time points.}
		\label{fig:dilation_d1}
	\end{subfigure}

	\begin{subfigure}[t]{\linewidth}
		\centering
		% !TEX root = ../../../../internsip_ubaid.tex

% Figure 4.2 -- the SAME five taps at dilation 16: they now span 65 time steps.
% Split out of panel (b) of fig_block_tikz.tex so it can be placed on its own.
\resizebox{0.92\linewidth}{!}{%
\begin{tikzpicture}[x=0.185cm, y=1cm, font=\small, >={Stealth[length=1.6mm]}]
  \foreach \i in {0,...,64} { \fill[black!22] (\i,0) circle (0.5mm); }
  \foreach \i in {0,16,32,48,64} {
    \draw[orange!65!black, thin] (\i,0) -- (32,0.85);
    \fill[orange!75!black] (\i,0) circle (0.9mm);
  }
  \fill[black!80] (32,0.85) circle (1.1mm);
  \node[anchor=west, black!80] at (34,0.85) {\ one output value};
  \node[anchor=east] at (-1.5,0) {$\delta = 16$\ };
  \node[anchor=north, black!60] at (32,-0.30)
       {\small the same 5 taps, now spread over 65 time steps};
  % the reach, drawn as a measured bar
  \draw[<->, black!55] (0,-0.95) -- node[below, black!55, font=\small] {receptive field $=65$} (64,-0.95);
\end{tikzpicture}}
		\caption{$\delta = 16$: the same five taps span $65$ time points.}
		\label{fig:dilation_d16}
	\end{subfigure}
	\caption{The effect of the dilation, drawn with the same five taps of the $k = 5$ filter in both rows. Widening the region that the filter sees costs no extra weight, which is why the receptive field of the encoder can be grown without paying for it in parameters.}
	\label{fig:dilation}
\end{figure}
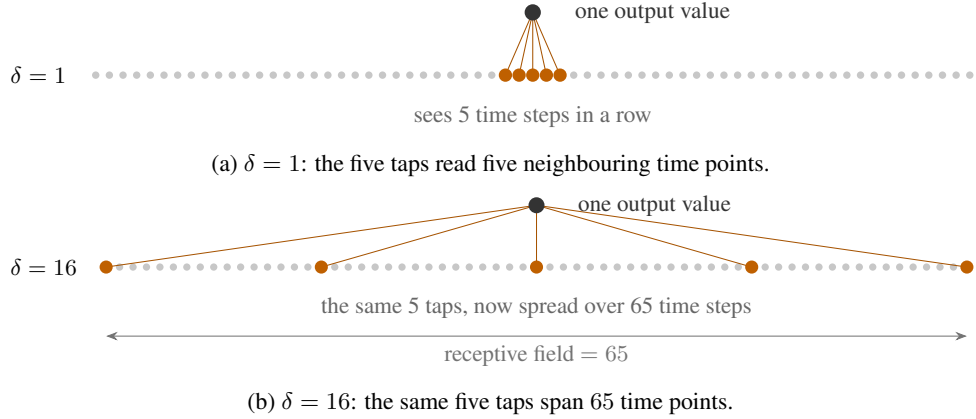

\Cref{fig:dilation} shows what the dilation of \Cref{sec:arch:block} gives. A kernel of size $5$ always holds five weights. At $\delta = 1$ they read five neighbouring time points; at $\delta = 16$ the same five weights read time points that are $16$ apart and together cover $65$. The number of parameters does not change, so the receptive field grows at no cost in weights, and the cap $\delta_{\max} = 16$ keeps it bounded, so that each embedding still describes a local region of the series.

% ---------------------------------------------------------------------
\subsection{Where the Bottleneck Saving Comes From}
\label{app:bottleneck}

\Cref{eq:bottleneck} gives the ratio $2r/d$; here is the arithmetic behind the $28\,\%$ saving. At $d = 64$ the pointwise step costs $d^2 = 4{,}096$ weights per unit, and with $r = d/4 = 16$ it costs $2dr = 2{,}048$, so each unit saves $2{,}048$ weights. A block holds two units and the model holds four blocks, which gives $8 \times 2{,}048 = 16{,}384$; the bias terms of the factorized layers bring the measured saving to $16{,}256$, i.e. from $57{,}580$ to $41{,}324$ parameters (\Cref{tab:encoder_bottleneck,tab:encoder_base}). The depthwise filters are left unchanged, because with the kernel sizes $\{5,11,23\}$ they hold only $39d$ weights per unit and are not where the cost is.

% ---------------------------------------------------------------------
\subsection{The Two Gated Variants, Written Out}
\label{app:gates}

\Cref{eq:generic_gate} writes both channel re-weighting variants in one line. Written separately, \emph{Gated (mean)} first averages the embeddings over time and computes one weight vector from that summary,
\begin{equation}
	s_i = \frac{1}{t}\sum_{j=1}^{t} z_i^j \in \mathbb{R}^{d},
	\qquad
	g_i = \sigma(W_g\, s_i) \in [0,1]^{d},
	\qquad
	W_g \in \mathbb{R}^{d \times d},
	\label{eq:app:gate_mean}
\end{equation}
and the same $g_i$ is then applied at every time point, i.e. $g_i^{\,j} = g_i$ for all $j$. The mean is one of three ways to summarize the time axis; the other two are the maximum and the last time point. The mean is used in the main experiments, and the other two appear in \Cref{tab:leaderboard_sweep}. The cost is one $d \times d$ matrix.

\emph{Input Gated} keeps the index $j$: the gate $g_i^{\,j}$ is computed from the raw series by a convolution with kernel size $7$ and ``same'' padding, so the gate has exactly $t$ entries, one per time point. Its cost is $7d$ weights, i.e. less than the mean gate, and its gradient reaches the input directly instead of passing through the encoder blocks.

% ---------------------------------------------------------------------
\subsection{\seanet{} Spike/Trend Encoder}
\label{app:spiketrend}

\Cref{eq:spike} writes the spike branch as one expression. In the implementation it is four steps in this order: a moving average with window $w = 5$, its subtraction from the input, and then $\operatorname{Conv}_3$, BatchNorm and ReLU, all with ``same'' padding, so the length $t$ is kept. The mixed features are gated by \Cref{eq:app:gate_mean} before they reach the head. We expected an advantage on datasets where the class evidence is a sharp spike, but \Cref{tab:ablation_encoder} shows that this was the weakest of the five encoders, so the expectation was not met.

% ---------------------------------------------------------------------
\subsection{\seanet{} Reconstruction-residual Encoder}
\label{app:recon}

The spike branch decides in advance what the encoder misses, namely the sharp changes. This variant learns it instead, by asking which part of the input the encoder features cannot rebuild. A $1\times1$ convolution reconstructs the raw series from the features, the reconstruction is subtracted from the real input, and the remainder becomes $d_{\text{res}} = 16$ extra channels:
\begin{equation}
	\widehat{X}_i = P_{\text{rec}}(Z_i),
	\qquad
	\tilde{z}_i^{\,j} = P\big([\,z_i^j \,;\, \phi(X_i - \widehat{X}_i)^j\,]\big),
	\label{eq:app:recon}
\end{equation}
where $\phi$ is the same small convolution, BatchNorm and ReLU used by the spike branch. There is no separate reconstruction loss: the branch is trained end to end with the classifier, so this is a learned residual feature and not an auto-encoder. It was the second strongest encoder in \Cref{tab:ablation_encoder}.

% ---------------------------------------------------------------------
\subsection{\seanet{} Multi-scale Channels (wrapper)}
\label{app:mschan}

A \emph{wrapper} changes the input or the output of an encoder but leaves the encoder itself unchanged. The stem receives a single raw channel, so any local statistic that it needs must be learned from scratch. This wrapper adds the rolling maximum, minimum and standard deviation over several windows, together with the sign of the first difference, as extra input channels:
\begin{equation}
	\begin{aligned}
	X_i^{+} = \Big[\, X_i \,;\; & \big(\text{roll\_max}_w(X_i)\big)_{w \in \mathcal{W}} \;;\;
	\big(\text{roll\_min}_w(X_i)\big)_{w \in \mathcal{W}} \;;\; \\
	& \big(\text{roll\_std}_w(X_i)\big)_{w \in \mathcal{W}} \;;\; \operatorname{sign}(\Delta X_i) \,\Big],
	\end{aligned}
	\label{eq:app:mschan}
\end{equation}
with $\mathcal{W} = \{3, 7, 15, 31\}$, all odd so that each window stays centred. The input grows from $1$ to $1 + 4\cdot3 + 1 = 14$ channels before the stem, and the rest is one of the encoders above. A rolling mean is not used on purpose: it is linear, so the stem convolution can learn it on its own, whereas the other three statistics are not linear. A BatchNorm first normalizes the $14$ channels, because they do not share the scale of the raw series; removing it reduced the accuracy.

% ---------------------------------------------------------------------
\subsection{\seanet{} Multi-scale Pyramid (wrapper)}
\label{app:pyramid}

The dilation cap bounds the receptive field, so evidence that is visible only over a long span cannot be seen. This wrapper runs the same encoder on the series downsampled by $1$, $2$, $4$ and $8$, stretches the outputs back to length $t$, and combines them with an attention over the scales:
\begin{equation}
	\tilde{z}_i^{\,j} = \sum_{s \in \mathcal{S}} \alpha_i^{\,j,s}\, z_i^{\,j,(s)},
	\qquad
	\alpha_i^{\,j,s} = \operatorname{softmax}_{s}\big(\text{score}(z_i^{\,j,(s)})\big),
	\qquad \mathcal{S} = \{1, 2, 4, 8\} .
	\label{eq:app:pyramid}
\end{equation}
The softmax normalizes over the \emph{scales} and never over time, so the per-time-point signal on which the two interpretation metrics depend is kept. One set of encoder weights is shared by all scales, so the number of parameters barely changes and the extra cost is compute, about $1.9\times$. This is the one-dimensional analogue of the multi-magnification MIL of \citet{hashimoto2020msdamil}, and it was the weakest variant we built (\Cref{tab:ablation_encoder}). It also has a constraint: the wrapped encoder must be shallow enough that its receptive field does not exceed $t / \max(\mathcal{S})$, otherwise the coarsest scales collapse to a nearly constant signal. The pyramid therefore uses a two-block base, and the pipeline checks this automatically.

\paragraph{Alternatives we rejected.}
A Transformer encoder costs $\mathcal{O}(t^2)$ in time and memory, which is too expensive for series of $1{,}000$--$1{,}500$ points on the target hardware. Its unrestricted attention also lets any output position read any input position, which removes the bounded receptive field that the interpretation relies on. Pruning or quantizing \texttt{InceptionTime} would reduce its size, but would not change where the cost is, i.e. in the encoder, which is the point of this paper.

% ---------------------------------------------------------------------
\subsection{The Two Encoders, Layer by Layer}
\label{app:encoder_tables}

\Cref{tab:encoder_bottleneck} and \Cref{tab:encoder_base} give one row per operation, with its output shape and its number of parameters. Read side by side, they show that the saving of \Cref{app:bottleneck} comes only from the pointwise layers.

\begin{table}[!htbp]
\caption{The \seanet{} bottleneck encoder ($d = 64$, 4 blocks, kernels 5/11/23, dilations 1/2/4/8), layer by layer, with the output shape and the number of parameters, built from \texttt{top/top\_bottleneck\_topk} for \webtraffic{} ($c = 10$). The point-wise convolution of every unit is factorised as $d \to d/4 \to d$, which is where the parameter saving comes from. Every layer keeps the series length $t$, so the head sees one embedding per time point. Depthwise = one filter per channel (groups $= d$).}
\label{tab:encoder_bottleneck}
\centering
\seatab
\begin{tabular}{L{2.55cm}L{5.35cm}C{2.75cm}R{1.75cm}}
\toprule
\textbf{Stage} & \textbf{Layer} & \textbf{Output} & \textbf{Params} \\
\midrule
Input & z-normalised series & $b \times 1 \times t$ & -- \\
Stem & Conv1d $1 \to 64$, kernel 7 & $b \times 64 \times t$ & 512 \\
\midrule
Block 1 (dilation 1), $2\times$ unit: & depthwise Conv1d $\times 3$ (kernels 5, 11, 23), summed & $b \times 64 \times t$ & 2{,}688 \\
 & pointwise Conv1d $64 \to 16$ (reduce) & $b \times 16 \times t$ & 1{,}040 \\
 & pointwise Conv1d $16 \to 64$ (expand) & $b \times 64 \times t$ & 1{,}088 \\
 & BatchNorm $+$ ReLU $+$ dropout & $b \times 64 \times t$ & 128 \\
 & \emph{after both units:} residual add $+$ ReLU & $b \times 64 \times t$ & \emph{block total 9{,}888} \\
\midrule
Block 2 (dilation 2), $2\times$ unit: & depthwise Conv1d $\times 3$ (kernels 5, 11, 23), summed & $b \times 64 \times t$ & 2{,}688 \\
 & pointwise Conv1d $64 \to 16$ (reduce) & $b \times 16 \times t$ & 1{,}040 \\
 & pointwise Conv1d $16 \to 64$ (expand) & $b \times 64 \times t$ & 1{,}088 \\
 & BatchNorm $+$ ReLU $+$ dropout & $b \times 64 \times t$ & 128 \\
 & \emph{after both units:} residual add $+$ ReLU & $b \times 64 \times t$ & \emph{block total 9{,}888} \\
\midrule
Block 3 (dilation 4), $2\times$ unit: & depthwise Conv1d $\times 3$ (kernels 5, 11, 23), summed & $b \times 64 \times t$ & 2{,}688 \\
 & pointwise Conv1d $64 \to 16$ (reduce) & $b \times 16 \times t$ & 1{,}040 \\
 & pointwise Conv1d $16 \to 64$ (expand) & $b \times 64 \times t$ & 1{,}088 \\
 & BatchNorm $+$ ReLU $+$ dropout & $b \times 64 \times t$ & 128 \\
 & \emph{after both units:} residual add $+$ ReLU & $b \times 64 \times t$ & \emph{block total 9{,}888} \\
\midrule
Block 4 (dilation 8), $2\times$ unit: & depthwise Conv1d $\times 3$ (kernels 5, 11, 23), summed & $b \times 64 \times t$ & 2{,}688 \\
 & pointwise Conv1d $64 \to 16$ (reduce) & $b \times 16 \times t$ & 1{,}040 \\
 & pointwise Conv1d $16 \to 64$ (expand) & $b \times 64 \times t$ & 1{,}088 \\
 & BatchNorm $+$ ReLU $+$ dropout & $b \times 64 \times t$ & 128 \\
 & \emph{after both units:} residual add $+$ ReLU & $b \times 64 \times t$ & \emph{block total 9{,}888} \\
\midrule
\textbf{Encoder total} &  & $b \times 64 \times t$ (time-point embs.) & 40{,}064 \\
Pooling head & Top-$k$ conjunctive ($a = 8$, $c = 10$) & $b \times 10$ & 1{,}260 \\
\textbf{Model total} &  &  & 41{,}324 \\
\bottomrule
\end{tabular}
\end{table}

\begin{table}[!htbp]
\caption{The \seanet{} base encoder ($d = 64$, 4 blocks, kernels 5/11/23, dilations 1/2/4/8), layer by layer, with the output shape and the number of parameters, built from \texttt{seanet/seanet\_slim\_topk} for \webtraffic{} ($c = 10$). It has the same skeleton with the plain $d \to d$ point-wise convolution, for comparison. Every layer keeps the series length $t$, so the head sees one embedding per time point. Depthwise = one filter per channel (groups $= d$).}
\label{tab:encoder_base}
\centering
\seatab
\begin{tabular}{L{2.55cm}L{5.35cm}C{2.75cm}R{1.75cm}}
\toprule
\textbf{Stage} & \textbf{Layer} & \textbf{Output} & \textbf{Params} \\
\midrule
Input & z-normalised series & $b \times 1 \times t$ & -- \\
Stem & Conv1d $1 \to 64$, kernel 7 & $b \times 64 \times t$ & 512 \\
\midrule
Block 1 (dilation 1), $2\times$ unit: & depthwise Conv1d $\times 3$ (kernels 5, 11, 23), summed & $b \times 64 \times t$ & 2{,}688 \\
 & pointwise Conv1d $64 \to 64$ & $b \times 64 \times t$ & 4{,}160 \\
 & BatchNorm $+$ ReLU $+$ dropout & $b \times 64 \times t$ & 128 \\
 & \emph{after both units:} residual add $+$ ReLU & $b \times 64 \times t$ & \emph{block total 13{,}952} \\
\midrule
Block 2 (dilation 2), $2\times$ unit: & depthwise Conv1d $\times 3$ (kernels 5, 11, 23), summed & $b \times 64 \times t$ & 2{,}688 \\
 & pointwise Conv1d $64 \to 64$ & $b \times 64 \times t$ & 4{,}160 \\
 & BatchNorm $+$ ReLU $+$ dropout & $b \times 64 \times t$ & 128 \\
 & \emph{after both units:} residual add $+$ ReLU & $b \times 64 \times t$ & \emph{block total 13{,}952} \\
\midrule
Block 3 (dilation 4), $2\times$ unit: & depthwise Conv1d $\times 3$ (kernels 5, 11, 23), summed & $b \times 64 \times t$ & 2{,}688 \\
 & pointwise Conv1d $64 \to 64$ & $b \times 64 \times t$ & 4{,}160 \\
 & BatchNorm $+$ ReLU $+$ dropout & $b \times 64 \times t$ & 128 \\
 & \emph{after both units:} residual add $+$ ReLU & $b \times 64 \times t$ & \emph{block total 13{,}952} \\
\midrule
Block 4 (dilation 8), $2\times$ unit: & depthwise Conv1d $\times 3$ (kernels 5, 11, 23), summed & $b \times 64 \times t$ & 2{,}688 \\
 & pointwise Conv1d $64 \to 64$ & $b \times 64 \times t$ & 4{,}160 \\
 & BatchNorm $+$ ReLU $+$ dropout & $b \times 64 \times t$ & 128 \\
 & \emph{after both units:} residual add $+$ ReLU & $b \times 64 \times t$ & \emph{block total 13{,}952} \\
\midrule
\textbf{Encoder total} &  & $b \times 64 \times t$ (time-point embs.) & 56{,}320 \\
Pooling head & Top-$k$ conjunctive ($a = 8$, $c = 10$) & $b \times 10$ & 1{,}260 \\
\textbf{Model total} &  &  & 57{,}580 \\
\bottomrule
\end{tabular}
\end{table}

% =====================================================================
\section{Pooling Heads Compared}
\label{app:heads}

\Cref{sec:method:head} describes the MIL pooling head $\psi_{pool}$. It has two branches: an attention branch, which decides how much each time point counts, and a classifier branch, which decides which class each time point argues for. Their product is the evidence $e_i^{\,j,k}$ of time point $j$ for class $k$; aggregating it over time gives the prediction $\hat{Y}_i$, and keeping it per time point gives the interpretation $\hat{I}_i$. The main text starts from the conjunctive pooling of \millet{} (\Cref{eq:conj_pool}), states its three limitations, and gives the three heads that remove them: class-wise (\Cref{eq:classwise_pool}), Soft-Max (\Cref{eq:softmax_pool}) and Top-$k$ (\Cref{eq:topk_pool}) conjunctive pooling.

This appendix adds three things. First, it draws every head in the same way, one full-width figure per head (\Cref{fig:pool_conj} to \Cref{fig:pool_dualstream}). All seven of our heads keep the classifier branch $\hat{y}_i^{\,j} = W_c z_i^j + \mathbf{b}_c$ unchanged and differ from the baseline only in two places, the attention head and the aggregation over time; these two places are drawn in blue, so two heads can be compared by looking at the blue boxes only. Second, \Cref{app:pool:adaptive} to \Cref{app:pool:dualstream} describe the four heads that the main text does not cover, with their equations and the reason why each was built. Third, \Cref{app:heads_tables} summarizes the eight heads and gives each of them layer by layer.
% Old sentence, for the two 2 x 2 grids (now commented out below):
% Every panel of \Cref{fig:pool_group1} and \Cref{fig:pool_group2} is drawn the same way, so two heads can be compared by looking at the two slots that differ: \Cref{fig:pool_group1} sets the baseline next to the three heads of the main text, and \Cref{fig:pool_group2} gathers the four heads that the main text only names.

% ---- small drawing helpers -------------------------------------------
% a cube with its bottom-left front corner at (#1,#2); 0.42 wide, 0.14 deep
\newcommand{\pcube}[2]{%
  \draw[fill=black!6,  draw=black!70] (#1,#2) rectangle ++(0.42,0.42);
  \draw[fill=black!18, draw=black!70] (#1,#2+0.42) -- ++(0.14,0.14) -- ++(0.42,0) -- ++(-0.14,-0.14) -- cycle;
  \draw[fill=black!30, draw=black!70] (#1+0.42,#2) -- ++(0.14,0.14) -- ++(0,0.42) -- ++(-0.14,-0.14) -- cycle;}
% one small square (a scalar per time step) with bottom-left at (#1,#2)
\newcommand{\psq}[2]{\draw[fill=white, draw=black!70] (#1,#2) rectangle ++(0.3,0.3);}
% a row "square square ... square", 1.7 wide, left edge at #1, bottom at #2
\newcommand{\psqrow}[2]{%
  \psq{#1}{#2}\psq{#1+0.4}{#2}
  \node at (#1+1.05,#2+0.15) {$\cdots$};
  \psq{#1+1.4}{#2}}

\newcommand{\poolfigwide}[5]{%
\begin{tikzpicture}[
  font=\small,
  % lines are set at their natural width (no text width), so the spaces in
  % a formula are never stretched; the boxes get a fixed minimum width so
  % that every head is drawn on the same grid
  box/.style  = {draw=black!55, fill=black!3, rounded corners=2pt, align=center, inner sep=4pt},
  slot/.style = {draw=blue!60!black, fill=blue!4, rounded corners=2pt, align=center,
                 inner sep=4pt, line width=0.7pt},
  arr/.style  = {->, >={Stealth[length=2mm]}, black!70, line width=0.6pt},
  op/.style   = {draw=black!65, circle, fill=white, inner sep=0.5pt, minimum size=5mm},
]
  % column 2: attention head (changes) above the classifier (fixed)
  \node[slot, minimum width=5.6cm] (att) at (0,0)
    {{\footnotesize\bfseries\color{blue!60!black} attention head}\\[2pt] #2};
  \node[box, minimum width=5.6cm, below=5mm of att] (clf)
    {{\footnotesize\bfseries classifier}\ {\footnotesize(the same in every head)}\\[2pt]
     $\hat{y}_i^{\,j} = W_c z_i^j + \mathbf{b}_c \in \R^{C}$};
  \coordinate (mid) at ($(att.south)!0.5!(clf.north)$);

  % column 1: the bag of time point embeddings, centred on the two branches;
  % the empty first line leaves room for the row of cubes drawn on top
  \node[box, minimum width=2.7cm, anchor=east] (emb) at ($(mid -| att.west)+(-9mm,0)$)
    {\rule{0pt}{7mm}\\
     {\footnotesize\bfseries time point}\\[-1pt] {\footnotesize\bfseries embeddings}\\[2pt]
     $z_i^j \in \R^{d}$\\ $j = 1,\dots,t$};
  \begin{scope}[shift={($(emb.north)+(-1.1,-0.78)$)}, scale=0.8]
    \pcube{0}{0}\pcube{0.72}{0}\node at (1.6,0.25) {$\cdots$};\pcube{2.2}{0}
  \end{scope}
  \draw[arr] (emb.east) -- ++(4.5mm,0) |- (att.west);
  \draw[arr] (emb.east) -- ++(4.5mm,0) |- (clf.west);

  % column 3: attention weighting, i.e. the product of the two branches
  \node[op] (mul) at ($(mid -| att.east)+(6mm,0)$) {$\times$};
  \draw[arr] (att.east) -| (mul.north);
  \draw[arr] (clf.east) -| (mul.south);
  \node[box, minimum width=3.3cm, anchor=west] (ev) at ($(mul.east)+(3.5mm,0)$)
    {{\footnotesize\bfseries evidence at}\\[-1pt] {\footnotesize\bfseries each time point}\\[2pt] #3};
  \draw[arr] (mul.east) -- (ev.west);

  % row 2: aggregation over time, then the time series prediction
  \coordinate (low) at (current bounding box.south);
  \path (att.west |- low) ++(0,-6mm) coordinate (aggnw);
  \path (ev.east |- low) coordinate (agge);
  \node[slot, anchor=north west, minimum width=10.1cm] (agg) at (aggnw)
    {{\footnotesize\bfseries\color{blue!60!black} aggregation over time: #4}\\[3pt] #5};
  \draw[arr] (ev.south) -- (ev.south |- agg.north);
  \node[box, minimum width=2.7cm, anchor=east] (out) at ($(agg.west)+(-9mm,0)$)
    {{\footnotesize\bfseries time series}\\[-1pt] {\footnotesize\bfseries prediction}\\[2pt] $\hat{Y}_i \in \R^{C}$};
  \draw[arr] (agg.west) -- (out.east);

  % title
  \node[anchor=south west, font=\small\bfseries, inner sep=0pt]
    at ($(current bounding box.north west)+(0,2mm)$) {#1};
\end{tikzpicture}}

% ---------------------------------------------------------------------
\subsection{Conjunctive Pooling}
\label{app:conjunctive}
\label{sec:method:mil}

% \poolgridone   % old 2 x 2 grid, switched off above

\begin{figure}[!htbp]
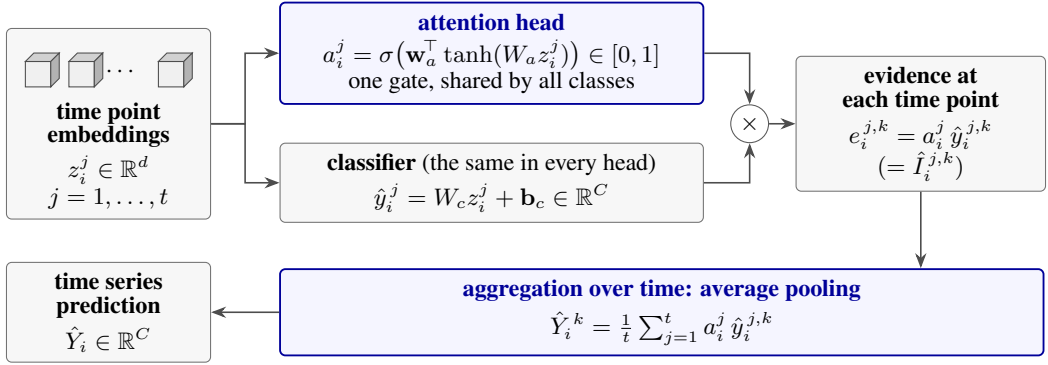

	\centering
	\poolfigwide{Conjunctive pooling (\millet{} baseline)}
		{$a_i^j = \sigma\big(\mathbf{w}_a^{\!\top}\tanh(W_a z_i^j)\big) \in [0,1]$\\ one gate, shared by all classes}
		{$e_i^{\,j,k} = a_i^j\,\hat{y}_i^{\,j,k}$\\ $(= \hat{I}_i^{\,j,k})$}
		{average pooling}
		{$\hat{Y}_i^{\,k} = \frac{1}{t}\sum_{j=1}^{t} a_i^j\,\hat{y}_i^{\,j,k}$}
	\caption{Conjunctive pooling, the \millet{} baseline of \Cref{eq:conj_pool}. Read from left to right: the time point embeddings feed an attention head and a classifier, their product is the evidence at each time point, which is also the interpretation, and aggregating that evidence over time gives the prediction for the series. The classifier is the same in every head; only the two blue boxes, the attention head and the aggregation, change from one head to the next, so \Cref{fig:pool_classwise} to \Cref{fig:pool_dualstream} are best read against this figure.}
	\label{fig:pool_conj}
\end{figure}

\Cref{fig:pool_conj} draws the baseline of \Cref{eq:conj_pool}. The head itself is almost free: replacing it by global average pooling on the same \texttt{InceptionTime} backbone removes only $1{,}041$ of the $423{,}707$ parameters, so it is the encoder, and not the interpretability, that makes the baseline large. \Cref{sec:method:head} states the three limitations of this head. We call them \textbf{L1} (one gate is shared by all classes), \textbf{L2} (the weights are not normalized over time) and \textbf{L3} (the fixed mean dilutes a few decisive time points in a long series), and use these names below and in \Cref{tab:heads_summary}.

% ---------------------------------------------------------------------
\subsection{Class-wise Conjunctive Pooling}
\label{app:pool:classwise}

\begin{figure}[!htbp]
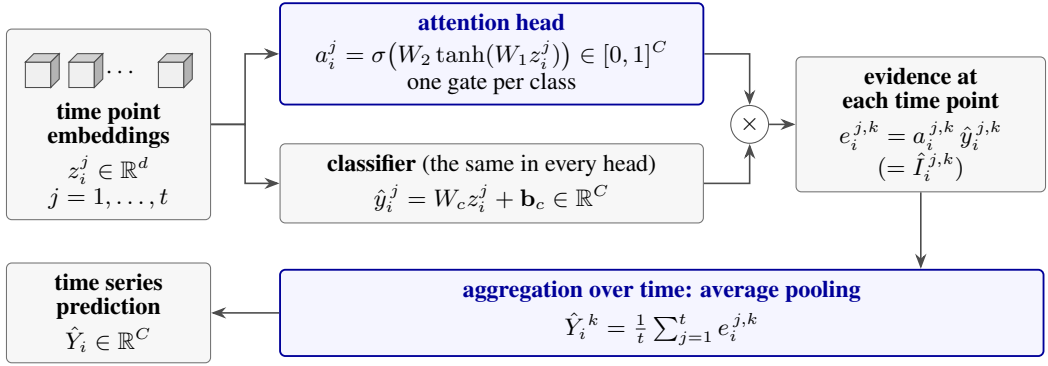

	\centering
	\poolfigwide{Class-wise conjunctive pooling}
		{$a_i^j = \sigma\big(W_2\tanh(W_1 z_i^j)\big) \in [0,1]^{C}$\\ one gate per class}
		{$e_i^{\,j,k} = a_i^{\,j,k}\,\hat{y}_i^{\,j,k}$\\ $(= \hat{I}_i^{\,j,k})$}
		{average pooling}
		{$\hat{Y}_i^{\,k} = \frac{1}{t}\sum_{j=1}^{t} e_i^{\,j,k}$}
	\caption{Class-wise conjunctive pooling. Against \Cref{fig:pool_conj}, only the attention head changes: widening its last layer to $C$ outputs costs $(C-1)(d_a+1)$ weights and makes the evidence, and therefore the interpretation, class-specific.}
	\label{fig:pool_classwise}
\end{figure}

Compared with \Cref{eq:conj_pool}, \Cref{eq:classwise_pool} changes only the attention head (\Cref{fig:pool_classwise}). The gate becomes a vector of length $C$, so the evidence is formed class by class, and a time point can support one class while arguing against another. This head targets \textbf{L1}.

% ---------------------------------------------------------------------
\subsection{Soft-Max Conjunctive Pooling}
\label{app:pool:softmax}

\begin{figure}[!htbp]
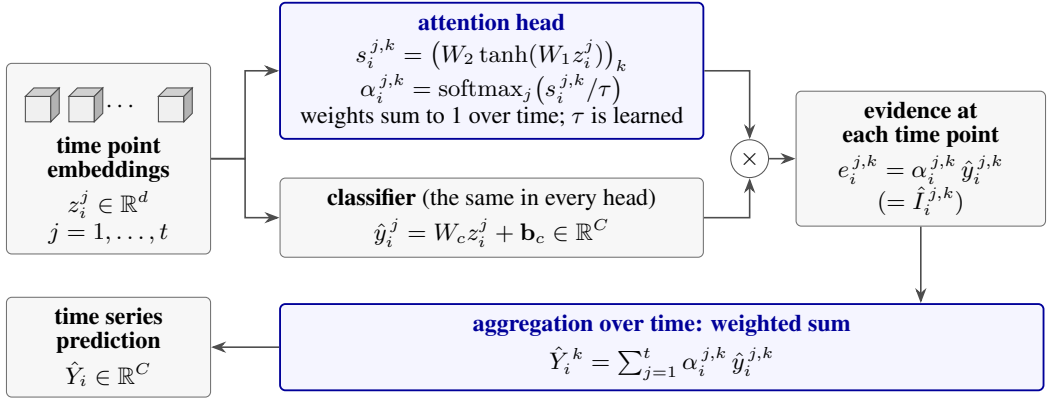

	\centering
	\poolfigwide{Soft-Max conjunctive pooling}
		{$s_i^{\,j,k} = \big(W_2\tanh(W_1 z_i^j)\big)_k$\\ $\alpha_i^{\,j,k} = \softmaxt\!\big(s_i^{\,j,k}/\tau\big)$\\ weights sum to 1 over time; $\tau$ is learned}
		{$e_i^{\,j,k} = \alpha_i^{\,j,k}\,\hat{y}_i^{\,j,k}$\\ $(= \hat{I}_i^{\,j,k})$}
		{weighted sum}
		{$\hat{Y}_i^{\,k} = \sum_{j=1}^{t} \alpha_i^{\,j,k}\,\hat{y}_i^{\,j,k}$}
	\caption{Soft-Max conjunctive pooling. Both blue boxes change compared with \Cref{fig:pool_conj}: the attention is normalized over time, and the aggregation is a weighted sum rather than a mean. The single extra weight is the sharpness $\tau = \exp(\rho)$.}
	\label{fig:pool_softmax}
\end{figure}

Here the aggregation changes as well, as \Cref{fig:pool_softmax} shows: the sigmoid gate becomes a Soft-Max over time, and the mean becomes a weighted sum. This head targets \textbf{L2}.

\begin{property}[Length invariance]
\label{prop:softmax}
Since the weights of \Cref{eq:softmax_pool} sum to one over $j$, $\hat{Y}_i^{\,k}$ does not depend on $t$. As $\tau \to \infty$ they flatten to $1/t$ and the head becomes a plain mean over time; as $\tau \to 0$ it keeps the single highest-scoring time point.
\end{property}

% ---------------------------------------------------------------------
\subsection{Top-$k$ Conjunctive Pooling}
\label{app:pool:topk}

\begin{figure}[!htbp]
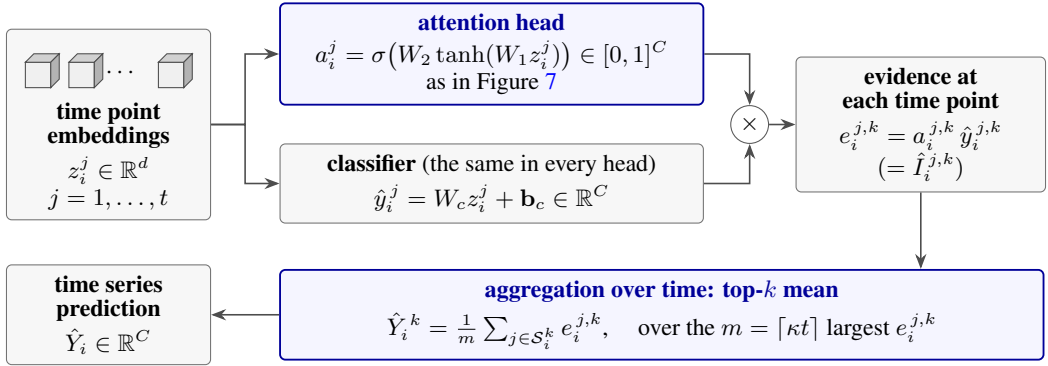

	\centering
	\poolfigwide{Top-$k$ conjunctive pooling}
		{$a_i^j = \sigma\big(W_2\tanh(W_1 z_i^j)\big) \in [0,1]^{C}$\\ as in \Cref{fig:pool_classwise}}
		{$e_i^{\,j,k} = a_i^{\,j,k}\,\hat{y}_i^{\,j,k}$\\ $(= \hat{I}_i^{\,j,k})$}
		{top-$k$ mean}
		{$\hat{Y}_i^{\,k} = \frac{1}{m}\sum_{j \in \mathcal{S}_i^{k}} e_i^{\,j,k}$, \quad over the $m = \lceil \kappa t \rceil$ largest $e_i^{\,j,k}$}
	\caption{Top-$k$ conjunctive pooling. The attention head is the one of \Cref{fig:pool_classwise}, and only the aggregation moves: the mean runs over the $m$ largest evidence values of the class instead of over all $t$. It adds no weight, because $\kappa$ is a hyperparameter.}
	\label{fig:pool_topk}
\end{figure}

The attention head is the class-wise one, and only the aggregation changes, as shown in \Cref{fig:pool_topk}. This head targets \textbf{L3}.

\begin{property}[Strict generalisation]
\label{prop:topk}
At $\kappa = 1$, \Cref{eq:topk_pool} is exactly \Cref{eq:classwise_pool}. For $\kappa < 1$ the gradient reaches only the selected time points.
\end{property}

\Cref{prop:topk} makes the $\kappa$ study of \Cref{tab:ablation_topk} a controlled experiment: the $\kappa = 1$ row \emph{is} the class-wise head, so any difference from it comes from the selection alone and not from a change of model.

The remaining four heads are not described in the main text. Each of them keeps the class-wise classifier and changes the attention head, the aggregation or both, and each targets one or two of the limitations above.

% ---------------------------------------------------------------------
\subsection{Adaptive Class-wise Pooling}
\label{app:pool:adaptive}

% \poolgridtwo   % old 2 x 2 grid, switched off above

\begin{figure}[!htbp]
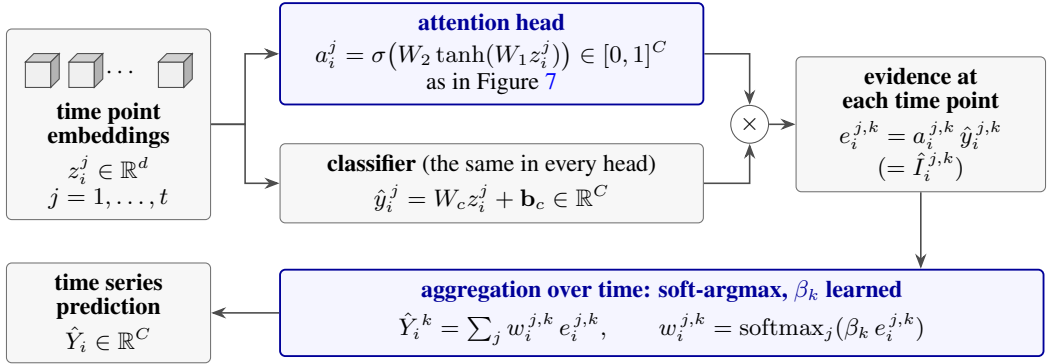

	\centering
	\poolfigwide{Adaptive class-wise pooling}
		{$a_i^j = \sigma\big(W_2\tanh(W_1 z_i^j)\big) \in [0,1]^{C}$\\ as in \Cref{fig:pool_classwise}}
		{$e_i^{\,j,k} = a_i^{\,j,k}\,\hat{y}_i^{\,j,k}$\\ $(= \hat{I}_i^{\,j,k})$}
		{soft-argmax, $\beta_k$ learned}
		{$\hat{Y}_i^{\,k} = \sum_{j} w_i^{\,j,k}\,e_i^{\,j,k}, \qquad w_i^{\,j,k} = \softmaxt(\beta_k\,e_i^{\,j,k})$}
	\caption{Adaptive class-wise pooling (\Cref{eq:app:adaptive}). The attention head is the one of \Cref{fig:pool_classwise}; the aggregation slides between the mean and the maximum according to one learned number per class.}
	\label{fig:pool_adaptive}
\end{figure}

This head also targets \textbf{L3}, but from another side. A fixed mean suits evidence that is spread over the series, and a maximum suits a single short event; no fixed operator is right for both. The head therefore keeps the class-wise gate and aggregates with a learnable sharpness $\beta_k \ge 0$ per class, written as a numerically stable soft-argmax (\Cref{fig:pool_adaptive}):
\begin{equation}
	\beta_k = \operatorname{softplus}(\theta_k),
	\qquad
	w_i^{\,j,k} = \softmaxt\!\big(\beta_k\, e_i^{\,j,k}\big),
	\qquad
	\hat{Y}_i^{\,k} = \sum_{j=1}^{t} w_i^{\,j,k}\, e_i^{\,j,k} .
	\label{eq:app:adaptive}
\end{equation}
It adds $C$ weights.

\begin{property}[Learned mean--max aggregator]
\label{prop:adaptive}
$\beta_k \to 0$ gives $w_i^{\,j,k} \to 1/t$, i.e. class-wise pooling; $\beta_k \to \infty$ gives $\hat{Y}_i^{\,k} \to \max_j e_i^{\,j,k}$. As $\hat{Y}_i^{\,k}$ is always a convex combination of the $e_i^{\,j,k}$, it stays inside their range, which keeps training stable.
\end{property}

We initialize $\beta_k$ at $0.3$, so that training starts close to the class-wise mean and becomes sharper only where this lowers the loss.

% ---------------------------------------------------------------------
\subsection{Attention-max Pooling}
\label{app:pool:attnmax}

\begin{figure}[!htbp]
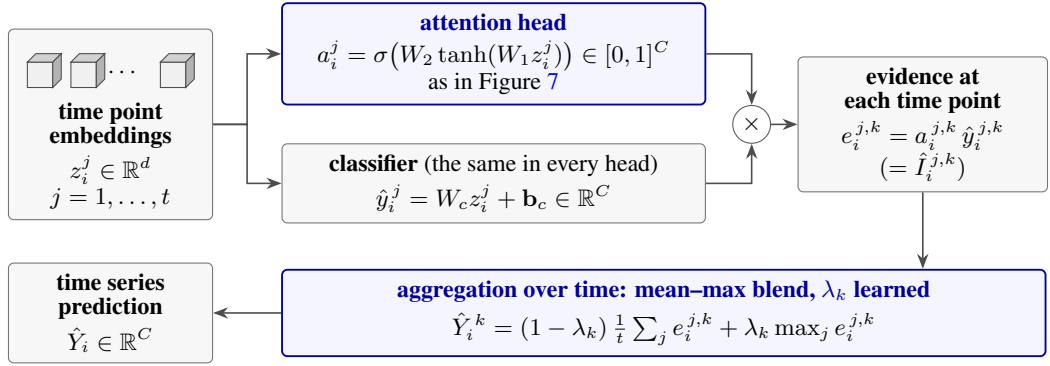

	\centering
	\poolfigwide{Attention-max pooling}
		{$a_i^j = \sigma\big(W_2\tanh(W_1 z_i^j)\big) \in [0,1]^{C}$\\ as in \Cref{fig:pool_classwise}}
		{$e_i^{\,j,k} = a_i^{\,j,k}\,\hat{y}_i^{\,j,k}$\\ $(= \hat{I}_i^{\,j,k})$}
		{mean--max blend, $\lambda_k$ learned}
		{$\hat{Y}_i^{\,k} = (1-\lambda_k)\,\frac{1}{t}\sum_{j} e_i^{\,j,k} + \lambda_k \max_{j} e_i^{\,j,k}$}
	\caption{Attention-max pooling (\Cref{eq:app:attnmax}), the hard counterpart of \Cref{fig:pool_adaptive}: one learned number per class blends the mean with the maximum directly.}
	\label{fig:pool_attnmax}
\end{figure}

This is the simplest hard alternative to the previous head, and it serves as a control: the soft-argmax still gives every time point some weight, and this head tests whether that is useful. The aggregation is a learnable per-class blend of the mean and the maximum (\Cref{fig:pool_attnmax}), again with $C$ extra weights:
\begin{equation}
	\hat{Y}_i^{\,k} = (1 - \lambda_k)\,\frac{1}{t}\sum_{j=1}^{t} e_i^{\,j,k}
	\;+\; \lambda_k \max_{j} e_i^{\,j,k},
	\qquad
	\lambda_k = \sigma(\theta_k) \in [0,1].
	\label{eq:app:attnmax}
\end{equation}
$\lambda_k \to 0$ recovers class-wise pooling and $\lambda_k \to 1$ gives max pooling, so each class learns on its own whether its evidence is sustained or a short event. \Cref{eq:app:attnmax} and \Cref{eq:app:adaptive} use the same gate and differ only in how they aggregate it, so comparing them isolates soft selection from hard selection.

% ---------------------------------------------------------------------
\subsection{Per-class Gated Attention}
\label{app:pool:gated}

\begin{figure}[!htbp]
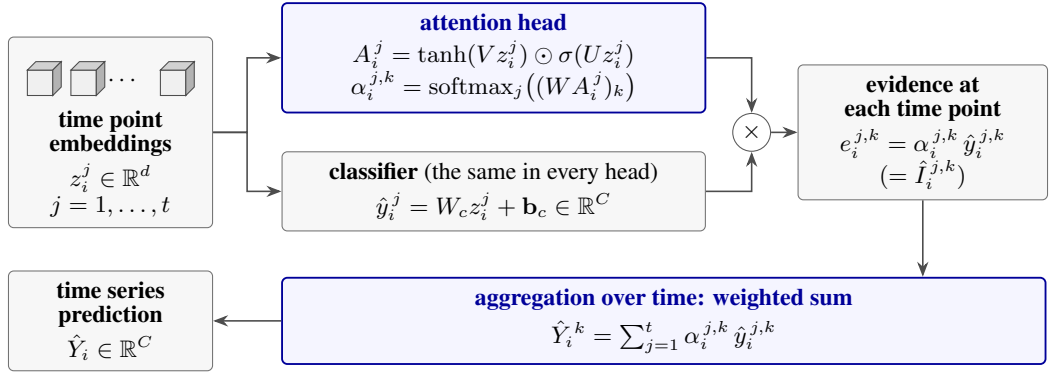

	\centering
	\poolfigwide{Per-class gated attention pooling}
		{$A_i^{\,j} = \tanh(V z_i^j) \odot \sigma(U z_i^j)$\\ $\alpha_i^{\,j,k} = \softmaxt\!\big((W A_i^{\,j})_k\big)$}
		{$e_i^{\,j,k} = \alpha_i^{\,j,k}\,\hat{y}_i^{\,j,k}$\\ $(= \hat{I}_i^{\,j,k})$}
		{weighted sum}
		{$\hat{Y}_i^{\,k} = \sum_{j=1}^{t} \alpha_i^{\,j,k}\,\hat{y}_i^{\,j,k}$}
	\caption{Per-class gated attention (\Cref{eq:app:ga_features} and \Cref{eq:app:ga_bag}). The attention head has two layers instead of one, and the second of them decides which attention features are admitted.}
	\label{fig:pool_gated}
\end{figure}

Gated attention was proposed by \citet{ilse2018attention} and is also used by \citet{hashimoto2020msdamil}, where a bag is a whole-slide image and an instance is a patch. It targets \textbf{L1} and \textbf{L3} together with a stronger attention head, because the baseline gate is a single $\tanh$ layer, which limits how selective it can be. We apply it to time series and make it per class, which the original is not (\Cref{fig:pool_gated}):
\begin{align}
	A_i^{\,j} &= \tanh(V z_i^j) \odot \sigma(U z_i^j) \in \mathbb{R}^{d_a},
	\qquad V, U \in \mathbb{R}^{d_a \times d},
	\label{eq:app:ga_features}\\
	\alpha_i^{\,j,k} &= \softmaxt\big((W A_i^{\,j})_k\big),
	\qquad
	\hat{Y}_i^{\,k} = \sum_{j=1}^{t} \alpha_i^{\,j,k}\, \hat{y}_i^{\,j,k} .
	\label{eq:app:ga_bag}
\end{align}
The $\tanh$ branch proposes attention features and the $\sigma$ branch decides which of them to let through, so a feature is used only where the second branch judges it reliable. The cost is a second $d \to d_a$ layer, which makes this the most expensive head here. Unlike the other heads, it has no blend parameter that recovers conjunctive pooling, so no setting of its weights returns the baseline.

% ---------------------------------------------------------------------
\subsection{Dual-stream Conjunctive Pooling}
\label{app:pool:dualstream}

\begin{figure}[!htbp]
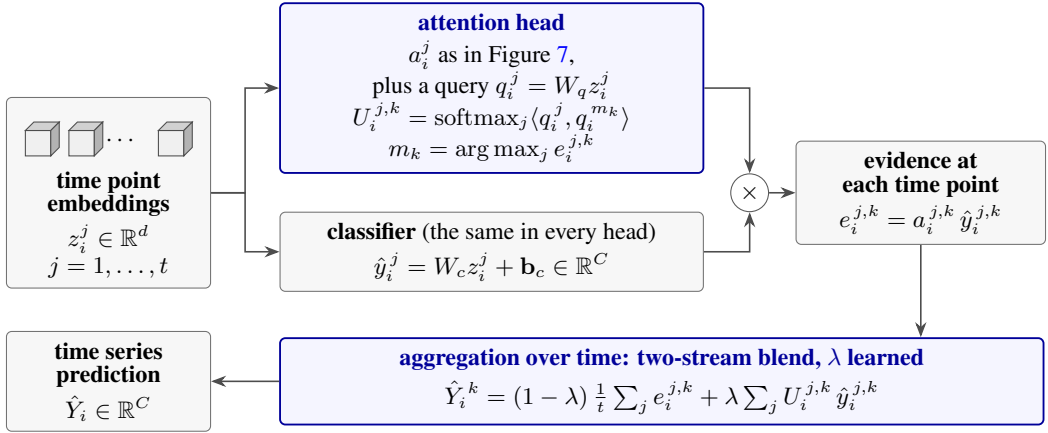

	\centering
	\poolfigwide{Dual-stream conjunctive pooling}
		{$a_i^j$ as in \Cref{fig:pool_classwise},\\ plus a query $q_i^{\,j} = W_q z_i^j$\\ $U_i^{\,j,k} = \softmaxt\langle q_i^{\,j}, q_i^{\,m_k}\rangle$\\ $m_k = \arg\max_j e_i^{\,j,k}$}
		{$e_i^{\,j,k} = a_i^{\,j,k}\,\hat{y}_i^{\,j,k}$}
		{two-stream blend, $\lambda$ learned}
		{$\hat{Y}_i^{\,k} = (1-\lambda)\,\frac{1}{t}\sum_{j} e_i^{\,j,k} + \lambda \sum_{j} U_i^{\,j,k}\,\hat{y}_i^{\,j,k}$}
	\caption{Dual-stream conjunctive pooling (\Cref{eq:app:ds_query} and \Cref{eq:app:ds_blend}). A second stream, built around the most supporting time point of each class, is blended into the class-wise stream of \Cref{fig:pool_classwise} by one learned scalar.}
	\label{fig:pool_dualstream}
\end{figure}

The two-stream mechanism was proposed by \citet{li2021dsmil}, again for whole-slide images: one stream aggregates all the instances, and a second one focuses on the most critical instance and on the instances similar to it. It targets \textbf{L3} when the evidence is scattered, i.e. one decisive time point should be able to dominate without the rest being discarded. We removed the contrastive pre-training of \citet{li2021dsmil}, made the critical instance and the query per class, and blended the two streams with a learnable scalar that starts close to the baseline. Stream A is the class-wise mean of \Cref{eq:classwise_pool}; stream B finds the most supporting time point of each class and re-weights the time points by their similarity to its query (\Cref{fig:pool_dualstream}):
\begin{align}
	m_k &= \argmax_{j}\; e_i^{\,j,k},
	\qquad
	q_i^{\,j} = W_q z_i^j \in \mathbb{R}^{d_a},
	\qquad
	U_i^{\,j,k} = \softmaxt\!\big(\langle q_i^{\,j}, q_i^{\,m_k}\rangle\big),
	\label{eq:app:ds_query}\\
	\hat{Y}_i^{\,k} &= (1-\lambda)\,\frac{1}{t}\sum_{j=1}^{t} e_i^{\,j,k}
	+ \lambda \sum_{j=1}^{t} U_i^{\,j,k}\, \hat{y}_i^{\,j,k},
	\qquad
	\lambda = \sigma(\theta), \;\; \lambda_0 = 0.05 .
	\label{eq:app:ds_blend}
\end{align}
This head is also the clearest negative result of the study: being able to return to the baseline is \textbf{not} a safety guarantee. Five of the seven heads contain conjunctive pooling as a special case, and this is one of them, yet it is the worst head we tested; its four pairings score $0.860$, $0.824$, $0.736$ and $0.556$ (\Cref{tab:ablation_pooling}). The training runs show why: $\lambda$ starts at $0.05$ and nothing in the loss pushes it back towards the baseline, so being \emph{able} to return there does not mean that training will go there. This is why the main text argues for \mbox{Top-$k$} from the measured \ndcg{} and not from \Cref{prop:topk} alone.

% ---------------------------------------------------------------------
\FloatBarrier
\subsection{The Eight Heads Together}
\label{app:heads_tables}

\Cref{tab:heads_summary} lists the eight heads by the two choices that define them, the attention head and the aggregation over time. \Cref{tab:arch_mil_conjunctive} to \Cref{tab:arch_sea_dualstream_conjunctive} then give the eight heads layer by layer, one table per head, with the input and the output shape of each operation. In these tables, $b$ is the batch size, $t$ the series length, $d$ the embedding size ($64$ for \seanet{}, $128$ for \millet{}), $a = 8$ the attention width and $c$ the number of classes.

\begin{table}[tbp]
\caption{The eight heads as two separate choices. \emph{Reduces at} gives the setting at which the head becomes conjunctive pooling exactly, and \emph{Targets} gives the limitations of \Cref{app:conjunctive} that it addresses.}
\label{tab:heads_summary}
\centering
\seatab
\begin{tabular}{r L{3.3cm} L{2.85cm} L{3.1cm} C{1.7cm} C{1.0cm}}
\toprule
\textbf{\#} & \textbf{Head} & \textbf{Attention} & \textbf{Reduction over time} & \textbf{Reduces at} & \textbf{Targets} \\
\midrule
1 & Conjunctive (baseline)    & shared gate                  & mean                             & ---                    & --- \\
2 & Class-wise conjunctive    & per-class gate               & mean                             & $a_i^{\,j,k} = a_i^j$  & L1 \\
3 & Soft-Max conjunctive      & per-class score              & softmax, learnable $\tau$        & $\tau \to \infty$      & L2 \\
4 & Top-$k$ conjunctive       & per-class gate               & mean of the $m$ largest          & $\kappa = 1$           & L3 \\
\midrule
5 & Adaptive class-wise       & per-class gate               & soft-argmax, learnable $\beta_k$ & $\beta_k \to 0$        & L3 \\
6 & Attention-max             & per-class gate               & blend of mean and max            & $\lambda_k \to 0$      & L3 \\
7 & Per-class gated attention & gated $\tanh\!\odot\!\sigma$ & softmax                          & \textbf{none}          & L1, L3 \\
8 & Dual-stream conjunctive   & gate $+$ critical query      & blend of two streams             & $\lambda \to 0$        & L3 \\
\bottomrule
\end{tabular}
\end{table}

\begin{table}[!htbp]
\caption{Conjunctive pooling, the \millet{} baseline head (their Table A.5; $d = 128$). One scalar gate per time point, shared by every class. Registered as \texttt{mil\_conjunctive}.}
\label{tab:arch_mil_conjunctive}
\centering
\seatab
\begin{tabular}{L{1.75cm}L{4.6cm}L{2.0cm}L{4.1cm}}
\toprule
\textbf{Process} & \textbf{Layer} & \textbf{Input} & \textbf{Output} \\
\midrule
Input & Positional encoding $+$ dropout & $b \times t \times d$ & $b \times t \times d$ (time-point embs.) \\
Attention & Linear $+$ tanh & $b \times t \times d$ & $b \times t \times a$ \\
 & Linear $+$ sigmoid & $b \times t \times a$ & $b \times t \times 1$ (attention scores) \\
Classifier & Linear & $b \times t \times d$ & $b \times t \times c$ (time-point logits) \\
Pooling & Attention weighting & $b \times t \times c$ & $b \times t \times c$ (interpretation) \\
 & Mean over $t$ & $b \times t \times c$ & $b \times c$ (series logits) \\
\bottomrule
\end{tabular}
\end{table}

\begin{table}[!htbp]
\caption{Class-wise conjunctive pooling: conjunctive pooling with one gate per class instead of one shared gate. Registered as \texttt{sea\_classwise\_conjunctive}.}
\label{tab:arch_sea_classwise_conjunctive}
\centering
\seatab
\begin{tabular}{L{1.75cm}L{4.6cm}L{2.0cm}L{4.1cm}}
\toprule
\textbf{Process} & \textbf{Layer} & \textbf{Input} & \textbf{Output} \\
\midrule
Input & Positional encoding $+$ dropout & $b \times t \times d$ & $b \times t \times d$ (time-point embs.) \\
Attention & Linear $+$ tanh & $b \times t \times d$ & $b \times t \times a$ \\
 & Linear $+$ sigmoid & $b \times t \times a$ & $b \times t \times c$ (one gate per class) \\
Classifier & Linear & $b \times t \times d$ & $b \times t \times c$ (time-point logits) \\
Evidence & Gate $\times$ logit (element-wise) & $b \times t \times c$ & $b \times t \times c$ (interpretation) \\
Pooling & Mean over $t$ & $b \times t \times c$ & $b \times c$ (series logits) \\
\bottomrule
\end{tabular}
\end{table}

\begin{table}[!htbp]
\caption{Top-$k$ conjunctive pooling, the head of every short-listed model: class-wise conjunctive pooling, but the series logit averages only the $k = \lceil \kappa t \rceil$ largest evidence values of each class ($\kappa = 0.1$). The interpretation is the full evidence map, so it is exactly zero wherever no evidence was used. Registered as \texttt{sea\_topk\_conjunctive}.}
\label{tab:arch_sea_topk_conjunctive}
\centering
\seatab
\begin{tabular}{L{1.75cm}L{4.6cm}L{2.0cm}L{4.1cm}}
\toprule
\textbf{Process} & \textbf{Layer} & \textbf{Input} & \textbf{Output} \\
\midrule
Input & Positional encoding $+$ dropout & $b \times t \times d$ & $b \times t \times d$ (time-point embs.) \\
Attention & Linear $+$ tanh & $b \times t \times d$ & $b \times t \times a$ \\
 & Linear $+$ sigmoid & $b \times t \times a$ & $b \times t \times c$ (one gate per class) \\
Classifier & Linear & $b \times t \times d$ & $b \times t \times c$ (time-point logits) \\
Evidence & Gate $\times$ logit (element-wise) & $b \times t \times c$ & $b \times t \times c$ (interpretation) \\
Pooling & Top-$k$ over $t$ (per class) & $b \times t \times c$ & $b \times k \times c$ \\
 & Mean over $k$ & $b \times k \times c$ & $b \times c$ (series logits) \\
\bottomrule
\end{tabular}
\end{table}

\begin{table}[!htbp]
\caption{Soft-Max conjunctive pooling: the attention is a distribution over time (a softmax with one learned temperature $\tau$), so the series logit is a true weighted average and long and short series are on one scale. Registered as \texttt{sea\_softmax\_conjunctive}.}
\label{tab:arch_sea_softmax_conjunctive}
\centering
\seatab
\begin{tabular}{L{1.75cm}L{4.6cm}L{2.0cm}L{4.1cm}}
\toprule
\textbf{Process} & \textbf{Layer} & \textbf{Input} & \textbf{Output} \\
\midrule
Input & Positional encoding $+$ dropout & $b \times t \times d$ & $b \times t \times d$ (time-point embs.) \\
Attention & Linear $+$ tanh & $b \times t \times d$ & $b \times t \times a$ \\
 & Linear (raw scores) & $b \times t \times a$ & $b \times t \times c$ \\
 & $\div\,\tau$, softmax over $t$ & $b \times t \times c$ & $b \times t \times c$ (weights, sum to 1) \\
Classifier & Linear & $b \times t \times d$ & $b \times t \times c$ (time-point logits) \\
Pooling & Weight $\times$ logit (element-wise) & $b \times t \times c$ & $b \times t \times c$ (interpretation) \\
 & Sum over $t$ & $b \times t \times c$ & $b \times c$ (series logits) \\
\bottomrule
\end{tabular}
\end{table}

\begin{table}[!htbp]
\caption{Adaptive class-wise pooling: class-wise conjunctive pooling with a learned mean-to-max aggregator, i.e. a soft-argmax with one sharpness $\beta_k = \mathrm{softplus}(\theta_k) \ge 0$ per class ($c$ extra parameters; $\beta \to 0$ gives the mean and $\beta \to \infty$ the max). Registered as \texttt{sea\_adaptive\_classwise}.}
\label{tab:arch_sea_adaptive_classwise}
\centering
\seatab
\begin{tabular}{L{1.75cm}L{4.6cm}L{2.0cm}L{4.1cm}}
\toprule
\textbf{Process} & \textbf{Layer} & \textbf{Input} & \textbf{Output} \\
\midrule
Input & Positional encoding $+$ dropout & $b \times t \times d$ & $b \times t \times d$ (time-point embs.) \\
Attention & Linear $+$ tanh & $b \times t \times d$ & $b \times t \times a$ \\
 & Linear $+$ sigmoid & $b \times t \times a$ & $b \times t \times c$ (one gate per class) \\
Classifier & Linear & $b \times t \times d$ & $b \times t \times c$ (time-point logits) \\
Evidence & Gate $\times$ logit (element-wise) & $b \times t \times c$ & $b \times t \times c$ (interpretation) \\
Pooling & softmax over $t$ of $\beta_k \cdot$ evidence & $b \times t \times c$ & $b \times t \times c$ (weights) \\
 & Weighted sum over $t$ & $b \times t \times c$ & $b \times c$ (series logits) \\
\bottomrule
\end{tabular}
\end{table}

\begin{table}[!htbp]
\caption{Attention-max pooling: class-wise conjunctive pooling with a learned per-class blend $\lambda_k = \sigma(\theta_k)$ of the mean and the max over time ($c$ extra parameters). Registered as \texttt{sea\_attention\_max}.}
\label{tab:arch_sea_attention_max}
\centering
\seatab
\begin{tabular}{L{1.75cm}L{4.6cm}L{2.0cm}L{4.1cm}}
\toprule
\textbf{Process} & \textbf{Layer} & \textbf{Input} & \textbf{Output} \\
\midrule
Input & Positional encoding $+$ dropout & $b \times t \times d$ & $b \times t \times d$ (time-point embs.) \\
Attention & Linear $+$ tanh & $b \times t \times d$ & $b \times t \times a$ \\
 & Linear $+$ sigmoid & $b \times t \times a$ & $b \times t \times c$ (one gate per class) \\
Classifier & Linear & $b \times t \times d$ & $b \times t \times c$ (time-point logits) \\
Evidence & Gate $\times$ logit (element-wise) & $b \times t \times c$ & $b \times t \times c$ (interpretation) \\
Pooling & Mean over $t$ & $b \times t \times c$ & $b \times c$ \\
 & Max over $t$ & $b \times t \times c$ & $b \times c$ \\
 & $(1-\lambda_k)\,$mean $+ \lambda_k\,$max & $b \times c$ & $b \times c$ (series logits) \\
\bottomrule
\end{tabular}
\end{table}

\begin{table}[!htbp]
\caption{Per-class gated attention pooling (\citealt{ilse2018attention}, made class-specific): a tanh branch multiplied by a sigmoid branch scores every time point, and a softmax over time turns the scores into weights. Registered as \texttt{sea\_gated\_attention}.}
\label{tab:arch_sea_gated_attention}
\centering
\seatab
\begin{tabular}{L{1.75cm}L{4.6cm}L{2.0cm}L{4.1cm}}
\toprule
\textbf{Process} & \textbf{Layer} & \textbf{Input} & \textbf{Output} \\
\midrule
Input & Positional encoding $+$ dropout & $b \times t \times d$ & $b \times t \times d$ (time-point embs.) \\
Attention & Linear $+$ tanh ($V$) & $b \times t \times d$ & $b \times t \times a$ \\
 & Linear $+$ sigmoid ($U$) & $b \times t \times d$ & $b \times t \times a$ \\
 & Element-wise product & $b \times t \times a$ & $b \times t \times a$ (gated features) \\
 & Linear (scores), softmax over $t$ & $b \times t \times a$ & $b \times t \times c$ (weights, sum to 1) \\
Classifier & Linear & $b \times t \times d$ & $b \times t \times c$ (time-point logits) \\
Pooling & Weight $\times$ logit (element-wise) & $b \times t \times c$ & $b \times t \times c$ (interpretation) \\
 & Sum over $t$ & $b \times t \times c$ & $b \times c$ (series logits) \\
\bottomrule
\end{tabular}
\end{table}

\begin{table}[!htbp]
\caption{Dual-stream conjunctive pooling (DSMIL, \citealt{li2021dsmil}, wrapped around class-wise conjunctive pooling): stream A is the plain mean, and stream B re-weights the time points by their similarity to the most supporting one (the critical time point) of each class; a learned scalar $\lambda = \sigma(\theta)$ blends the two and starts near stream A. Registered as \texttt{sea\_dualstream\_conjunctive}.}
\label{tab:arch_sea_dualstream_conjunctive}
\centering
\seatab
\begin{tabular}{L{1.75cm}L{4.6cm}L{2.0cm}L{4.1cm}}
\toprule
\textbf{Process} & \textbf{Layer} & \textbf{Input} & \textbf{Output} \\
\midrule
Input & Positional encoding $+$ dropout & $b \times t \times d$ & $b \times t \times d$ (time-point embs.) \\
Attention & Linear $+$ tanh & $b \times t \times d$ & $b \times t \times a$ \\
 & Linear $+$ sigmoid & $b \times t \times a$ & $b \times t \times c$ (one gate per class) \\
Classifier & Linear & $b \times t \times d$ & $b \times t \times c$ (time-point logits) \\
Evidence & Gate $\times$ logit (element-wise) & $b \times t \times c$ & $b \times t \times c$ (interpretation) \\
Stream A & Mean over $t$ & $b \times t \times c$ & $b \times c$ \\
Stream B & Linear (query) & $b \times t \times d$ & $b \times t \times a$ \\
 & argmax over $t$ of the evidence & $b \times t \times c$ & $b \times c$ (critical index) \\
 & Gather the critical query & $b \times t \times a$ & $b \times c \times a$ \\
 & Dot product with every query, softmax over $t$ & $b \times t \times a$ & $b \times t \times c$ (weights) \\
 & Weight $\times$ logit, sum over $t$ & $b \times t \times c$ & $b \times c$ \\
Blend & $(1-\lambda)\,$A $+ \lambda\,$B & $b \times c$ & $b \times c$ (series logits) \\
\bottomrule
\end{tabular}
\end{table}

% =====================================================================
\section{Reproducing the Experiments}
\label{app:setup_all}

\Cref{sec:setup_new} gives the experimental setup in short: the two datasets, \webtraffic{} and the $85$ \ucr{} datasets used by \millet{}, the metrics, i.e. the test accuracy for the classification and AOPCR and \ndcg{} for the interpretation, and the \millet{} baseline. This appendix gives the details needed to reproduce the experiments. \Cref{app:datasets} shows the \webtraffic{} classes and lists every dataset, \Cref{app:baselines} describes the baselines and how they were trained, \Cref{app:hparams} gives the training configuration shared by all models, \Cref{app:metrics} defines the two interpretation metrics, and \Cref{app:hparams_all} lists every hyperparameter and the size of every trained model.

% ---------------------------------------------------------------------
\subsection{Datasets}
\label{app:datasets}

\begin{figure}[!htbp]
	\centering
	\includegraphics[width=\linewidth]{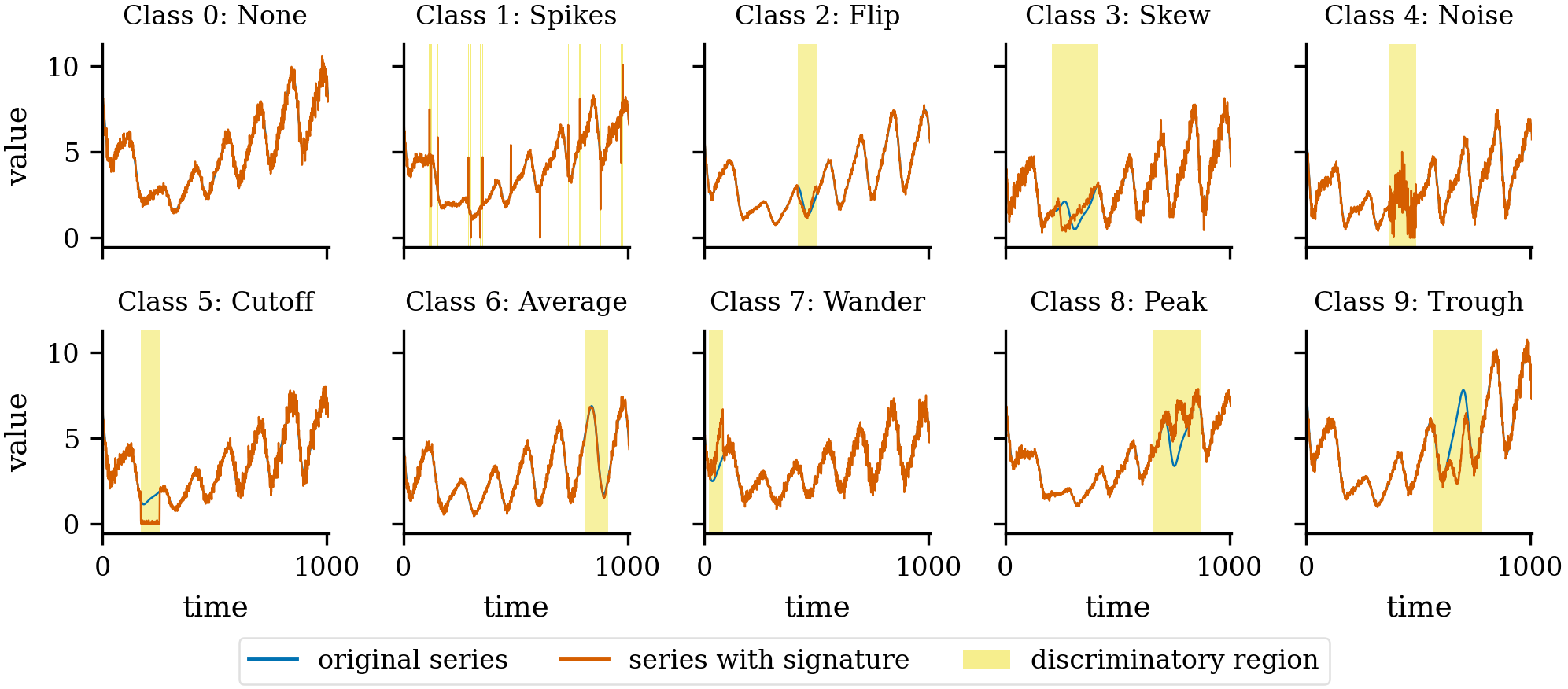}
	\caption{One example of each class of the \webtraffic{} dataset. Each series is a daily seasonal signal with a weekly trend and noise, and the signature that defines the class is injected inside a single window, shown shaded; the exception is class $0$ (None), which has no signature. These windows are the per-time-point ground truth against which \ndcg{} is scored, and they are never used during training.}
	\label{fig:web_classes}
\end{figure}

\webtraffic{} is the only dataset in this work with a ground truth for each time point, so it is the only one on which the interpretation can be scored with \ndcg{}. \Cref{fig:web_classes} shows one example of each of its classes. \Cref{tab:datasets_all} lists \webtraffic{} and the $85$ \ucr{} datasets with their train and test sizes, lengths, numbers of classes and class imbalance.

\begingroup
\seatab\renewcommand{\arraystretch}{1.12}
\begin{longtable}{r L{4.7cm}R{1.35cm}R{1.35cm}R{1.35cm}R{1.25cm}R{1.6cm}}
\caption{The datasets: \webtraffic{} (the only one with per-time-point labels, hence the only one with \ndcg{}) and the 85 \ucr{} datasets that \millet{} publishes, in the order of \millet{}. \emph{Imbalance} is the largest training-class share divided by the smallest (1.00 = balanced). The sizes are the original train / test splits; our training holds out 20\,\% of the training set for validation when it has at least 100 series.}
\label{tab:datasets_all} \\
\toprule
\textbf{\#} & \textbf{Dataset} & \textbf{Train} & \textbf{Test} & \textbf{Length} & \textbf{Classes} & \textbf{Imbalance} \\
\midrule
\endfirsthead
\multicolumn{7}{l}{\emph{\tablename~\thetable{} (continued)}} \\
\toprule
\textbf{\#} & \textbf{Dataset} & \textbf{Train} & \textbf{Test} & \textbf{Length} & \textbf{Classes} & \textbf{Imbalance} \\
\midrule
\endhead
\midrule
\multicolumn{7}{r}{\emph{continued on the next page}} \\
\endfoot
\bottomrule
\endlastfoot
 & WebTraffic & 500 & 500 & 1{,}008 & 10 & 1.00 \\
\midrule
1 & Adiac & 390 & 391 & 176 & 37 & 3.75 \\
2 & ArrowHead & 36 & 175 & 251 & 3 & 1.00 \\
3 & Beef & 30 & 30 & 470 & 5 & 1.00 \\
4 & BeetleFly & 20 & 20 & 512 & 2 & 1.00 \\
5 & BirdChicken & 20 & 20 & 512 & 2 & 1.00 \\
6 & Car & 60 & 60 & 577 & 4 & 1.55 \\
7 & CBF & 30 & 900 & 128 & 3 & 1.50 \\
8 & ChlorineConcentration & 467 & 3{,}840 & 166 & 3 & 2.88 \\
9 & CinCECGTorso & 40 & 1{,}380 & 1{,}639 & 4 & 2.60 \\
10 & Coffee & 28 & 28 & 286 & 2 & 1.00 \\
11 & Computers & 250 & 250 & 720 & 2 & 1.00 \\
12 & CricketX & 390 & 390 & 300 & 12 & 1.56 \\
13 & CricketY & 390 & 390 & 300 & 12 & 1.29 \\
14 & CricketZ & 390 & 390 & 300 & 12 & 1.64 \\
15 & DiatomSizeReduction & 16 & 306 & 345 & 4 & 6.00 \\
16 & DistalPhalanxOutlineAgeGroup & 400 & 139 & 80 & 3 & 8.57 \\
17 & DistalPhalanxOutlineCorrect & 600 & 276 & 80 & 2 & 1.70 \\
18 & DistalPhalanxTW & 400 & 139 & 80 & 6 & 11.83 \\
19 & Earthquakes & 322 & 139 & 512 & 2 & 4.55 \\
20 & ECG200 & 100 & 100 & 96 & 2 & 2.23 \\
21 & ECG5000 & 500 & 4{,}500 & 140 & 5 & 146.00 \\
22 & ECGFiveDays & 23 & 861 & 136 & 2 & 1.56 \\
23 & ElectricDevices & 8{,}926 & 7{,}711 & 96 & 7 & 4.73 \\
24 & FaceAll & 560 & 1{,}690 & 131 & 14 & 1.00 \\
25 & FaceFour & 24 & 88 & 350 & 4 & 2.67 \\
26 & FacesUCR & 200 & 2{,}050 & 131 & 14 & 8.25 \\
27 & FiftyWords & 450 & 455 & 270 & 50 & 52.00 \\
28 & Fish & 175 & 175 & 463 & 7 & 1.33 \\
29 & FordA & 3{,}601 & 1{,}320 & 500 & 2 & 1.05 \\
30 & FordB & 3{,}636 & 810 & 500 & 2 & 1.05 \\
31 & GunPoint & 50 & 150 & 150 & 2 & 1.08 \\
32 & Ham & 109 & 105 & 431 & 2 & 1.10 \\
33 & HandOutlines & 1{,}000 & 370 & 2{,}709 & 2 & 1.76 \\
34 & Haptics & 155 & 308 & 1{,}092 & 5 & 2.00 \\
35 & Herring & 64 & 64 & 512 & 2 & 1.56 \\
36 & InlineSkate & 100 & 550 & 1{,}882 & 7 & 2.00 \\
37 & InsectWingbeatSound & 220 & 1{,}980 & 256 & 11 & 1.00 \\
38 & ItalyPowerDemand & 67 & 1{,}029 & 24 & 2 & 1.03 \\
39 & LargeKitchenAppliances & 375 & 375 & 720 & 3 & 1.00 \\
40 & Lightning2 & 60 & 61 & 637 & 2 & 2.00 \\
41 & Lightning7 & 70 & 73 & 319 & 7 & 3.80 \\
42 & Mallat & 55 & 2{,}345 & 1{,}024 & 8 & 5.50 \\
43 & Meat & 60 & 60 & 448 & 3 & 1.00 \\
44 & MedicalImages & 381 & 760 & 99 & 10 & 33.83 \\
45 & MiddlePhalanxOutlineAgeGroup & 400 & 154 & 80 & 3 & 4.31 \\
46 & MiddlePhalanxOutlineCorrect & 600 & 291 & 80 & 2 & 1.83 \\
47 & MiddlePhalanxTW & 399 & 154 & 80 & 6 & 6.40 \\
48 & MoteStrain & 20 & 1{,}252 & 84 & 2 & 1.00 \\
49 & NonInvasiveFetalECGThorax1 & 1{,}800 & 1{,}965 & 750 & 42 & 1.60 \\
50 & NonInvasiveFetalECGThorax2 & 1{,}800 & 1{,}965 & 750 & 42 & 1.60 \\
51 & OliveOil & 30 & 30 & 570 & 4 & 3.25 \\
52 & OSULeaf & 200 & 242 & 427 & 6 & 3.53 \\
53 & PhalangesOutlinesCorrect & 1{,}800 & 858 & 80 & 2 & 1.87 \\
54 & Phoneme & 214 & 1{,}896 & 1{,}024 & 39 & 24.00 \\
55 & Plane & 105 & 105 & 144 & 7 & 2.22 \\
56 & ProximalPhalanxOutlineAgeGroup & 400 & 205 & 80 & 3 & 2.62 \\
57 & ProximalPhalanxOutlineCorrect & 600 & 291 & 80 & 2 & 2.09 \\
58 & ProximalPhalanxTW & 400 & 205 & 80 & 6 & 11.25 \\
59 & RefrigerationDevices & 375 & 375 & 720 & 3 & 1.00 \\
60 & ScreenType & 375 & 375 & 720 & 3 & 1.00 \\
61 & ShapeletSim & 20 & 180 & 500 & 2 & 1.00 \\
62 & ShapesAll & 600 & 600 & 512 & 60 & 1.00 \\
63 & SmallKitchenAppliances & 375 & 375 & 720 & 3 & 1.00 \\
64 & SonyAIBORobotSurface1 & 20 & 601 & 70 & 2 & 2.33 \\
65 & SonyAIBORobotSurface2 & 27 & 953 & 65 & 2 & 1.45 \\
66 & StarLightCurves & 1{,}000 & 8{,}236 & 1{,}024 & 3 & 3.77 \\
67 & Strawberry & 613 & 370 & 235 & 2 & 1.80 \\
68 & SwedishLeaf & 500 & 625 & 128 & 15 & 1.62 \\
69 & Symbols & 25 & 995 & 398 & 6 & 2.67 \\
70 & SyntheticControl & 300 & 300 & 60 & 6 & 1.00 \\
71 & ToeSegmentation1 & 40 & 228 & 277 & 2 & 1.00 \\
72 & ToeSegmentation2 & 36 & 130 & 343 & 2 & 1.00 \\
73 & Trace & 100 & 100 & 275 & 4 & 1.48 \\
74 & TwoLeadECG & 23 & 1{,}139 & 82 & 2 & 1.09 \\
75 & TwoPatterns & 1{,}000 & 4{,}000 & 128 & 4 & 1.14 \\
76 & UWaveGestureLibraryAll & 896 & 3{,}582 & 945 & 8 & 1.27 \\
77 & UWaveGestureLibraryX & 896 & 3{,}582 & 315 & 8 & 1.27 \\
78 & UWaveGestureLibraryY & 896 & 3{,}582 & 315 & 8 & 1.27 \\
79 & UWaveGestureLibraryZ & 896 & 3{,}582 & 315 & 8 & 1.27 \\
80 & Wafer & 1{,}000 & 6{,}164 & 152 & 2 & 9.31 \\
81 & Wine & 57 & 54 & 234 & 2 & 1.11 \\
82 & WordSynonyms & 267 & 638 & 270 & 25 & 30.00 \\
83 & Worms & 181 & 77 & 900 & 5 & 4.47 \\
84 & WormsTwoClass & 181 & 77 & 900 & 2 & 1.38 \\
85 & Yoga & 300 & 3{,}000 & 426 & 2 & 1.19 \\
\midrule
 & \emph{UCR-85: min / median / max} & 16 / 267 / 8926 & 20 / 375 / 8236 & 24 / 300 / 2709 & 2 / 3 / 60 & 1.00 / 1.60 / 146.00 \\
\end{longtable}
\endgroup

% ---------------------------------------------------------------------
\subsection{Baselines}
\label{app:baselines}

\millet{} (\texttt{InceptionTime} encoder $+$ \texttt{Conjunctive} pooling) is the baseline of this work. We also train FCN and ResNet \citep{wang2017fcn} with the same pooling head, to have two more reference points. The faithfulness numbers of two different training recipes cannot be compared (\Cref{sec:res:aopcr}), so we retrain every baseline ourselves with our own training budget, which is the only fair comparison. We also list the published numbers of \millet{} \citep{early2024millet} for reference (\Cref{tab:published}).
% Removed: the results of MILLET retrained with its published recipe in our code.
% We report \millet{} twice: once with our own training budget, which is the only fair comparison, and once with its published $1{,}500$-epoch recipe, which is the harder target. The two retrained rows agree with the published numbers to within a few thousandths (\Cref{tab:published}), which confirms that the difference between them comes from the training budget and not from our re-implementation.

Two things differ between the recipe behind the published numbers and ours. First, they keep no validation split and use no early stopping, so over-fitting cannot be observed during training. Second, they train with hard targets, i.e. the whole probability mass is put on one class, whereas we add label smoothing.

% ---------------------------------------------------------------------
\subsection{Training Configuration}
\label{app:hparams}

We use Optuna to search for the hyperparameters, and the values it selects are then used for every model of this work. Hence, a difference between two rows of a result table is a difference between two models, and not between two training setups. \Cref{tab:protocol} gives the shared setting, and \Cref{tab:hparams} gives the remaining hyperparameters, with one column per model.

\begin{table}[!htbp]
\caption{Training configuration, identical for every model of this study. The complete hyperparameter list, including the head-specific initial values, is given in \Cref{tab:hparams}.}
\label{tab:protocol}
\centering
\seatab
\begin{tabular}{L{4.0cm}L{9.2cm}}
\toprule
\textbf{Setting} & \textbf{Value} \\
\midrule
Encoder size (every \seanet{} model) & width $d = 64$, 4 blocks; 41{,}324--69{,}768 parameters \\
Optimizer              & Adam \citep{kingma2015adam}, lr $=1.25\times10^{-3}$, weight decay $=1.2\times10^{-4}$ \\
Epochs / early stopping& max 400 epochs; stop after 60 epochs with no improvement \\
Batch size             & $\min(\lfloor n_{\text{train}}/10 \rfloor, 16)$, floor of 2 \\
Validation split       & stratified 80/20 when $n_{\text{train}} \geq 100$; else train-loss early stopping \\
Label smoothing        & 0.13 \\
Dropout                & 0.2 \\
Attention width $d_a$  & 8 \\
Top-$k$ fraction $\kappa$ & 0.1 (\Cref{eq:topk_pool}) \\
Head initial values    & $\tau_0 = 1.0$, $\beta_0 = 0.3$, $\lambda_0 = 0.5$ (attention-max) or $0.05$ (dual-stream) \\
Focus penalty $\lambda_{\text{focus}}$ & 0.01; 0 for the \millet{} baselines, since the term is ours \\
Seeds                  & 3 for the models of \Cref{tab:seeds}; 1 for the rest \\
Hardware               & GPU nodes of a shared academic computing cluster \\
\bottomrule
\end{tabular}
\end{table}

The wide encoder uses the same setting, with $d = 128$ and 6 blocks.
% Removed (published-recipe run): The published-recipe baseline uses \millet{}'s own settings instead, where the main difference is $1{,}500$ training epochs against our $400$.
\webtraffic{} is generated with the released default settings, and each series is standardized on its own; nothing else is done to the data.

% ---------------------------------------------------------------------
\subsection{Interpretation Metrics}
\label{app:metrics}
\label{app:metrics_new}

\Cref{sec:setup_new} describes the two interpretation metrics in words: AOPCR removes the time points in the order of their estimated importance and measures how much the score of the predicted class drops, and \ndcg{} compares the ranking of the time points with the ground truth. Here we give their exact definitions.

\paragraph{AOPCR.}
Let $\hat{Y}_i^{\,c}(X_i)$ be the logit of the predicted class $c$ on the unchanged time series, and let $X_i^{(m)}$ be the same series in which the $m$ time points that the model itself ranks as the most important for class $c$ are replaced by the mean of that series. AOPCR adds up the drop of the logit along this order,
\begin{equation}
	\mathrm{AOPCR} = \frac{1}{M+1}\sum_{m=0}^{M}
	\Big( \hat{Y}_i^{\,c}(X_i) - \hat{Y}_i^{\,c}\big(X_i^{(m)}\big) \Big),
	\label{eq:aopcr_new}
\end{equation}
following \citet{samek2017aopc} as adapted to time series by \citet{early2024millet}, with $M$ set to $10\,\%$ of the series length in steps of $1\,\%$. A higher value is better. Note that \Cref{eq:aopcr_new} is a difference of logits, and logits have no fixed scale. This causes the comparison problem of \Cref{sec:res:aopcr}: two models whose logits lie on different scales give AOPCR values on different scales, however good or bad their ranking of the time points is.

\paragraph{\ndcg{}.}
When the ground truth of each time point is available, as in \webtraffic{}, we rank the $t$ time points of the series by the importance $\hat{I}_i^{\,j,c}$ that the model gives them for the predicted class $c$, and score this ranking with the normalized discounted cumulative gain \citep{jarvelin2002ndcg} at the cut-off $n$, where $n$ is the number of time points that the generator marked as carrying the class signature,
\begin{equation}
	\mathrm{DCG}@n = \sum_{p=1}^{n} \frac{r_p}{\log_2(p+1)}, \qquad
	\mathrm{NDCG}@n = \frac{\mathrm{DCG}@n}{\mathrm{IDCG}@n} \in [0,1],
	\label{eq:ndcg_new}
\end{equation}
where $r_p \in \{0,1\}$ tells whether the time point placed at rank $p$ by the model belongs to the true signature window, and $\mathrm{IDCG}@n$ is the same sum for the best possible ranking of the same series. Since the score is divided by the best possible score, \ndcg{} lies in $[0,1]$ and can be compared between models, datasets and papers. This is why the interpretation claims of this paper rest on \ndcg{} and not on AOPCR.

% ---------------------------------------------------------------------
\subsection{Full Hyperparameters}
\label{app:hparams_all}

\Cref{tab:hparams} lists every hyperparameter, with one row per hyperparameter and one column per model, read directly from the configuration files. The two \seanet{} columns and the \millet{} column with our recipe share the same training block, so any difference between them is in the architecture. \Cref{tab:params_grid}, set on landscape pages, then gives the number of parameters of every encoder $\times$ pooling head pair that was trained, measured on \webtraffic{} with its $10$ classes, together with the change relative to the same encoder with the class-wise conjunctive head.

\begin{table}[!htbp]
\caption{Every hyperparameter, read from the configuration files under \texttt{configs/models/}. The two \seanet{} columns and \millet{} (our recipe) share one training block, so any difference between them is the architecture. -- = not applicable (the InceptionTime backbone is the code of \millet{} with its published defaults, so it has no settings here).}
\label{tab:hparams}
\centering
\seatab\setlength{\tabcolsep}{3pt}
\begin{tabular}{L{3.4cm}C{3.0cm}C{3.0cm}C{3.0cm}}
\toprule
\textbf{Parameter} & \textbf{SEA-Net bottleneck $+$ Top-$k$ (the paper's model)} & \textbf{SEA-Net gated (mean) $+$ Top-$k$} & \textbf{MILLET, our recipe} \\
\midrule
\multicolumn{4}{l}{\emph{Training}} \\
max epochs & 400 & 400 & 400 \\
early-stopping patience & 60 & 60 & 60 \\
batch size (max) & 16 & 16 & 16 \\
learning rate & 0.00125 & 0.00125 & 0.00125 \\
weight decay & 0.00012 & 0.00012 & 0.00012 \\
label smoothing & 0.13 & 0.13 & 0.13 \\
$\lambda_{\text{focus}}$ (attention entropy) & 0.01 & 0.01 & 0.0 \\
min.\ train size for a val.\ split & 100 & 100 & 100 \\
validation fraction & 0.2 & 0.2 & 0.2 \\
optimizer & adam & adam & adam \\
\midrule
\multicolumn{4}{l}{\emph{Encoder}} \\
type (registry name) & \texttt{sea\_\allowbreak mstcn\_\allowbreak sep\_\allowbreak bottleneck} & \texttt{sea\_\allowbreak mstcn\_\allowbreak sep\_\allowbreak gated} & \texttt{mil\_\allowbreak inceptiontime} \\
width $d$ & 64 & 64 & -- \\
blocks & 4 & 4 & -- \\
dropout & 0.2 & 0.2 & -- \\
dilation cap & 16 & 16 & -- \\
kernel sizes & \{5, 11, 23\} & \{5, 11, 23\} & -- \\
bottleneck ratio ($d / r$) & 4 & -- & -- \\
gate summary statistic & -- & mean & -- \\
\midrule
\multicolumn{4}{l}{\emph{Pooling head}} \\
type (registry name) & \texttt{sea\_\allowbreak topk\_\allowbreak conjunctive} & \texttt{sea\_\allowbreak topk\_\allowbreak conjunctive} & \texttt{mil\_\allowbreak conjunctive} \\
attention width $a$ & 8 & 8 & 8 \\
dropout & 0.2 & 0.2 & 0.1 \\
positional encoding & True & True & True \\
Top-$k$ fraction $\kappa$ & 0.1 & 0.1 & -- \\
\bottomrule
\end{tabular}
\end{table}

\begin{landscape}
\begingroup
\seatab\setlength{\tabcolsep}{2.5pt}
\begin{longtable}{r L{4.3cm}C{1.42cm}C{1.42cm}C{1.42cm}C{1.42cm}C{1.42cm}C{1.42cm}C{1.42cm}C{1.42cm}C{1.42cm}C{1.42cm}}
\caption{Number of parameters of every encoder $\times$ pooling head pair that was trained, measured on \webtraffic{} (10 classes; the size of a head grows with the number of classes). The percentages give the change relative to the same encoder with class-wise conjunctive pooling, in the manner of Table A.13 of \millet{}: every attention head costs well under 1\,\% of the encoder, so the size of a model is set by its encoder. The bottom rows are the baseline encoders with the conjunctive head of \millet{}.}
\label{tab:params_grid} \\
\toprule
\textbf{\#} & \textbf{Encoder $\backslash$ head} & \textbf{class-wise conj.} & \textbf{Top-$k$ conj.} & \textbf{softmax conj.} & \textbf{adaptive cw.} & \textbf{attn-max} & \textbf{gated attn.} & \textbf{dual-stream} & \textbf{conj.$^{\mathrm{M}}$} & \textbf{additive$^{\mathrm{M}}$} & \textbf{attention$^{\mathrm{M}}$} \\
\midrule
\endfirsthead
\multicolumn{12}{l}{\emph{\tablename~\thetable{} (continued)}} \\
\toprule
\textbf{\#} & \textbf{Encoder $\backslash$ head} & \textbf{class-wise conj.} & \textbf{Top-$k$ conj.} & \textbf{softmax conj.} & \textbf{adaptive cw.} & \textbf{attn-max} & \textbf{gated attn.} & \textbf{dual-stream} & \textbf{conj.$^{\mathrm{M}}$} & \textbf{additive$^{\mathrm{M}}$} & \textbf{attention$^{\mathrm{M}}$} \\
\midrule
\endhead
\midrule
\multicolumn{12}{r}{\emph{continued on the next page}} \\
\endfoot
\bottomrule
\endlastfoot
1 & bottleneck (shallow) & -- & 21{,}548 & -- & -- & -- & -- & -- & -- & -- & -- \\
2 & multi-scale pyramid (bottleneck base, over channels) & -- & 28{,}441 & -- & -- & -- & -- & -- & -- & -- & -- \\
3 & bottleneck & 41{,}324 & 41{,}324 \newline (+0.00\%) & 41{,}325 \newline (+0.00\%) & 41{,}334 \newline (+0.02\%) & 41{,}334 \newline (+0.02\%) & 41{,}844 \newline (+1.26\%) & 41{,}845 \newline (+1.26\%) & -- & -- & -- \\
4 & multi-scale pyramid (bottleneck base) & -- & 42{,}365 & -- & -- & -- & -- & -- & -- & -- & -- \\
5 & multi-scale channels (bottleneck base, lean) & -- & 43{,}576 & -- & -- & -- & -- & -- & -- & -- & -- \\
6 & multi-scale channels (bottleneck base) & -- & 47{,}176 & -- & -- & -- & -- & -- & -- & -- & -- \\
7 & base & 57{,}580 & 57{,}580 \newline (+0.00\%) & 57{,}581 \newline (+0.00\%) & 57{,}590 \newline (+0.02\%) & 57{,}590 \newline (+0.02\%) & 58{,}100 \newline (+0.90\%) & 58{,}101 \newline (+0.90\%) & -- & -- & -- \\
8 & input-gated & 58{,}092 & 58{,}092 \newline (+0.00\%) & 58{,}093 \newline (+0.00\%) & 58{,}102 \newline (+0.02\%) & 58{,}102 \newline (+0.02\%) & -- & -- & -- & -- & -- \\
9 & gated & 61{,}740 & -- & -- & -- & -- & -- & -- & -- & -- & 58{,}100 \newline (-5.90\%) \\
10 & gated (mean) & 61{,}740 & 61{,}740 \newline (+0.00\%) & 61{,}741 \newline (+0.00\%) & 61{,}750 \newline (+0.02\%) & 61{,}750 \newline (+0.02\%) & -- & -- & -- & -- & -- \\
11 & gated (max) & 61{,}740 & 61{,}740 \newline (+0.00\%) & 61{,}741 \newline (+0.00\%) & 61{,}750 \newline (+0.02\%) & 61{,}750 \newline (+0.02\%) & -- & -- & -- & -- & -- \\
12 & gated (last) & 61{,}740 & 61{,}740 \newline (+0.00\%) & 61{,}741 \newline (+0.00\%) & 61{,}750 \newline (+0.02\%) & 61{,}750 \newline (+0.02\%) & 62{,}260 \newline (+0.84\%) & 62{,}261 \newline (+0.84\%) & -- & -- & -- \\
13 & multi-scale pyramid (gated base) & -- & 62{,}781 & -- & -- & -- & -- & -- & -- & -- & -- \\
14 & reconstruction-residual & 62{,}925 & 62{,}925 \newline (+0.00\%) & 62{,}926 \newline (+0.00\%) & 62{,}935 \newline (+0.02\%) & 62{,}935 \newline (+0.02\%) & -- & -- & -- & -- & -- \\
15 & spike/trend & 67{,}020 & 67{,}020 \newline (+0.00\%) & 67{,}021 \newline (+0.00\%) & 67{,}030 \newline (+0.01\%) & 67{,}030 \newline (+0.01\%) & 67{,}540 \newline (+0.78\%) & 67{,}541 \newline (+0.78\%) & -- & -- & -- \\
16 & multi-scale channels (gated base) & -- & 67{,}592 & -- & -- & -- & -- & -- & -- & -- & -- \\
17 & multi-scale channels (input-gated base) & -- & 69{,}768 & -- & -- & -- & -- & -- & -- & -- & -- \\
18 & base (wide, $d=128$) & 269{,}164 & -- & 269{,}165 \newline (+0.00\%) & 269{,}174 \newline (+0.00\%) & -- & -- & -- & 269{,}083 \newline (-0.03\%) & 269{,}083 \newline (-0.03\%) & -- \\
\midrule
19 & InceptionTime (MILLET) & -- & -- & -- & -- & -- & -- & -- & 423{,}707 & -- & -- \\
20 & ResNet & -- & -- & -- & -- & -- & -- & -- & 506{,}331 & -- & -- \\
21 & FCN & -- & -- & -- & -- & -- & -- & -- & 267{,}035 & -- & -- \\
\end{longtable}
\endgroup
\end{landscape}

% =====================================================================
\section{Complete Results}
\label{app:additional}

\Cref{sec:res:main} reports the main results in short. On \webtraffic{}, our \seanet{} bottleneck $+$ Top-$k$ model is more accurate than \millet{} and gives a better \ndcg{} with about ten times fewer parameters. On the $85$ \ucr{} datasets it stays slightly behind, and this gap comes from the reduced capacity and not from the design. The main text also answers RQ4 (the bottleneck) and RQ5 (Top-$k$ pooling), and shows that AOPCR can only be compared within one training recipe.

This appendix gives the complete results behind these statements, in the same order as the main text. \Cref{app:web_all} gives every model trained on \webtraffic{}, \Cref{app:cost_all} the cost of the models, \Cref{app:ablations_all} the ablation studies, \Cref{app:seeds_all} the results of every seed, \Cref{app:ucr_all} the complete \ucr{} results, and \Cref{app:published_all} compares our \millet{} baseline with the published numbers of \millet{} \citep{early2024millet}.

% ---------------------------------------------------------------------
\subsection{Complete \webtraffic{} Results}
\label{app:web_all}
\label{app:leaderboard}

\begin{figure}[!htbp]
	\centering
	\includegraphics[width=0.95\linewidth]{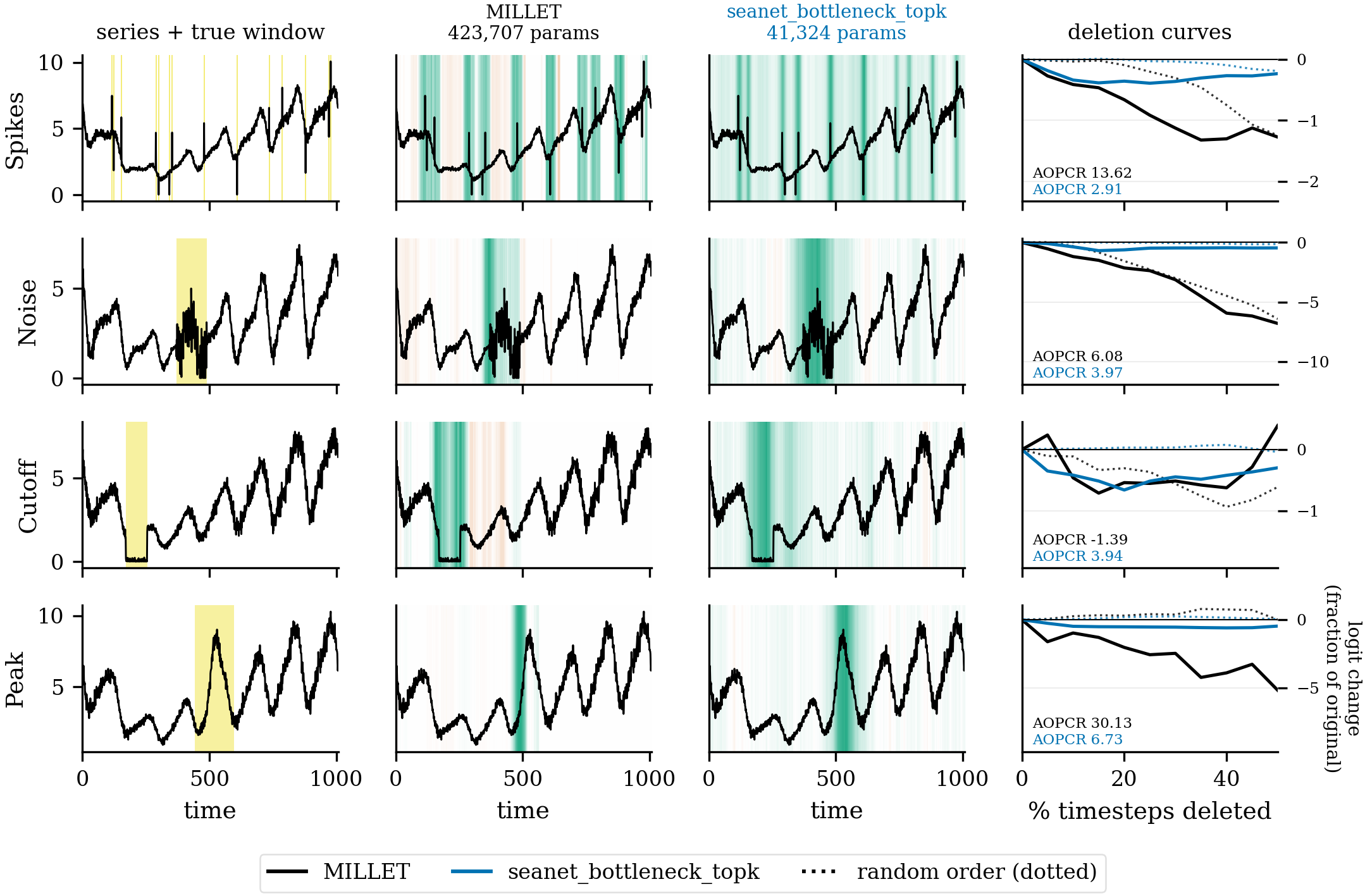}
	\caption{Head-to-head comparison of the explanations of \millet{} (published recipe) and of our
		\seanet{} bottleneck $+$ Top-$k$ model on four \webtraffic{} classes. The panels show, left to right, the series with its true signature window, the interpretation heatmap for the true class, and the perturbation curves under each model's own deletion order (solid) against a random one (dotted).
		%		{\color{violet}\rdt{XXI}  \emph{Left:} the series with
			%		its true signature window. \emph{Middle:} the interpretation heatmap of each model for the true
			%		class. \emph{Right:} the perturbation curves, i.e. the change of the true class logit as the time
			%		points are deleted in the order the explanation of each model proposes (solid) or in a random
			%		order (dotted); the faster the curve falls, the more the prediction really depended on the time
			%		points the model pointed at.}
		The vertical axis is the change as a fraction of the original logit,
		because the two models produce logits on very different scales (\Cref{app:published}). The AOPCR of each panel is the area between the two curves; \millet{} wins it on three of the four classes while our heatmaps are the more concentrated --- the difference between the two metrics is analysed in \Cref{sec:res:aopcr}.}
	\label{fig:head_to_head}
\end{figure}

\Cref{fig:head_to_head} first compares the explanations of \millet{} and of our model on four \webtraffic{} classes. It shows, on single series, the difference between the two interpretation metrics: \millet{} wins AOPCR on three of the four classes, while our heatmaps are the more concentrated.

The tables below give every trained model. \Cref{tab:leaderboard_sweep} is the complete \webtraffic{} leaderboard of the $71$-model sweep, ranked by accuracy, where one row is one encoder $+$ pooling head pair; the rows are grouped into five accuracy bands, so that the table can be read one band at a time. \Cref{tab:pooling_round} is a later round in which every new head was trained on the bottleneck encoder, so the row of our model is the one to beat there. \Cref{tab:grid_a} and \Cref{tab:grid_b} give the same sweep as a grid, with one encoder per row and one pooling head per column, for the accuracy and for AOPCR / \ndcg{} respectively; a dash means that the pair was not trained. These three tables are wide and are set on landscape pages.

\begin{landscape}
\begingroup
\seatab
\begin{longtable}{r L{2.4cm}L{6.2cm}C{1.1cm}C{1.1cm}C{1.1cm}C{1.5cm}C{1.4cm}R{1.4cm}C{1.8cm}R{1.2cm}}
\caption{The 2026-09 pooling-head round on \webtraffic{} (results/top\_results, one seed each): every new head is trained on the bottleneck encoder (one on the gated encoder), so row~1 (the model of the paper, trained again) is the number to beat. The \millet{} row is the baseline retrained with our recipe. \emph{Epochs run / best} gives the epoch at which early stopping fired and the epoch whose weights were kept; the voting heads that fail to fit are visible in their NLL.}
\label{tab:pooling_round} \\
\toprule
\textbf{\#} & \textbf{Encoder} & \textbf{Pooling head} & \textbf{Acc.} & \textbf{AUROC} & \textbf{NLL} & \textbf{AOPCR} & \textbf{NDCG@$n$} & \textbf{Params} & \textbf{Epochs run / best} & \textbf{Train (s)} \\
\midrule
\endfirsthead
\multicolumn{11}{l}{\emph{\tablename~\thetable{} (continued)}} \\
\toprule
\textbf{\#} & \textbf{Encoder} & \textbf{Pooling head} & \textbf{Acc.} & \textbf{AUROC} & \textbf{NLL} & \textbf{AOPCR} & \textbf{NDCG@$n$} & \textbf{Params} & \textbf{Epochs run / best} & \textbf{Train (s)} \\
\midrule
\endhead
\midrule
\multicolumn{11}{r}{\emph{continued on the next page}} \\
\endfoot
\bottomrule
\endlastfoot
1 & bottleneck & Top-$k$ conjunctive (paper model, re-run) & 0.942 & 0.996 & 0.294 & 2.951 & 0.786 & 41{,}324 & 180 / 120 & 54 \\
2 & bottleneck & Top-$k$ conjunctive, thresholded & 0.930 & 0.996 & 0.330 & 2.603 & 0.744 & 41{,}324 & 172 / 112 & 52 \\
% hidden: MILLET trained with its published recipe in our code (not reported)
% 3 & InceptionTime & conjunctive (MILLET, their recipe) & 0.916 & 0.993 & 0.413 & 10.852$^\dagger$ & 0.624 & 423{,}707 & 1500 / 1056 & 8602 \\
3 & bottleneck & softmax conjunctive, per-class $\tau$ & 0.914 & 0.991 & 0.377 & 2.076 & 0.732 & 41{,}334 & 151 / 91 & 65 \\
4 & bottleneck & deviation-max (MLP form) & 0.912 & 0.990 & 0.369 & 2.394 & 0.742 & 42{,}544 & 164 / 104 & 276 \\
5 & bottleneck & DSMIL (batched rewrite) & 0.900 & 0.991 & 0.477 & 1.462 & 0.667 & 71{,}956 & 379 / 319 & 611 \\
6 & InceptionTime & conjunctive (MILLET, our recipe) & 0.890 & 0.985 & 0.531 & 2.120 & 0.692 & 423{,}707 & 184 / 124 & 443 \\
7 & bottleneck & DSMIL (original code) & 0.888 & 0.991 & 0.488 & 1.242 & 0.536 & 71{,}956 & 319 / 259 & 2152 \\
8 & bottleneck & Top-$k$ voting, thresholded & 0.886 & 0.987 & 0.468 & 2.195 & 0.604 & 41{,}325 & 247 / 187 & 71 \\
9 & bottleneck & deviation-max (dual form) & 0.878 & 0.988 & 0.451 & 2.356 & 0.723 & 41{,}944 & 176 / 116 & 69 \\
10 & bottleneck & Top-$k$ voting & 0.870 & 0.983 & 0.516 & 2.212 & 0.601 & 41{,}325 & 218 / 158 & 61 \\
11 & bottleneck & adaptive class-wise + MLP & 0.836 & 0.982 & 0.653 & 1.528 & 0.618 & 41{,}375 & 195 / 135 & 78 \\
12 & bottleneck & DSMIL (batched rewrite, no pos.) & 0.832 & 0.976 & 0.632 & 1.312 & 0.639 & 71{,}956 & 205 / 145 & 318 \\
13 & gated (mean) & DSMIL (original code) & 0.810 & 0.971 & 0.692 & 0.996 & 0.341 & 92{,}372 & 126 / 66 & 1043 \\
14 & bottleneck & Top-$k$ voting, hard & 0.672 & 0.902 & 1.173 & 1.770 & 0.697 & 41{,}325 & 168 / 108 & 50 \\
15 & bottleneck & simple voting & 0.100 & 0.531 & 4.511 & 0.150 & 0.222 & 41{,}324 & 73 / 13 & 22 \\
\end{longtable}
\endgroup
\newpage
\begingroup
\seatab\setlength{\tabcolsep}{3pt}\renewcommand{\arraystretch}{1.1}
\begin{longtable}{r L{3.6cm}C{1.45cm}C{1.45cm}C{1.45cm}C{1.45cm}C{1.45cm}C{1.45cm}C{1.45cm}C{1.45cm}C{1.45cm}C{1.45cm}}
\caption{\webtraffic{} accuracy for every encoder $\times$ pooling head pair that was trained (71-model sweep; the $\kappa$ / $\lambda$ ablation copies are left out). -- = pair not trained; a superscript count means that several configurations share the cell and the most accurate one is shown. $^{\mathrm{M}}$ = a \millet{} head. The best value of each column is in bold.}
\label{tab:grid_a} \\
\toprule
\textbf{\#} & \textbf{Encoder $\backslash$ head} & \textbf{class-wise conj.} & \textbf{Top-$k$ conj.} & \textbf{softmax conj.} & \textbf{adaptive cw.} & \textbf{attn-max} & \textbf{gated attn.} & \textbf{dual-stream} & \textbf{conj.$^{\mathrm{M}}$} & \textbf{additive$^{\mathrm{M}}$} & \textbf{attention$^{\mathrm{M}}$} \\
\midrule
\endfirsthead
\multicolumn{12}{l}{\emph{\tablename~\thetable{} (continued)}} \\
\toprule
\textbf{\#} & \textbf{Encoder $\backslash$ head} & \textbf{class-wise conj.} & \textbf{Top-$k$ conj.} & \textbf{softmax conj.} & \textbf{adaptive cw.} & \textbf{attn-max} & \textbf{gated attn.} & \textbf{dual-stream} & \textbf{conj.$^{\mathrm{M}}$} & \textbf{additive$^{\mathrm{M}}$} & \textbf{attention$^{\mathrm{M}}$} \\
\midrule
\endhead
\midrule
\multicolumn{12}{r}{\emph{continued on the next page}} \\
\endfoot
\bottomrule
\endlastfoot
1 & gated (mean) & 0.902 & \textbf{0.955} & 0.912 & 0.860 & 0.832 & -- & -- & -- & -- & -- \\
2 & base (wide, $d=128$) & \textbf{0.946} & -- & 0.880 & 0.918 & -- & -- & -- & \textbf{0.954} & \textbf{0.942} & -- \\
3 & gated (max) & 0.944 & 0.952 & 0.914 & 0.906 & 0.886 & -- & -- & -- & -- & -- \\
4 & spike/trend & 0.904 & 0.950 & 0.878 & 0.934 & 0.854 & \textbf{0.930} & 0.556 & -- & -- & -- \\
5 & multi-scale channels (input-gated base) & -- & 0.948 & -- & -- & -- & -- & -- & -- & -- & -- \\
6 & multi-scale channels (gated base) & -- & 0.948 & -- & -- & -- & -- & -- & -- & -- & -- \\
7 & reconstruction-residual & 0.910 & 0.940 & 0.932 & \textbf{0.946} & \textbf{0.892} & -- & -- & -- & -- & -- \\
8 & input-gated & 0.914 & 0.946 & 0.894 & 0.905 & 0.858 & -- & -- & -- & -- & -- \\
9 & gated (last) & 0.942 & 0.942 & 0.920 & 0.938 & 0.854 & 0.918 & 0.736 & -- & -- & -- \\
10 & bottleneck & 0.902 & 0.931 & \textbf{0.940} & 0.868 & 0.816 & \textbf{0.930} & 0.824 & -- & -- & -- \\
11 & base & 0.916 & 0.932 & 0.890 & 0.930 & 0.714 & 0.894 & \textbf{0.860} & -- & -- & -- \\
12 & gated & 0.930 & -- & -- & -- & -- & -- & -- & -- & -- & \textbf{0.888} \\
13 & multi-scale channels (bottleneck base) & -- & 0.926 & -- & -- & -- & -- & -- & -- & -- & -- \\
14 & multi-scale channels (bottleneck base, lean) & -- & 0.924 & -- & -- & -- & -- & -- & -- & -- & -- \\
15 & InceptionTime & -- & -- & -- & -- & -- & -- & -- & 0.887 & -- & -- \\
16 & bottleneck (shallow) & -- & 0.846 & -- & -- & -- & -- & -- & -- & -- & -- \\
17 & multi-scale pyramid (gated base) & -- & 0.832 & -- & -- & -- & -- & -- & -- & -- & -- \\
18 & multi-scale pyramid (bottleneck base) & -- & 0.786 & -- & -- & -- & -- & -- & -- & -- & -- \\
19 & ResNet & -- & -- & -- & -- & -- & -- & -- & 0.772 & -- & -- \\
20 & FCN & -- & -- & -- & -- & -- & -- & -- & 0.742 & -- & -- \\
21 & multi-scale pyramid (bottleneck base, over channels) & -- & 0.696 & -- & -- & -- & -- & -- & -- & -- & -- \\
\end{longtable}
\endgroup
\newpage
\begingroup
\seatab\setlength{\tabcolsep}{2.5pt}\renewcommand{\arraystretch}{1.1}
\begin{longtable}{r L{3.2cm}C{1.6cm}C{1.6cm}C{1.6cm}C{1.6cm}C{1.6cm}C{1.6cm}C{1.6cm}C{1.6cm}C{1.6cm}C{1.6cm}}
\caption{\webtraffic{} AOPCR / \ndcg{} for every encoder $\times$ pooling head pair that was trained (71-model sweep; the $\kappa$ / $\lambda$ ablation copies are left out). -- = pair not trained; a superscript count means that several configurations share the cell and the most accurate one is shown. $^{\mathrm{M}}$ = a \millet{} head. AOPCR is unnormalised (the same recipe is used everywhere here).}
\label{tab:grid_b} \\
\toprule
\textbf{\#} & \textbf{Encoder $\backslash$ head} & \textbf{class-wise conj.} & \textbf{Top-$k$ conj.} & \textbf{softmax conj.} & \textbf{adaptive cw.} & \textbf{attn-max} & \textbf{gated attn.} & \textbf{dual-stream} & \textbf{conj.$^{\mathrm{M}}$} & \textbf{additive$^{\mathrm{M}}$} & \textbf{attention$^{\mathrm{M}}$} \\
\midrule
\endfirsthead
\multicolumn{12}{l}{\emph{\tablename~\thetable{} (continued)}} \\
\toprule
\textbf{\#} & \textbf{Encoder $\backslash$ head} & \textbf{class-wise conj.} & \textbf{Top-$k$ conj.} & \textbf{softmax conj.} & \textbf{adaptive cw.} & \textbf{attn-max} & \textbf{gated attn.} & \textbf{dual-stream} & \textbf{conj.$^{\mathrm{M}}$} & \textbf{additive$^{\mathrm{M}}$} & \textbf{attention$^{\mathrm{M}}$} \\
\midrule
\endhead
\midrule
\multicolumn{12}{r}{\emph{continued on the next page}} \\
\endfoot
\bottomrule
\endlastfoot
1 & gated (mean) & 2.49 / 0.714 & 2.23 / 0.750 & 2.09 / 0.765 & 2.24 / 0.731 & 1.67 / 0.704 & -- & -- & -- & -- & -- \\
2 & base (wide, $d=128$) & 1.72 / 0.686 & -- & 1.02 / 0.605 & 1.61 / 0.619 & -- & -- & -- & 1.50 / 0.698 & 1.58 / 0.729 & -- \\
3 & gated (max) & 1.89 / 0.592 & 2.27 / 0.719 & 1.88 / 0.776 & 2.07 / 0.729 & 1.12 / 0.575 & -- & -- & -- & -- & -- \\
4 & spike/trend & 2.15 / 0.699 & 2.32 / 0.756 & 1.43 / 0.665 & 1.91 / 0.747 & 1.65 / 0.644 & 1.59 / 0.721 & 1.64 / 0.510 & -- & -- & -- \\
5 & multi-scale channels (input-gated base) & -- & 2.32 / 0.757 & -- & -- & -- & -- & -- & -- & -- & -- \\
6 & multi-scale channels (gated base) & -- & 2.22 / 0.717 & -- & -- & -- & -- & -- & -- & -- & -- \\
7 & reconstruction-residual & 2.50 / 0.680 & 2.51 / 0.698 & 1.64 / 0.723 & 2.60 / 0.768 & 2.32 / 0.725 & -- & -- & -- & -- & -- \\
8 & input-gated & 2.48 / 0.715 & 2.80 / 0.748 & 1.64 / 0.696 & 2.65 / 0.732 & 2.15 / 0.700 & -- & -- & -- & -- & -- \\
9 & gated (last) & 2.52 / 0.742 & 2.31 / 0.748 & 1.99 / 0.738 & 2.39 / 0.793 & 1.78 / 0.668 & 1.35 / 0.757 & 1.95 / 0.637 & -- & -- & -- \\
10 & bottleneck & 2.57 / 0.733 & 2.50 / 0.765 & 1.68 / 0.740 & 1.97 / 0.680 & 1.74 / 0.610 & 2.23 / 0.741 & 2.04 / 0.652 & -- & -- & -- \\
11 & base & 2.48 / 0.719 & 2.69 / 0.778 & 1.83 / 0.705 & 2.45 / 0.793 & 1.63 / 0.532 & 2.30 / 0.714 & 2.40 / 0.745 & -- & -- & -- \\
12 & gated & 2.41 / 0.637 & -- & -- & -- & -- & -- & -- & -- & -- & 2.40 / 0.710 \\
13 & multi-scale channels (bottleneck base) & -- & 2.24 / 0.777 & -- & -- & -- & -- & -- & -- & -- & -- \\
14 & multi-scale channels (bottleneck base, lean) & -- & 2.76 / 0.735 & -- & -- & -- & -- & -- & -- & -- & -- \\
15 & InceptionTime & -- & -- & -- & -- & -- & -- & -- & 2.57 / 0.677 & -- & -- \\
16 & bottleneck (shallow) & -- & 2.82 / 0.624 & -- & -- & -- & -- & -- & -- & -- & -- \\
17 & multi-scale pyramid (gated base) & -- & 0.98 / 0.591 & -- & -- & -- & -- & -- & -- & -- & -- \\
18 & multi-scale pyramid (bottleneck base) & -- & 1.03 / 0.569 & -- & -- & -- & -- & -- & -- & -- & -- \\
19 & ResNet & -- & -- & -- & -- & -- & -- & -- & 2.95 / 0.554 & -- & -- \\
20 & FCN & -- & -- & -- & -- & -- & -- & -- & 3.83 / 0.533 & -- & -- \\
21 & multi-scale pyramid (bottleneck base, over channels) & -- & 2.17 / 0.653 & -- & -- & -- & -- & -- & -- & -- & -- \\
\end{longtable}
\endgroup
\end{landscape}

\begingroup
\seatab\setlength{\tabcolsep}{3pt}
\begin{longtable}{r L{2.4cm}L{2.45cm}C{0.8cm}C{0.8cm}C{1.2cm}C{1.4cm}R{1.15cm}C{1.2cm}}
\caption{The complete \webtraffic{} leaderboard of the 71-model sweep (results/old\_results), ranked by accuracy and grouped into accuracy bands. $^\ast$ marks the mean over three seeds (seeds 0--2, the runs of \Cref{tab:seeds}); every other row is a single run with seed 0, so a comparison between a $^\ast$ row and an unmarked row sets a three-run mean against one run. \emph{UCR-85 acc.} is the mean accuracy over the 85 \ucr{} datasets for the models that completed the archive, and -- for the models that were only screened on \webtraffic{}. \emph{Encoder} drops the \seanet{} prefix (MS = multi-scale); \emph{wide} is the $d = 128$ encoder; $^{\mathrm{M}}$ = a \millet{} head. AUROC is given in the csv file next to this table.}
\label{tab:leaderboard_sweep} \\
\toprule
\textbf{\#} & \textbf{Encoder} & \textbf{Pooling head} & \textbf{Acc.} & \textbf{NLL} & \textbf{AOPCR} & \textbf{NDCG@$n$} & \textbf{Params} & \textbf{UCR-85 acc.} \\
\midrule
\endfirsthead
\multicolumn{9}{l}{\emph{\tablename~\thetable{} (continued)}} \\
\toprule
\textbf{\#} & \textbf{Encoder} & \textbf{Pooling head} & \textbf{Acc.} & \textbf{NLL} & \textbf{AOPCR} & \textbf{NDCG@$n$} & \textbf{Params} & \textbf{UCR-85 acc.} \\
\midrule
\endhead
\midrule
\multicolumn{9}{r}{\emph{continued on the next page}} \\
\endfoot
\bottomrule
\endlastfoot
\multicolumn{9}{l}{\emph{Accuracy $\geq 0.94$}} \\
1 & gated (mean) & Top-$k$ conj.$^\ast$ & 0.955 & 0.254 & 2.225 & 0.750 & 61{,}740 & 0.8083 \\
2 & base (wide, $d=128$) & conj.$^{\mathrm{M}}$ & 0.954 & 0.262 & 1.502 & 0.698 & 269{,}083 & 0.8238 \\
3 & gated (max) & Top-$k$ conj. & 0.952 & 0.308 & 2.268 & 0.719 & 61{,}740 & 0.8108 \\
4 & bottleneck & Top-$k$ conj. ($\lambda_{\text{focus}}{=}0$) & 0.950 & 0.263 & 2.303 & 0.765 & 41{,}324 & -- \\
5 & spike/trend & Top-$k$ conj. & 0.950 & 0.288 & 2.316 & 0.756 & 67{,}020 & 0.8089 \\
6 & MS channels (input-gated base) & Top-$k$ conj. & 0.948 & 0.318 & 2.316 & 0.757 & 69{,}768 & -- \\
7 & MS channels (gated base) & Top-$k$ conj. & 0.948 & 0.289 & 2.223 & 0.717 & 67{,}592 & -- \\
8 & base (wide, $d=128$) & class-wise conj. & 0.946 & 0.292 & 1.722 & 0.686 & 269{,}164 & 0.8292 \\
9 & input-gated & Top-$k$ conj. & 0.946 & 0.261 & 2.803 & 0.748 & 58{,}092 & 0.8131 \\
10 & reconstruction-residual & adaptive cw. & 0.946 & 0.296 & 2.601 & 0.768 & 62{,}935 & 0.8140 \\
11 & gated (max) & class-wise conj. & 0.944 & 0.306 & 1.889 & 0.592 & 61{,}740 & 0.8237 \\
12 & gated (last) & class-wise conj. & 0.942 & 0.327 & 2.520 & 0.742 & 61{,}740 & 0.7888 \\
13 & gated (last) & Top-$k$ conj. & 0.942 & 0.328 & 2.312 & 0.748 & 61{,}740 & 0.8015 \\
14 & base (wide, $d=128$) & additive$^{\mathrm{M}}$ & 0.942 & 0.260 & 1.579 & 0.729 & 269{,}083 & 0.8298 \\
15 & bottleneck & Top-$k$ conj. ($\kappa{=}0.25$) & 0.940 & 0.277 & 2.648 & 0.733 & 41{,}324 & -- \\
16 & reconstruction-residual & Top-$k$ conj. & 0.940 & 0.321 & 2.505 & 0.698 & 62{,}925 & 0.8063 \\
17 & bottleneck & softmax conj. & 0.940 & 0.345 & 1.685 & 0.740 & 41{,}325 & 0.8106 \\
\midrule
\multicolumn{9}{l}{\emph{Accuracy in $[0.92, 0.94)$}} \\
18 & gated (last) & adaptive cw. & 0.938 & 0.289 & 2.387 & 0.793 & 61{,}750 & 0.8029 \\
19 & spike/trend & adaptive cw. & 0.934 & 0.301 & 1.913 & 0.747 & 67{,}030 & 0.8010 \\
20 & base & Top-$k$ conj. & 0.932 & 0.306 & 2.691 & 0.778 & 57{,}580 & 0.8093 \\
21 & reconstruction-residual & softmax conj. & 0.932 & 0.341 & 1.642 & 0.723 & 62{,}926 & -- \\
22 & bottleneck & Top-$k$ conj.$^\ast$ & 0.931 & 0.312 & 2.498 & 0.765 & 41{,}324 & 0.8097 \\
23 & gated & class-wise conj. & 0.930 & 0.359 & 2.408 & 0.637 & 61{,}740 & 0.7900 \\
24 & base & adaptive cw. & 0.930 & 0.336 & 2.448 & 0.793 & 57{,}590 & -- \\
25 & bottleneck & gated attn. & 0.930 & 0.313 & 2.232 & 0.741 & 41{,}844 & 0.8146 \\
26 & spike/trend & gated attn. & 0.930 & 0.350 & 1.593 & 0.721 & 67{,}540 & 0.8039 \\
27 & MS channels (bottleneck base) & Top-$k$ conj. & 0.926 & 0.357 & 2.238 & 0.777 & 47{,}176 & -- \\
28 & MS channels (bottleneck base, lean) & Top-$k$ conj. & 0.924 & 0.389 & 2.762 & 0.735 & 43{,}576 & -- \\
29 & gated (last) & softmax conj. & 0.920 & 0.381 & 1.990 & 0.738 & 61{,}741 & -- \\
% hidden: MILLET trained with its published recipe in our code (not reported)
% 30 & InceptionTime & conj.$^{\mathrm{M}}$ & 0.920 & 0.371 & 13.268$^\dagger$ & 0.661 & 423{,}707 & 0.8434 \\
\midrule
\multicolumn{9}{l}{\emph{Accuracy in $[0.90, 0.92)$}} \\
30 & base (wide, $d=128$) & adaptive cw. & 0.918 & 0.360 & 1.608 & 0.619 & 269{,}174 & 0.8134 \\
31 & gated (last) & gated attn. & 0.918 & 0.365 & 1.354 & 0.757 & 62{,}260 & 0.8048 \\
32 & base & class-wise conj. & 0.916 & 0.383 & 2.478 & 0.719 & 57{,}580 & -- \\
33 & gated (max) & softmax conj. & 0.914 & 0.372 & 1.884 & 0.776 & 61{,}741 & -- \\
34 & input-gated & class-wise conj. & 0.914 & 0.398 & 2.481 & 0.715 & 58{,}092 & -- \\
35 & gated (mean) & softmax conj. & 0.912 & 0.368 & 2.093 & 0.765 & 61{,}741 & -- \\
36 & reconstruction-residual & class-wise conj. & 0.910 & 0.417 & 2.504 & 0.680 & 62{,}925 & -- \\
37 & bottleneck & Top-$k$ conj. ($\kappa{=}1.00$) & 0.908 & 0.395 & 2.436 & 0.722 & 41{,}324 & -- \\
38 & gated (max) & adaptive cw. & 0.906 & 0.444 & 2.071 & 0.729 & 61{,}750 & -- \\
39 & input-gated & adaptive cw.$^\ast$ & 0.905 & 0.390 & 2.651 & 0.732 & 58{,}102 & 0.8153 \\
40 & spike/trend & class-wise conj. & 0.904 & 0.395 & 2.155 & 0.699 & 67{,}020 & -- \\
41 & bottleneck & class-wise conj. & 0.902 & 0.395 & 2.572 & 0.733 & 41{,}324 & -- \\
42 & gated (mean) & class-wise conj. & 0.902 & 0.393 & 2.493 & 0.714 & 61{,}740 & -- \\
\midrule
\multicolumn{9}{l}{\emph{Accuracy in $[0.85, 0.90)$}} \\
43 & input-gated & softmax conj. & 0.894 & 0.454 & 1.636 & 0.696 & 58{,}093 & -- \\
44 & base & gated attn. & 0.894 & 0.403 & 2.296 & 0.714 & 58{,}100 & -- \\
45 & reconstruction-residual & attn-max & 0.892 & 0.499 & 2.324 & 0.725 & 62{,}935 & -- \\
46 & base & softmax conj. & 0.890 & 0.419 & 1.832 & 0.705 & 57{,}581 & -- \\
47 & gated & attention$^{\mathrm{M}}$ & 0.888 & 0.426 & 2.403 & 0.710 & 58{,}100 & -- \\
48 & bottleneck & Top-$k$ conj. ($\kappa{=}0.05$) & 0.888 & 0.401 & 2.288 & 0.715 & 41{,}324 & -- \\
49 & InceptionTime & conj.$^{\mathrm{M}\ast}$ & 0.887 & 0.507 & 2.569 & 0.677 & 423{,}707 & 0.8274 \\
50 & gated (max) & attn-max & 0.886 & 0.588 & 1.124 & 0.575 & 61{,}750 & -- \\
51 & bottleneck & Top-$k$ conj. ($\kappa{=}0.50$) & 0.882 & 0.418 & 3.114 & 0.732 & 41{,}324 & -- \\
52 & base (wide, $d=128$) & softmax conj. & 0.880 & 0.455 & 1.015 & 0.605 & 269{,}165 & 0.8131 \\
53 & spike/trend & softmax conj. & 0.878 & 0.488 & 1.435 & 0.665 & 67{,}021 & -- \\
54 & bottleneck & adaptive cw. & 0.868 & 0.559 & 1.969 & 0.680 & 41{,}334 & -- \\
55 & base & dual-stream & 0.860 & 0.498 & 2.399 & 0.745 & 58{,}101 & -- \\
56 & gated (mean) & adaptive cw. & 0.860 & 0.549 & 2.238 & 0.731 & 61{,}750 & -- \\
57 & input-gated & attn-max & 0.858 & 0.625 & 2.153 & 0.700 & 58{,}102 & -- \\
58 & spike/trend & attn-max & 0.854 & 0.637 & 1.646 & 0.644 & 67{,}030 & -- \\
59 & gated (last) & attn-max & 0.854 & 0.598 & 1.779 & 0.668 & 61{,}750 & -- \\
\midrule
\multicolumn{9}{l}{\emph{Accuracy $< 0.85$}} \\
60 & bottleneck (shallow) & Top-$k$ conj. & 0.846 & 0.579 & 2.817 & 0.624 & 21{,}548 & -- \\
61 & gated (mean) & attn-max & 0.832 & 0.665 & 1.672 & 0.704 & 61{,}750 & -- \\
62 & MS pyramid (gated base) & Top-$k$ conj. & 0.832 & 0.602 & 0.980 & 0.591 & 62{,}781 & -- \\
63 & bottleneck & dual-stream & 0.824 & 0.638 & 2.037 & 0.652 & 41{,}845 & -- \\
64 & bottleneck & attn-max & 0.816 & 0.732 & 1.737 & 0.610 & 41{,}334 & -- \\
65 & MS pyramid (bottleneck base) & Top-$k$ conj. & 0.786 & 0.759 & 1.035 & 0.569 & 42{,}365 & -- \\
66 & ResNet & conj.$^{\mathrm{M}}$ & 0.772 & 0.824 & 2.952 & 0.554 & 506{,}331 & 0.8146 \\
67 & FCN & conj.$^{\mathrm{M}}$ & 0.742 & 0.910 & 3.826 & 0.533 & 267{,}035 & 0.8141 \\
68 & gated (last) & dual-stream & 0.736 & 0.804 & 1.947 & 0.637 & 62{,}261 & -- \\
69 & base & attn-max & 0.714 & 0.855 & 1.630 & 0.532 & 57{,}590 & -- \\
70 & MS pyramid (bottleneck base, over channels) & Top-$k$ conj. & 0.696 & 0.992 & 2.167 & 0.653 & 28{,}441 & -- \\
71 & spike/trend & dual-stream & 0.556 & 1.260 & 1.643 & 0.510 & 67{,}541 & -- \\
\end{longtable}
\endgroup

% ---------------------------------------------------------------------
\subsection{Model Cost}
\label{app:cost_all}
\label{sec:res:cost}

The main text states that our model is $10.3\times$ smaller and needs $11.1\times$ less computation than \millet{}. \Cref{tab:cost} gives the full cost and separates what improved from what did not. At 32-bit precision the weights go from about $1.6$\,MB to $161$\,KB, which moves the model from above to well below the roughly 1\,MB flash budget of the micro-controller class described by \citet{lin2020mcunet} and \citet{banbury2021micronets}, and standard \emph{int8} quantization \citep{jacob2018quantization} would reduce it to about $40$\,KB. In this sense, the model becomes a candidate for such a device.
%\begin{table}[!tbp]
%\caption{Cost, measured at batch 32, length 1008. Weight memory is \emph{calculated} from the parameter count (4\,B and 1\,B per weight); nothing was quantised or run on a device. Latency and peak memory are GPU measurements, not estimates for a device.}
%\label{tab:cost}
%	\begin{center}
	%	\small
	%	\setlength{\tabcolsep}{5pt}
	%		\begin{tabular}{lrrrrrr}
		%			\toprule
		%			\textbf{Model} & Params & \multicolumn{2}{c}{Weights (KB)} & MFLOPs & Lat. & Mem. \\
		%			\cmidrule(lr){3-4}
		%			 & & fp32 & int8 & & (ms) & (MB) \\
		%			\midrule
		%			\millet{}               & 423{,}707 & 1655 & 414 & 847.6 & 0.203 & 112.3 \\
		%			\seanet{} wide          & 269{,}164 & 1051 & 263 & 523.7 & 0.359 & 123.7 \\
		%			\seanet{} gated         &  61{,}740 &  241 &  60 & 109.7 & 0.128 &  64.9 \\
		%			\textbf{\seanet{} bottl.} & \textbf{41{,}324} & \textbf{161} & \textbf{40} & \textbf{76.7} & 0.131 & 179.0 \\
		%			\seanet{} bottl., 2 blocks & 21{,}548 &   84 &  21 &  40.0 & 0.069 & 178.9 \\
		%			\bottomrule
		%		\end{tabular}
	%	\end{center}
%\end{table}
The other two columns do not improve. Latency improves by only $1.6\times$ even though the FLOPs fall $11.1\times$, because depthwise convolutions are limited by memory speed, and a GPU at batch 32 is the wrong tool for a claim about small hardware; the $269$\,K wide model is in fact the \emph{slowest} row. Peak memory is $179.0$\,MB against $112.3$\,MB for the baseline, which is worse, and this comes from the design: the encoder keeps the length unchanged, so the activations have the shape $(\dmodel, \nsteps)$ at every stage, however few weights produce them, and \Cref{eq:bottleneck} reduces the weights, not the activations. On micro-controllers, the SRAM needed for the activations is the limit at least as often as the flash needed for the weights, so this is a real limitation of the design and not a measurement error (\Cref{sec:disc}).

% The cost table is copied here from tables/01_main/tab03_cost/tab03_cost_ver1.tex
% (that file is left unchanged), with the row of MILLET trained with its
% published recipe in our code switched off.
% \input{tables/01_main/tab03_cost/tab03_cost_ver1}
\begin{table}[tbp]
\caption{Cost on the WebTraffic input (length 1{,}008, 10 classes), one GPU, batch 32. \emph{Params}$\times$ and \emph{FLOPs}$\times$ are size and compute relative to the re-trained MILLET baseline; column definitions in Appendix~\ref{app:metrics}; lower is better throughout.}
\label{tab:cost}
\begin{center}
\footnotesize
\setlength{\tabcolsep}{6pt}
\renewcommand{\arraystretch}{1.15}
\ifdefined\seatablebox\else\newsavebox\seatablebox\fi
\sbox\seatablebox{\begin{tabular}{lcccccc}
\toprule
\textbf{Model} & \textbf{Params} & \textbf{Size} & \textbf{FLOPs} & \textbf{Latency} & \textbf{Params}$\times$ & \textbf{FLOPs}$\times$ \\
\midrule
\multicolumn{7}{l}{\emph{SEA-Net (ours)}} \\
bottleneck + Top-$k$ conj. & \textbf{41{,}324} & \textbf{1.43} & \textbf{76.7} & 0.131 & 0.10$\times$ & 0.09$\times$ \\
bottleneck + Top-$k$ conj. ($\lambda_{\text{focus}}=0$) & \textbf{41{,}324} & \textbf{1.43} & \textbf{76.7} & 0.130 & 0.10$\times$ & 0.09$\times$ \\
input-gated + Top-$k$ conj. & 58{,}092 & 1.49 & 110.6 & 0.130 & 0.14$\times$ & 0.13$\times$ \\
gated (mean) + Top-$k$ conj. & 61{,}740 & 1.50 & 109.7 & 0.128 & 0.15$\times$ & 0.13$\times$ \\
spike/trend + Top-$k$ conj. & 67{,}020 & 1.53 & 120.1 & 0.132 & 0.16$\times$ & 0.14$\times$ \\
bottleneck + Top-$k$ conj., 2026-09 re-run & \textbf{41{,}324} & \textbf{1.43} & \textbf{76.7} & 0.131 & 0.10$\times$ & 0.09$\times$ \\
\midrule
\multicolumn{7}{l}{\emph{Baselines}} \\
MILLET (InceptionTime + conjunctive), our recipe & 423{,}707 & 4.11 & 847.6 & 0.203 & 1.00$\times$ & 1.00$\times$ \\
% hidden: MILLET trained with its published recipe in our code (not reported)
% MILLET, their recipe (1500 epochs, no early stopping) & 423{,}707 & 4.11 & 847.6 & 0.202 & 1.00$\times$ & 1.00$\times$ \\
ResNet + conjunctive & 506{,}331 & 4.42 & 1013.2 & 0.139 & 1.20$\times$ & 1.20$\times$ \\
FCN + conjunctive & 267{,}035 & 3.47 & 535.2 & \textbf{0.068} & 0.63$\times$ & 0.63$\times$ \\
\bottomrule
\end{tabular}}
\ifdim\wd\seatablebox>\linewidth
  \resizebox{\linewidth}{!}{\usebox\seatablebox}
\else
  \usebox\seatablebox
\fi
\end{center}
\end{table}

\begin{figure}[!htbp]
  \centering
  \includegraphics[width=\linewidth]{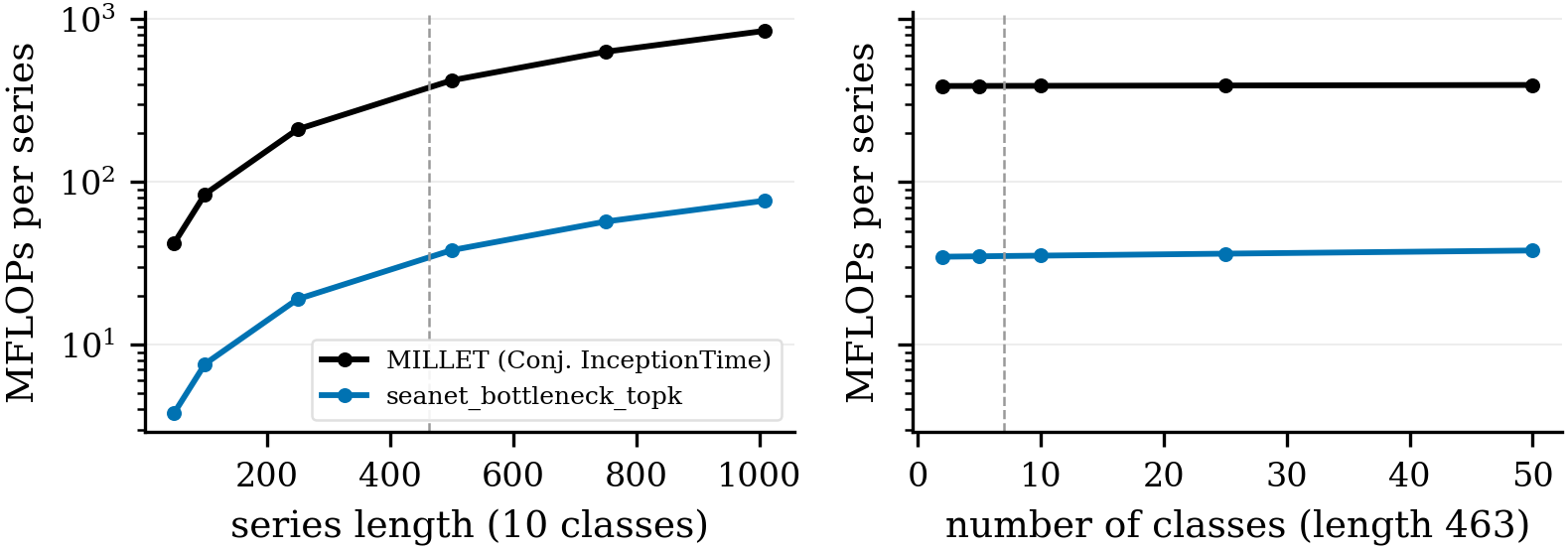}
  \caption{Compute cost of one prediction, in millions of floating point operations, for our
  \seanet{} bottleneck $+$ Top-$k$ model and for the \millet{} baseline (\texttt{InceptionTime} $+$
  \texttt{Conjunctive}). \emph{Left:} as the length of the series grows, at $10$ classes.
  \emph{Right:} as the number of classes grows, at length $463$. The dashed lines mark the mean
  series length and the mean number of classes over the $85$ \ucr{} datasets.}
  \label{fig:time_complexity}
\end{figure}

\Cref{fig:time_complexity} draws the compute column of \Cref{tab:cost} as a function of the size of the problem. The left panel varies the length of the series at $10$ classes, and the right panel varies the number of classes at length $463$, and the two dashed lines mark the mean length and the mean number of classes over the $85$ \ucr{} datasets. The cost of both models grows linearly with the length and hardly depends on the number of classes, and our model needs about ten times fewer operations at every length.

\Cref{tab:complexity} completes the picture on the pooling side, by counting the real multiplications of each head as a function of $t$, $C$ and $d$. No head adds a term that grows faster than those of \texttt{Conjunctive} pooling, so the saving of \Cref{tab:cost} comes from the encoder and is not lost again in the head.

\begin{table}[!htbp]
\caption{Time complexity of the pooling heads as the number of real multiplications, counted from the code (matrix products only, as in Table A.15 of \millet{}; additions, comparisons, the exponential of a softmax and the gather of the critical time point are not counted). $t$ = series length, $c$ = classes, $d$ = embedding size, $a$ = attention width, $k = \lceil \kappa t \rceil$. The last column evaluates the formula for \webtraffic{} ($t = 1008$, $c = 10$, $a = 8$, $\kappa = 0.1$, so $k = 101$) with the own $d$ of each head: the \seanet{} heads read $d = 64$ embeddings and the \millet{} head $d = 128$, so every \seanet{} head is cheaper than the baseline head in absolute terms, although its formula has more terms. $^{\mathrm{M}}$ = \millet{}.}
\label{tab:complexity}
\centering
\seatab\setlength{\tabcolsep}{3pt}
\begin{tabular}{r L{2.35cm}L{3.3cm}L{2.75cm}L{2.2cm}C{1.55cm}}
\toprule
\textbf{\#} & \textbf{Pooling head} & \textbf{Real multiplications} & \textbf{Scale w.r.t.\ $t$} & \textbf{Scale w.r.t.\ $c$} & \textbf{At WebTraffic (M)} \\
\midrule
1 & conjunctive$^{\mathrm{M}}$ & $dtc + dta + (a + 2c)\,t$ & $dc + da + a + 2c$ & $dt + 2t$ & 2.351 ($d = 128$) \\
2 & class-wise conjunctive & $dtc + dta + (ac + 2c)\,t$ & $dc + da + ac + 2c$ & $dt + at + 2t$ & 1.262 ($d = 64$) \\
3 & Top-$k$ conjunctive & $dtc + dta + (ac + c)\,t + ck$ & $dc + da + ac + c + c\kappa$ & $dt + at + t + k$ & 1.253 ($d = 64$) \\
4 & softmax conjunctive & $dtc + dta + (ac + 3c)\,t$ & $dc + da + ac + 3c$ & $dt + at + 3t$ & 1.272 ($d = 64$) \\
5 & adaptive class-wise & $dtc + dta + (ac + 4c)\,t$ & $dc + da + ac + 4c$ & $dt + at + 4t$ & 1.282 ($d = 64$) \\
6 & attention-max & $dtc + dta + (ac + 2c)\,t + 2c$ & $dc + da + ac + 2c$ & $dt + at + 2t + 2$ & 1.262 ($d = 64$) \\
7 & gated attention & $dtc + 2dta + (ac + a + 2c)\,t$ & $dc + 2da + ac + a + 2c$ & $dt + at + 2t$ & 1.786 ($d = 64$) \\
8 & dual-stream conjunctive & $dtc + 2dta + (2ac + 4c)\,t + 2c$ & $dc + 2da + 2ac + 4c$ & $dt + 2at + 4t + 2$ & 1.879 ($d = 64$) \\
\bottomrule
\end{tabular}
\end{table}

% ---------------------------------------------------------------------
\subsection{Ablation Studies}
\label{app:ablations_all}

The main text answers RQ4 and RQ5 with matched pairs of models, i.e. two models that differ in one part only. The ablations below give the wider view: the effect of the pooling head and of the encoder averaged over the whole sweep (A and B), the effect of the Top-$k$ fraction $\kappa$ (C), and the effect of the attention entropy term (D). A short look at the wrong predictions of the two models closes the subsection.

\paragraph{A. Effect of the pooling head, averaged over the encoders.}
In \Cref{tab:ablation_pooling}, one row is one pooling head, and the columns are the accuracy, AOPCR and \ndcg{} averaged over every encoder that the head was paired with, where $n$ is the number of pairings. Averaged this way, the pooling head alone changes the \webtraffic{} accuracy by almost $18$ points, which is why we treat the choice of the head as a design decision and not as a detail.

\begin{table}[!htbp]
\caption{Effect of the pooling head: \webtraffic{} accuracy, AOPCR and \ndcg{} averaged over every encoder that each head was paired with in the 71-model sweep ($n$ = number of pairings; the $\kappa$ / $\lambda$ ablation copies are excluded, so that no pairing is counted twice). The last column is the best single pairing of that head. A head paired once ($n = 1$) is not comparable with one averaged over ten.}
\label{tab:ablation_pooling}
\centering
\seatab\setlength{\tabcolsep}{3pt}
\begin{tabular}{r L{2.9cm}C{1.15cm}C{1.4cm}C{1.4cm}C{0.45cm}L{4.35cm}}
\toprule
\textbf{\#} & \textbf{Pooling head} & \textbf{Mean acc.} & \textbf{Mean AOPCR} & \textbf{Mean NDCG@$n$} & \textbf{$n$} & \textbf{Best pairing (acc.)} \\
\midrule
\multicolumn{7}{l}{\emph{SEA-Net heads}} \\
1 & class-wise conjunctive & 0.921 & 2.322 & 0.692 & 10 & 0.946 (\texttt{seanet\_\allowbreak classwise}) \\
2 & gated attention & 0.918 & 1.869 & \textbf{0.733} & 4 & 0.930 (\texttt{seanet\_\allowbreak bottleneck\_\allowbreak gatedattn}) \\
3 & adaptive class-wise & 0.912 & 2.210 & 0.732 & 9 & 0.946 (\texttt{seanet\_\allowbreak recon\_\allowbreak adaptive}) \\
4 & softmax conjunctive & 0.907 & 1.690 & 0.712 & 9 & 0.940 (\texttt{seanet\_\allowbreak bottleneck\_\allowbreak softmax}) \\
5 & Top-$k$ conjunctive & 0.903 & 2.260 & 0.712 & 16 & 0.955 (\texttt{seanet\_\allowbreak gated\_\allowbreak mean\_\allowbreak topk}) \\
6 & attention-max & 0.838 & 1.758 & 0.645 & 8 & 0.892 (\texttt{seanet\_\allowbreak recon\_\allowbreak attnmax}) \\
7 & dual-stream conjunctive & 0.744 & 2.007 & 0.636 & 4 & 0.860 (\texttt{seanet\_\allowbreak slim\_\allowbreak dualstream}) \\
\midrule
\multicolumn{7}{l}{\emph{MILLET heads (reference; $n = 1$ rows are one pairing only)}} \\
8 & additive & \textbf{0.942} & 1.579 & 0.729 & 1 & 0.942 (\texttt{seanet}) \\
9 & attention & 0.888 & 2.403 & 0.710 & 1 & 0.888 (\texttt{seanet\_\allowbreak slim\_\allowbreak gatedattn}) \\
10 & conjunctive & 0.839 & \textbf{2.712} & 0.615 & 4 & 0.954 (\texttt{seanet\_\allowbreak conjunctive}) \\
\bottomrule
\end{tabular}
\end{table}

\paragraph{B. Effect of the encoder, averaged over the pooling heads.}
\Cref{tab:ablation_encoder} is the same table for the encoders, i.e. one row is one encoder, and the values are averaged over every head that it was paired with. Its spread, from $0.937$ down to $0.771$, is close to the spread of the pooling table above, so the two choices are about equally important. Two cautions apply to both tables. First, several groups contain very few models and not every pair was trained, so the grid of \Cref{tab:grid_a} is the more reliable view. Second, the mean of the bottleneck encoder ($0.884$) is lower than its best pairing suggests, because it was also paired with the weak heads.

\begin{table}[!htbp]
\caption{Effect of the encoder: \webtraffic{} accuracy, AOPCR and \ndcg{} averaged over every pooling head that each encoder was paired with ($n$ = number of pairings, with the same exclusions as in \Cref{tab:ablation_pooling}). The gated encoder has the largest $n$ because it was paired with the most heads, which pulls its mean towards the weak heads; the last column shows what each encoder reaches with its best head.}
\label{tab:ablation_encoder}
\centering
\seatab\setlength{\tabcolsep}{3pt}
\begin{tabular}{r L{2.9cm}C{1.15cm}C{1.4cm}C{1.4cm}C{0.45cm}L{4.35cm}}
\toprule
\textbf{\#} & \textbf{Encoder} & \textbf{Mean acc.} & \textbf{Mean AOPCR} & \textbf{Mean NDCG@$n$} & \textbf{$n$} & \textbf{Best pairing (acc.)} \\
\midrule
\multicolumn{7}{l}{\emph{SEA-Net encoders ($\dagger$ = a wrapper around another encoder)}} \\
1 & SEA-Net multi-scale channels$^\dagger$ & \textbf{0.936} & 2.385 & \textbf{0.747} & 4 & 0.948 (\texttt{seanet\_\allowbreak inputgate\_\allowbreak mschan}) \\
2 & SEA-Net reconstruction-residual & 0.924 & 2.315 & 0.719 & 5 & 0.946 (\texttt{seanet\_\allowbreak recon\_\allowbreak adaptive}) \\
3 & SEA-Net input-gated & 0.903 & 2.345 & 0.718 & 5 & 0.946 (\texttt{seanet\_\allowbreak inputgate\_\allowbreak topk}) \\
4 & SEA-Net gated & 0.902 & 2.056 & 0.710 & 19 & 0.955 (\texttt{seanet\_\allowbreak gated\_\allowbreak mean\_\allowbreak topk}) \\
5 & SEA-Net base & 0.898 & 1.933 & 0.693 & 12 & 0.954 (\texttt{seanet\_\allowbreak conjunctive}) \\
6 & SEA-Net bottleneck & 0.882 & 2.193 & 0.693 & 8 & 0.940 (\texttt{seanet\_\allowbreak bottleneck\_\allowbreak softmax}) \\
7 & SEA-Net spike/trend & 0.858 & 1.814 & 0.677 & 7 & 0.950 (\texttt{seanet\_\allowbreak spiketrend\_\allowbreak topk}) \\
8 & SEA-Net multi-scale pyramid$^\dagger$ & 0.771 & 1.394 & 0.604 & 3 & 0.832 (\texttt{seanet\_\allowbreak gated\_\allowbreak pyramid}) \\
\midrule
\multicolumn{7}{l}{\emph{Baseline encoders (each paired with conjunctive pooling only)}} \\
9 & InceptionTime & 0.887 & 2.569 & 0.677 & 1 & 0.887 (\texttt{millet}) \\
10 & ResNet & 0.772 & 2.952 & 0.554 & 1 & 0.772 (\texttt{resnet}) \\
11 & FCN & 0.742 & \textbf{3.826} & 0.533 & 1 & 0.742 (\texttt{fcn}) \\
\bottomrule
\end{tabular}
\end{table}

\paragraph{C. Effect of the Top-$k$ fraction $\kappa$.}
\Cref{tab:ablation_topk} is the same model trained five times with only the $\kappa$ of \Cref{eq:topk_pool} changed, all of them on seed $0$, and the column $k$ gives how many time points are actually kept at each fraction. The selection is what helps: $\kappa = 1$, which by \Cref{prop:topk} is exactly the class-wise conjunctive pooling of \Cref{eq:classwise_pool}, scores $0.908$, clearly below the $0.938$--$0.940$ obtained at $\kappa = 0.1$ and $\kappa = 0.25$.

\begin{table}[!htbp]
\caption{Top-$k$ fraction ablation on \webtraffic{}: the bottleneck $+$ Top-$k$ model trained five times with only $\kappa$ changed (seed 0 everywhere, so the $\kappa = 0.10$ row is the seed-0 run of that model, and the last row is its mean over every run). $k = \lceil \kappa t \rceil$ is the number of time points that the series logit averages; $\kappa = 1$ recovers class-wise conjunctive pooling exactly. The best single-seed value of each column is in bold.}
\label{tab:ablation_topk}
\centering
\seatab
\begin{tabular}{r L{2.6cm}C{1.3cm}C{1.95cm}C{1.95cm}C{1.95cm}C{1.95cm}}
\toprule
\textbf{\#} & \textbf{$\kappa$} & \textbf{$k$ at $t = 1008$} & \textbf{Accuracy} & \textbf{NLL} & \textbf{AOPCR} & \textbf{NDCG@$n$} \\
\midrule
1 & 0.05 & 51 & 0.888 & 0.401 & 2.288 & 0.715 \\
2 & 0.10 \emph{(default)} & 101 & 0.888 & 0.391 & 2.409 & \textbf{0.755} \\
3 & 0.25 & 252 & \textbf{0.940} & \textbf{0.277} & 2.648 & 0.733 \\
4 & 0.50 & 504 & 0.882 & 0.418 & \textbf{3.114} & 0.732 \\
5 & 1.00 \emph{(= class-wise conjunctive)} & 1008 & 0.908 & 0.395 & 2.436 & 0.722 \\
\midrule
 & 0.10, mean $\pm$ std over 3 runs & 101 & $0.931 \pm 0.032$ & $0.312 \pm 0.063$ & $2.498 \pm 0.149$ & $0.765 \pm 0.013$ \\
\bottomrule
\end{tabular}
\end{table}

\paragraph{D. Effect of the attention entropy term.}
\Cref{tab:ablation_focus} is one matched pair, i.e.\ the same model with only $\lambda_{\text{focus}}$ switched on and off, on seed $0$. Switching the term off raises the accuracy from $0.888$ to $0.950$ and \ndcg{} from $0.755$ to $0.765$, and lowers AOPCR from $2.409$ to $2.303$. Seed $0$ is, however, the weakest of the three runs of the default model ($0.888$, $0.938$ and $0.966$; \Cref{tab:seeds}): against its three-run mean ($0.931 \pm 0.032$ accuracy, $0.765 \pm 0.013$ \ndcg{}), the run without the term lies within one standard deviation in accuracy and matches it exactly in \ndcg{}. We therefore keep the value selected by the hyperparameter search and read this pair as showing no measurable effect of the term in either direction.

\begin{table}[!htbp]
\caption{The attention-entropy term of the loss, switched on and off, on \webtraffic{} (seed 0): the bottleneck $+$ Top-$k$ model with only $\lambda_{\text{focus}}$ changed. This is one matched pair on one seed, so it indicates a direction rather than an effect size.}
\label{tab:ablation_focus}
\centering
\seatab
\begin{tabular}{L{3.6cm}C{1.8cm}C{1.8cm}C{1.8cm}C{1.8cm}}
\toprule
\textbf{$\lambda_{\text{focus}}$} & \textbf{Accuracy} & \textbf{NLL} & \textbf{AOPCR} & \textbf{NDCG@$n$} \\
\midrule
0.01 \emph{(on, the default)} & 0.888 & 0.391 & \textbf{2.409} & 0.755 \\
0.00 \emph{(off)} & \textbf{0.950} & \textbf{0.263} & 2.303 & \textbf{0.765} \\
\bottomrule
\end{tabular}
\end{table}

\begin{figure}[!htbp]
  \centering
  \includegraphics[width=0.95\linewidth]{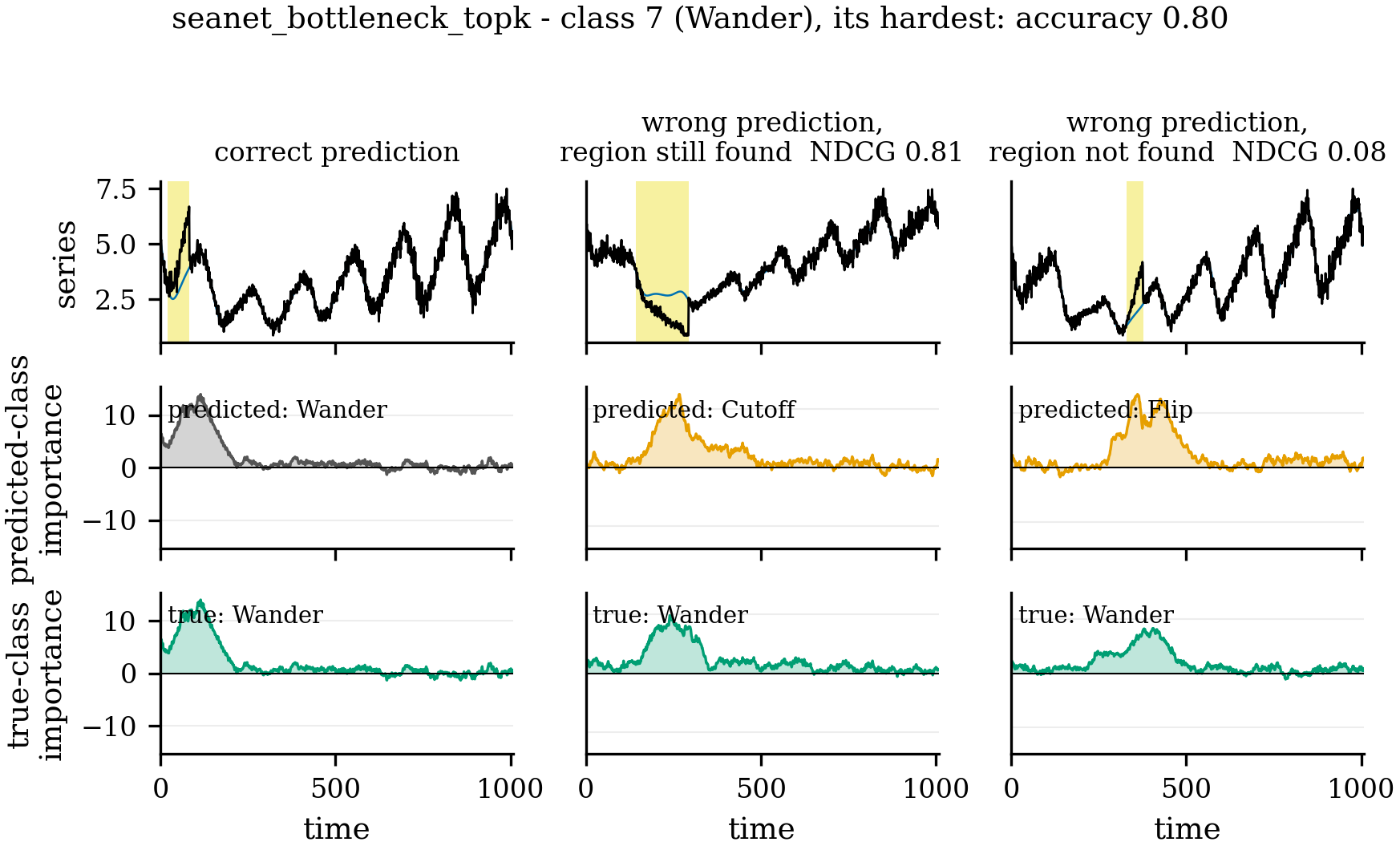}
  \caption{\seanet{} bottleneck $+$ Top-$k$ (ours). Wrong predictions on the hardest \webtraffic{} class, i.e. the class with the lowest accuracy. \emph{Top:} the series with its true signature window in yellow. \emph{Middle:} the interpretation for the class that the model predicted. \emph{Bottom:} the interpretation for the true class, with its \ndcg{}.}
  \label{fig:fn_ours}
\end{figure}

\begin{figure}[!htbp]
  \centering
  \includegraphics[width=0.95\linewidth]{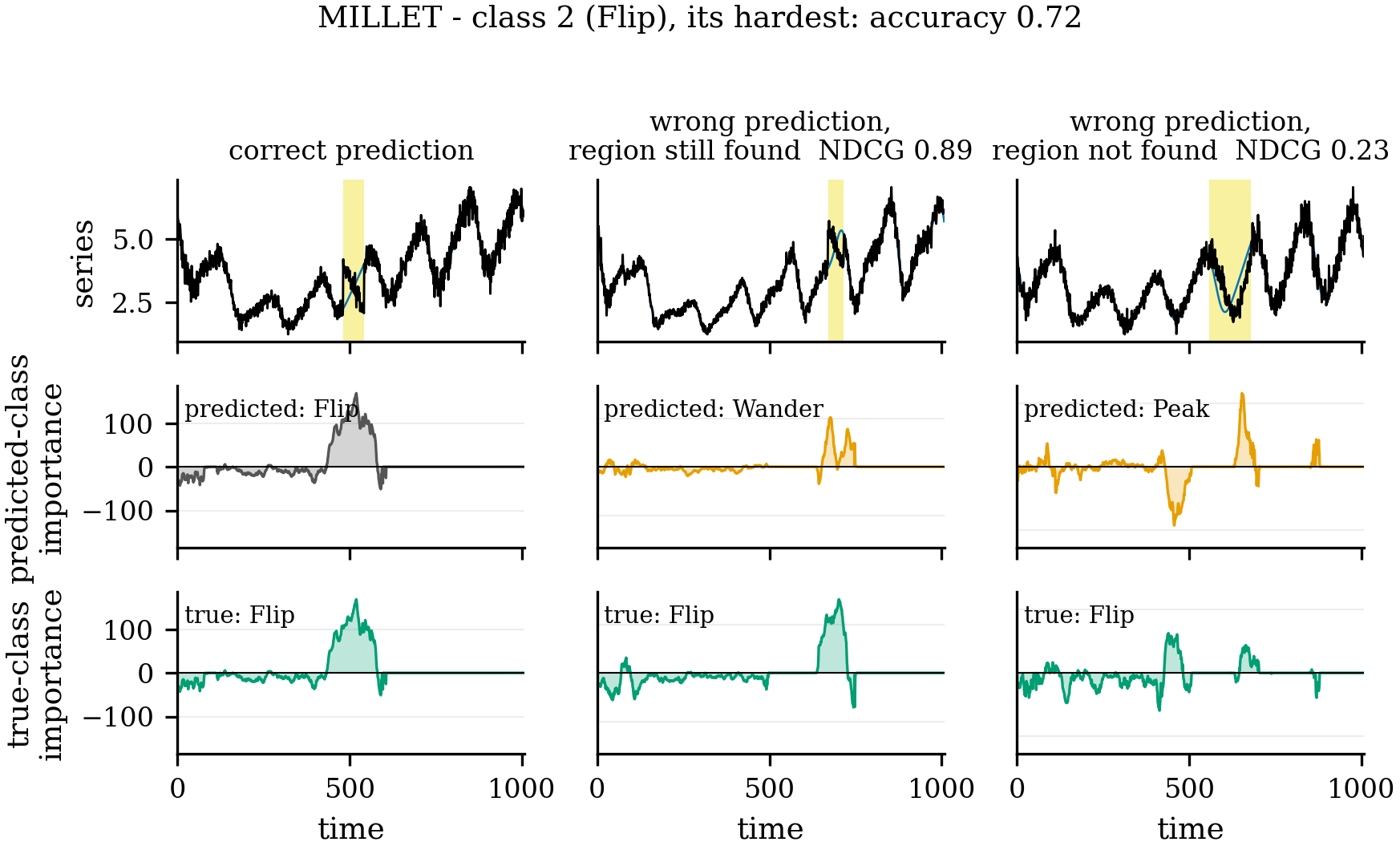}
  \caption{\millet{} (published recipe). Wrong predictions on the hardest \webtraffic{} class, i.e. the class with the lowest accuracy. \emph{Top:} the series with its true signature window in yellow. \emph{Middle:} the interpretation for the class that the model predicted. \emph{Bottom:} the interpretation for the true class, with its \ndcg{}.}
  \label{fig:fn_millet}
\end{figure}

\paragraph{Wrong predictions.}
Finally, \Cref{fig:fn_ours,fig:fn_millet} look at the wrong predictions of the two models on their hardest \webtraffic{} class, i.e. the class with the lowest accuracy. The errors are of two kinds: in the first, the model still points at the right window but names the wrong class, so the \ndcg{} stays high; in the second, the signature is too weak or too close to the seasonal pattern, and the model never finds it.

% \begin{figure}[!htbp]
%   \centering
%   \begin{subfigure}[t]{0.495\linewidth}
%     \centering
%     \includegraphics[width=\linewidth]{selected_figures/02_supplementary/figA4_false_negatives/figA4_false_negatives_seanet_bottleneck_topk.png}
%     \caption{\seanet{} bottleneck $+$ Top-$k$ (ours).}
%     \label{fig:fn_ours}
%   \end{subfigure}\hfill
%   \begin{subfigure}[t]{0.495\linewidth}
%     \centering
%     \includegraphics[width=\linewidth]{selected_figures/02_supplementary/figA4_false_negatives/figA4_false_negatives_millet.png}
%     \caption{\millet{} (published recipe).}
%     \label{fig:fn_millet}
%   \end{subfigure}
%   \caption{Wrong predictions of the two models on their hardest \webtraffic{} class, i.e. the class
%   with the lowest accuracy, whose name and accuracy are given in the title of each panel.
%   \emph{Left column:} a correct prediction, for reference. \emph{Middle and right columns:} two
%   wrong predictions. \emph{Top:} the series with its true signature window in yellow.
%   \emph{Middle:} the interpretation for the class that the model predicted. \emph{Bottom:} the
%   interpretation for the true class, with its \ndcg{}.}
%   \label{fig:false_negatives}
% \end{figure}

% ---------------------------------------------------------------------
\subsection{Per-seed Results}
\label{app:seeds_all}
\label{app:seeds}

\begin{figure}[!htbp]
  \centering
  \includegraphics[width=\linewidth]{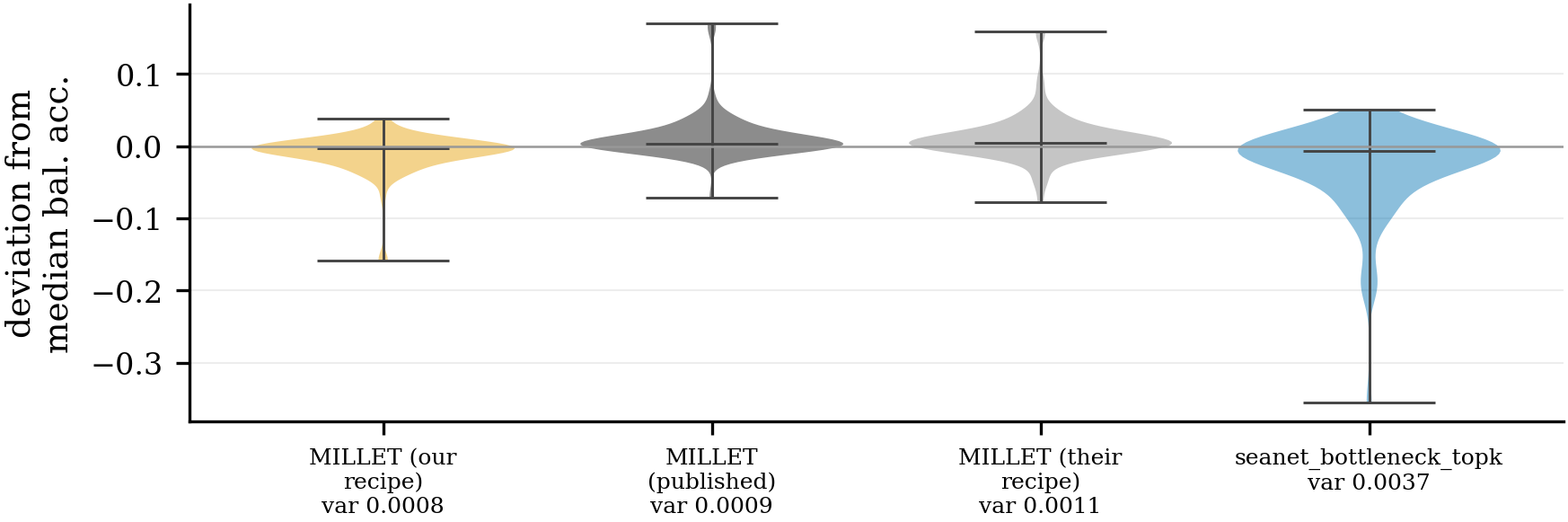}
  \caption{Model variance over the $85$ \ucr{} datasets, in the style of \citet{early2024millet}.
  For each dataset and each seed, the balanced accuracy minus the median balanced accuracy of that
  model on that dataset, pooled over the datasets; the models are ordered from left to right by
  their total variance, which is printed under each of them. Our \seanet{} model is the least stable
  of the four and has a long lower tail, i.e. on a few datasets one seed falls far below the other
  two, which is the same effect that \Cref{tab:seeds} shows on \webtraffic{}.}
  \label{fig:variance}
\end{figure}

The main text compares every difference with the seed-to-seed spread measured on \webtraffic{}, $\pm 0.016$ for the accuracy and $\pm 0.013$ for \ndcg{}. \Cref{tab:seeds} gives the runs behind these numbers: every \webtraffic{} run of the models that were trained more than once, with one row per model and per metric, one column per run, and the mean $\pm$ the standard deviation in the last column. These are the numbers behind every $\pm$ of the main tables and the noise level used throughout the paper: we never report a difference smaller than this spread as a result. For every model, the spread of AOPCR is far larger than that of the accuracy, which is why we never read an AOPCR difference of a few tenths as a win. \Cref{fig:variance} asks the same question on the $85$ \ucr{} datasets.

\begin{table}[!htbp]
\caption{Every \webtraffic{} run of the models that were trained more than once, run by run (\emph{archive} = the 71-model sweep in results/old\_results, \emph{2026-09} = results/top\_results; a seed run in both folders is two different trainings). These are the numbers behind every $\pm$ in the main tables. The spread on AOPCR is far larger than on accuracy for every model, which is why AOPCR differences of a few tenths are not read as wins.}
\label{tab:seeds}
\centering
\seatab\setlength{\tabcolsep}{3pt}
\begin{tabular}{r L{2.8cm}L{1.35cm}C{1.4cm}C{1.4cm}C{1.4cm}C{1.4cm}C{2.1cm}}
\toprule
\textbf{\#} & \textbf{Model} & \textbf{Metric} & \textbf{Run 1} & \textbf{Run 2} & \textbf{Run 3} & \textbf{Run 4} & \textbf{Mean $\pm$ std} \\
\midrule
1 & SEA-Net gated (mean) + Top-$k$ conjunctive & accuracy & 0.954 & 0.958 & 0.952 & -- & $0.955 \pm 0.002$ \\
 &  & AOPCR & 2.293 & 2.074 & 2.308 & -- & $2.225 \pm 0.107$ \\
 &  & NDCG@$n$ & 0.772 & 0.749 & 0.728 & -- & $0.750 \pm 0.018$ \\
 &  & \emph{run} & \emph{archive, seed 0} & \emph{archive, seed 1} & \emph{archive, seed 2} &  &  \\
\midrule
2 & SEA-Net bottleneck + Top-$k$ conjunctive & accuracy & 0.888 & 0.938 & 0.966 & -- & $0.931 \pm 0.032$ \\
 &  & AOPCR & 2.409 & 2.708 & 2.378 & -- & $2.498 \pm 0.149$ \\
 &  & NDCG@$n$ & 0.755 & 0.783 & 0.758 & -- & $0.765 \pm 0.013$ \\
 &  & \emph{run} & \emph{archive, seed 0} & \emph{archive, seed 1} & \emph{archive, seed 2} &  &  \\
\midrule
% hidden: MILLET trained with its published recipe in our code (not reported)
% 3 & MILLET (InceptionTime + conjunctive), their recipe & accuracy & 0.920 & 0.916 & -- & -- & $0.918 \pm 0.002$ \\
%  &  & AOPCR & 13.268 & 10.852 & -- & -- & $12.060 \pm 1.208$ \\
%  &  & NDCG@$n$ & 0.661 & 0.624 & -- & -- & $0.643 \pm 0.019$ \\
%  &  & \emph{run} & \emph{archive, seed 0} & \emph{2026-09, seed 0} &  &  &  \\
% \midrule
3 & SEA-Net input-gated + adaptive class-wise & accuracy & 0.938 & 0.880 & 0.898 & -- & $0.905 \pm 0.024$ \\
 &  & AOPCR & 3.126 & 2.266 & 2.560 & -- & $2.651 \pm 0.357$ \\
 &  & NDCG@$n$ & 0.777 & 0.692 & 0.728 & -- & $0.732 \pm 0.035$ \\
 &  & \emph{run} & \emph{archive, seed 0} & \emph{archive, seed 1} & \emph{archive, seed 2} &  &  \\
\midrule
4 & MILLET (InceptionTime + conjunctive), our recipe & accuracy & 0.894 & 0.876 & 0.892 & 0.890 & $0.888 \pm 0.007$ \\
 &  & AOPCR & 1.594 & 2.795 & 3.319 & 2.120 & $2.457 \pm 0.655$ \\
 &  & NDCG@$n$ & 0.671 & 0.695 & 0.665 & 0.692 & $0.681 \pm 0.013$ \\
 &  & \emph{run} & \emph{archive, seed 0} & \emph{archive, seed 1} & \emph{archive, seed 2} & \emph{2026-09, seed 0} &  \\
\bottomrule
\end{tabular}
\end{table}

% ---------------------------------------------------------------------
\FloatBarrier
\subsection{Complete \ucr{} Results}
\label{app:ucr_all}
\label{app:ucr}

On the \ucr{} archive, the main text reports that our model reaches $0.810$ mean accuracy, behind \millet{} with the matched budget ($0.827$) and behind the published \millet{} accuracy ($0.8445$, \citealt{early2024millet}), and that the wide variant of the same design closes this gap. \Cref{tab:ucr85_rank} is the evidence behind these statements (\Cref{sec:res:ucr}). With the training budget fixed, changing the backbone at a similar size moves the mean rank from $12.60$ to $11.53$, about $1.1$ places, so our design is not what loses accuracy; the reduced capacity is.

\begin{table}[!htbp]
\caption{\ucr{}-85 accuracy and mean rank. Mean rank is over the 84 datasets finished by all 28
models we ran on the archive; lower is better. Win/tie/loss is counted per dataset against
\millet{}'s \emph{published} accuracies. $^{\ddagger}$84 of 85 datasets finished.}
\label{tab:ucr85_rank}
\begin{center}
\footnotesize
\setlength{\tabcolsep}{5pt}
\begin{tabular}{lrccc}
\toprule
\textbf{Model} & Params & Acc. & Mean rank & W/T/L \\
\midrule
\millet{}, published \citep{early2024millet}   & 423{,}707 & 0.8445 & --- & --- \\
% hidden: MILLET trained with its published recipe in our code (not reported)
% \millet{}, published recipe, our code          & 423{,}707 & 0.8434$^{\ddagger}$ & \textbf{9.41} & 26/32/26 \\
\seanet{} wide $+$ class-wise conjunctive      & 269{,}164 & 0.8292 & \textbf{11.53} & 26/19/39 \\
\seanet{} wide $+$ \millet{} additive          & 269{,}083 & 0.8298 & 11.98 & 23/19/42 \\
\seanet{} wide $+$ \millet{} conjunctive       & 269{,}083 & 0.8238 & 12.43 & 22/20/42 \\
\millet{}, matched budget                      & 423{,}707 & 0.8274$^{\ddagger}$ & 12.60 & 13/25/46 \\
ResNet $+$ conjunctive                         & 506{,}331 & 0.8146 & 13.54 & 25/11/49 \\
FCN $+$ conjunctive                            & 267{,}035 & 0.8141 & 14.48 & 20/12/52 \\
\seanet{} gated $+$ Top-$k$                    &  61{,}740 & 0.8083 & 15.29 & 19/13/53 \\
\textbf{\seanet{} bottleneck $+$ Top-$k$}      & \textbf{41{,}324} & 0.8097 & 15.40 & 20/15/50 \\
\bottomrule
\end{tabular}
\end{center}
\end{table}

The remaining tables give the full detail. \Cref{tab:ucr85_all} lists every model that completed the archive, sorted by mean accuracy, where $n$ is the number of datasets it finished out of the $85$, and \Cref{tab:ucr85_wtl} counts the wins, ties and losses of the same models against the published per-dataset numbers of \millet{}. \Cref{tab:per_dataset_acc} and \Cref{tab:per_dataset_aopcr} then give one row per dataset and one column per model, for the accuracy and for AOPCR respectively, so that a single dataset can be looked up instead of a mean.

A Wilcoxon--Holm test over the same datasets finds that most of the differences in the middle of \Cref{tab:ucr85_all} are not significant at $\alpha = 0.05$. This is one more reason why the main text argues from \webtraffic{} and from the cost rather than from the \ucr{} rank, and why we give the win / tie / loss counts instead of relying on the rank order alone.

\begin{figure}[!htbp]
  \centering
  \includegraphics[width=0.62\linewidth]{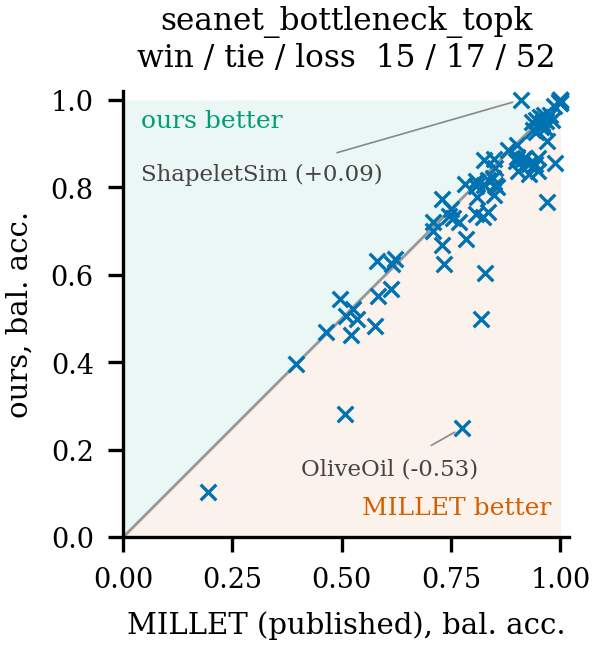}
  \caption{Direct comparison of our \seanet{} bottleneck $+$ Top-$k$ model against the published
  \millet{} accuracies on the $85$ \ucr{} datasets, one point per dataset. The points above the
  diagonal are the datasets where our model is the more accurate one. The title gives the number of
  wins / ties / losses (15 / 17 / 52, where a tie is a difference below $0.005$), and the datasets
  with the largest positive and negative differences are named.}
  \label{fig:direct_comparison}
\end{figure}

\begin{figure}[!htbp]
  \centering
  \includegraphics[width=\linewidth]{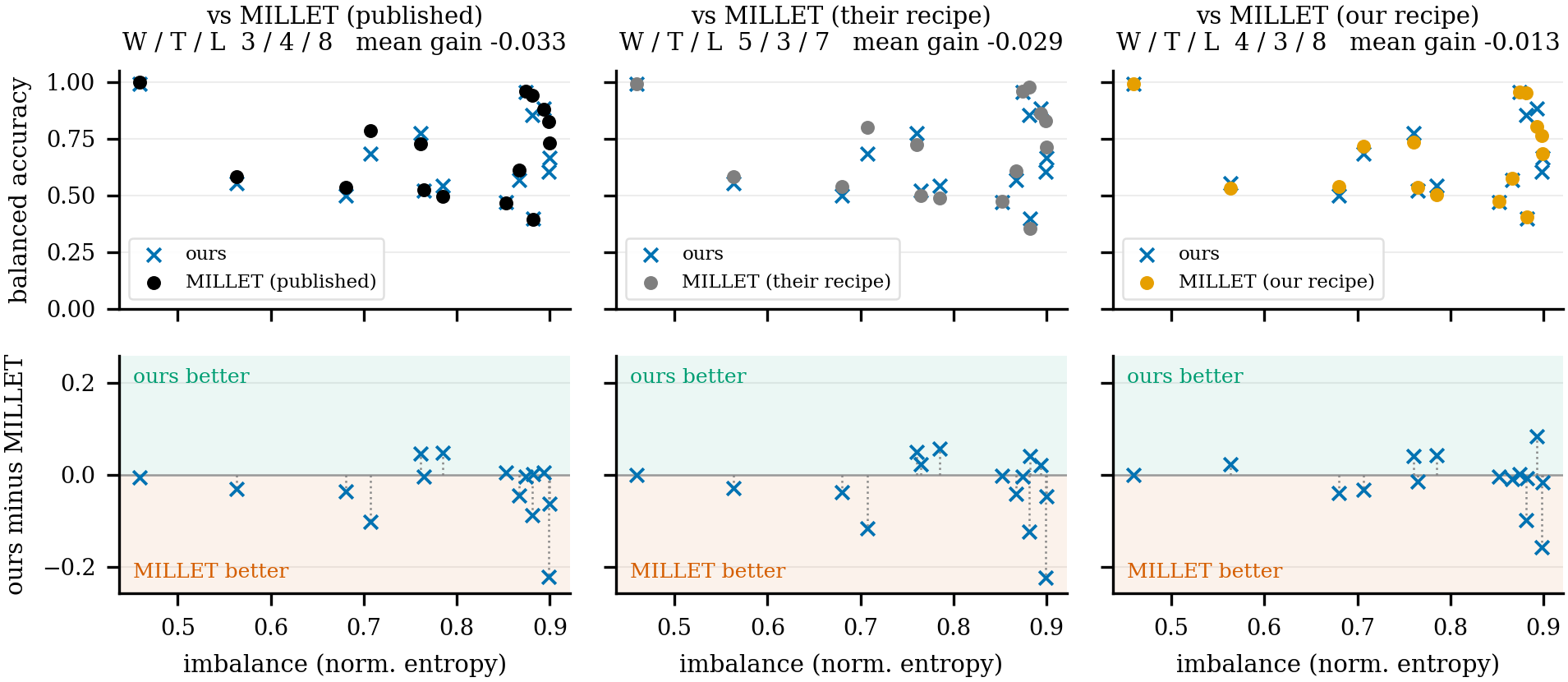}
  \caption{The effect of the class imbalance, on the \ucr{} datasets whose training set is
  imbalanced (normalized class entropy below one). \emph{Top:} the balanced accuracy of our model
  and of \millet{} measured three ways, i.e. published, re-trained with the published recipe, and
  re-trained with the configuration of \Cref{tab:protocol}. \emph{Bottom:} the difference, ours
  minus \millet{}, where green means that our model is the better one. Each title gives the wins /
  ties / losses and the mean gain. The imbalance does not hurt the smaller model more than the
  baseline: the differences are small on both sides, and they do not grow with the imbalance.}
  \label{fig:imbalance}
\end{figure}

\begin{figure}[!htbp]
  \centering
  \includegraphics[width=\linewidth]{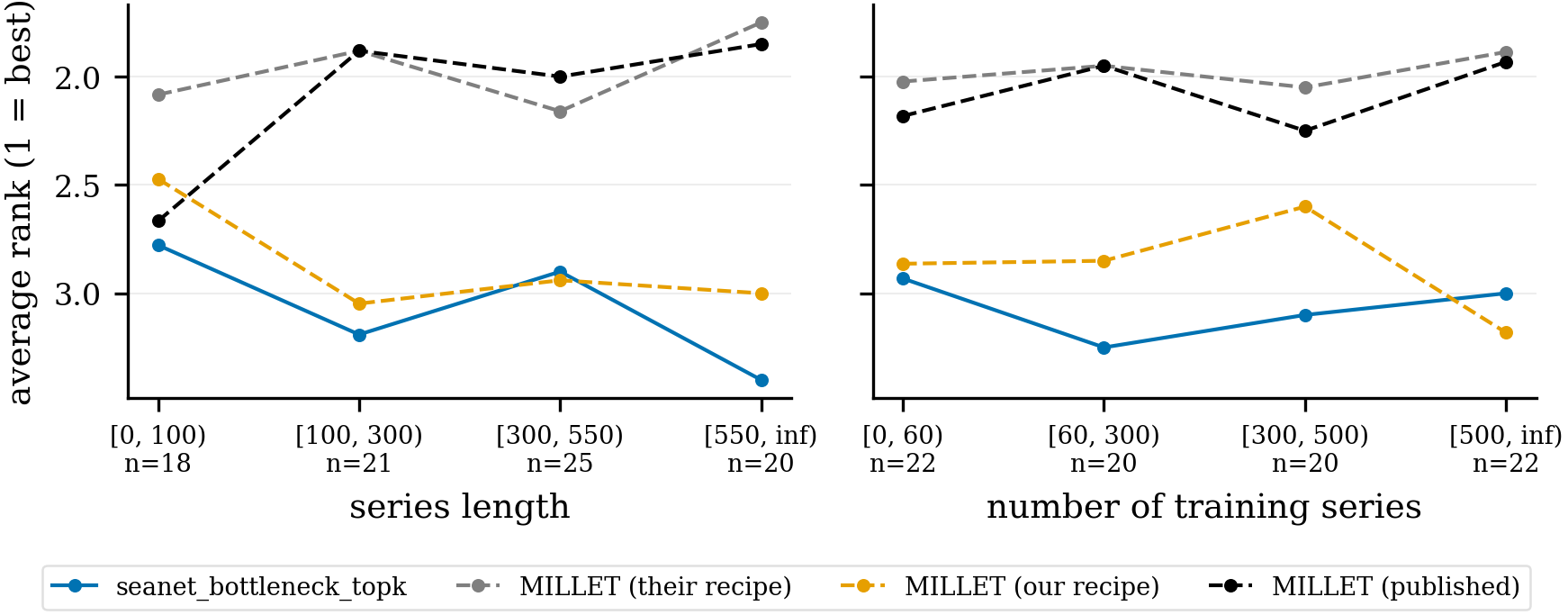}
  \caption{Average accuracy rank ($1 =$ best, lower is better) of our model and of the three
  \millet{} measurements, binned by two properties of the dataset: the length of the series
  (\emph{left}) and the number of training series (\emph{right}). The bin edges are chosen to give
  bins of roughly equal size, and $n$ is the number of datasets in each bin. The ranking is stable
  across the bins, i.e. our \seanet{} model does not fall further behind on the long series or on
  the small training sets.}
  \label{fig:dataset_properties}
\end{figure}

\Cref{fig:direct_comparison} shows the accuracy comparison with one point per dataset, our model against the published \millet{} accuracy \citep{early2024millet}. \Cref{fig:imbalance} then checks, on the \ucr{} datasets whose normalized class entropy is below one, whether an imbalanced training set hurts the smaller model more, and \Cref{fig:dataset_properties} checks the same for the length of the series and for the number of training series. In all three cases the answer is the same: the ranking of the models is stable, so our model does not fall further behind on the imbalanced, the long or the small datasets.

\begingroup
\seatab\setlength{\tabcolsep}{3pt}
\begin{longtable}{r L{4.45cm}C{0.45cm}C{0.95cm}C{0.95cm}C{1.2cm}C{0.95cm}C{1.15cm}R{1.25cm}}
\caption{The 85 \ucr{} datasets: every model that completed the archive, sorted by mean accuracy ($n$ = number of datasets finished, out of the 85 that \millet{} publishes). The means are over the datasets that each model finished; the win / tie / loss counts of the same rows are given in \Cref{tab:ucr85_wtl}. AOPCR is unnormalised: the rows trained with our recipe can be compared with each other, whereas the $^\dagger$ row (the published numbers of \millet{}) is on another scale. The best value of each column is in bold (the $^\dagger$ row is excluded for AOPCR).}
\label{tab:ucr85_all} \\
\toprule
\textbf{\#} & \textbf{Model} & \textbf{$n$} & \textbf{Acc.} & \textbf{Bal.\ acc.} & \textbf{AUROC} & \textbf{NLL} & \textbf{AOPCR} & \textbf{Params} \\
\midrule
\endfirsthead
\multicolumn{9}{l}{\emph{\tablename~\thetable{} (continued)}} \\
\toprule
\textbf{\#} & \textbf{Model} & \textbf{$n$} & \textbf{Acc.} & \textbf{Bal.\ acc.} & \textbf{AUROC} & \textbf{NLL} & \textbf{AOPCR} & \textbf{Params} \\
\midrule
\endhead
\midrule
\multicolumn{9}{r}{\emph{continued on the next page}} \\
\endfoot
\bottomrule
\endlastfoot
% hidden: MILLET trained with its published recipe in our code (not reported)
% 1 & MILLET (InceptionTime + conjunctive), their recipe & 84 & \textbf{0.8434} & \textbf{0.8212} & \textbf{0.9363} & 1.1704 & 3.926$^\dagger$ & 423{,}707 \\
1 & SEA-Net base (wide, $d=128$) + additive (MILLET) & 85 & 0.8298 & 0.8077 & 0.9295 & \textbf{0.5088} & 0.551 & 269{,}083 \\
2 & SEA-Net base (wide, $d=128$) + class-wise conjunctive & 85 & 0.8292 & 0.8039 & 0.9255 & 0.5113 & 0.564 & 269{,}164 \\
3 & MILLET (InceptionTime + conjunctive), our recipe & 84 & 0.8274 & 0.7997 & 0.9275 & 0.5297 & 0.812 & 423{,}707 \\
4 & SEA-Net base (wide, $d=128$) + conjunctive (MILLET) & 85 & 0.8238 & 0.7993 & 0.9281 & 0.5180 & 0.626 & 269{,}083 \\
5 & SEA-Net gated (max) + class-wise conjunctive & 85 & 0.8237 & 0.7979 & 0.9288 & 0.5214 & 0.619 & 61{,}740 \\
6 & SEA-Net input-gated + adaptive class-wise & 85 & 0.8153 & 0.7905 & 0.9208 & 0.5545 & 0.964 & 58{,}102 \\
7 & SEA-Net bottleneck + gated attention & 85 & 0.8146 & 0.7879 & 0.9254 & 0.5330 & 0.688 & 41{,}844 \\
8 & ResNet + conjunctive & 85 & 0.8146 & 0.7839 & 0.9232 & 0.5709 & 1.330 & 506{,}331 \\
9 & FCN + conjunctive & 85 & 0.8141 & 0.7846 & 0.9216 & 0.5727 & \textbf{1.378} & 267{,}035 \\
10 & SEA-Net reconstruction-residual + adaptive class-wise & 85 & 0.8140 & 0.7859 & 0.9232 & 0.5502 & 1.091 & 62{,}935 \\
11 & SEA-Net base (wide, $d=128$) + adaptive class-wise & 85 & 0.8134 & 0.7849 & 0.9199 & 0.5354 & 0.891 & 269{,}174 \\
12 & SEA-Net base (wide, $d=128$) + softmax conjunctive & 85 & 0.8131 & 0.7858 & 0.9212 & 0.5436 & 1.247 & 269{,}165 \\
13 & SEA-Net input-gated + Top-$k$ conjunctive & 85 & 0.8131 & 0.7877 & 0.9223 & 0.5365 & 0.964 & 58{,}092 \\
14 & SEA-Net gated (max) + Top-$k$ conjunctive & 85 & 0.8108 & 0.7850 & 0.9172 & 0.5464 & 0.936 & 61{,}740 \\
15 & SEA-Net bottleneck + softmax conjunctive & 85 & 0.8106 & 0.7856 & 0.9167 & 0.5509 & 0.759 & 41{,}325 \\
16 & SEA-Net bottleneck + Top-$k$ conjunctive & 85 & 0.8097 & 0.7827 & 0.9179 & 0.5506 & 1.091 & 41{,}324 \\
17 & SEA-Net base + Top-$k$ conjunctive & 85 & 0.8093 & 0.7833 & 0.9182 & 0.5525 & 1.018 & 57{,}580 \\
18 & SEA-Net spike/trend + Top-$k$ conjunctive & 85 & 0.8089 & 0.7832 & 0.9137 & 0.5460 & 1.108 & 67{,}020 \\
19 & SEA-Net gated (mean) + Top-$k$ conjunctive & 85 & 0.8083 & 0.7803 & 0.9168 & 0.5560 & 1.064 & 61{,}740 \\
20 & SEA-Net reconstruction-residual + Top-$k$ conjunctive & 85 & 0.8063 & 0.7793 & 0.9156 & 0.5530 & 1.222 & 62{,}925 \\
21 & SEA-Net gated (last) + gated attention & 85 & 0.8048 & 0.7804 & 0.9157 & 0.5564 & 0.500 & 62{,}260 \\
22 & SEA-Net spike/trend + gated attention & 85 & 0.8039 & 0.7793 & 0.9112 & 0.5537 & 0.682 & 67{,}540 \\
23 & SEA-Net gated (last) + adaptive class-wise & 85 & 0.8029 & 0.7797 & 0.9139 & 0.5633 & 0.796 & 61{,}750 \\
24 & SEA-Net gated (last) + Top-$k$ conjunctive & 85 & 0.8015 & 0.7749 & 0.9131 & 0.5531 & 0.967 & 61{,}740 \\
25 & SEA-Net spike/trend + adaptive class-wise & 85 & 0.8010 & 0.7766 & 0.9135 & 0.5641 & 0.677 & 67{,}030 \\
26 & SEA-Net gated + class-wise conjunctive & 85 & 0.7900 & 0.7657 & 0.9055 & 0.5848 & 0.486 & 61{,}740 \\
27 & SEA-Net gated (last) + class-wise conjunctive & 85 & 0.7888 & 0.7658 & 0.9027 & 0.5922 & 0.535 & 61{,}740 \\
\midrule
28 & MILLET conjunctive InceptionTime, published (mean of 5 seeds) & 85 & \textbf{0.8445} & \textbf{0.8235} & \textbf{0.9329} & 1.2241 & 4.553$^\dagger$ & 423{,}300 \\
\end{longtable}
\endgroup

\begingroup
\seatab
\begin{longtable}{r L{6.3cm}C{2.0cm}C{2.0cm}C{2.0cm}}
\caption{Wins / ties / losses (W/T/L) of the models of \Cref{tab:ucr85_all}, with the same rows and the same numbering, against the published per-dataset numbers of \millet{}, with tie bands of 0.005 (accuracy), 0.01 (NLL) and 0.1 (AOPCR). The published AOPCR values come from \millet{}'s recipe and are on another scale, which is why every model ``loses'' the AOPCR W/T/L against them.}
\label{tab:ucr85_wtl} \\
\toprule
\textbf{\#} & \textbf{Model} & \textbf{W/T/L acc.} & \textbf{W/T/L NLL} & \textbf{W/T/L AOPCR} \\
\midrule
\endfirsthead
\multicolumn{5}{l}{\emph{\tablename~\thetable{} (continued)}} \\
\toprule
\textbf{\#} & \textbf{Model} & \textbf{W/T/L acc.} & \textbf{W/T/L NLL} & \textbf{W/T/L AOPCR} \\
\midrule
\endhead
\midrule
\multicolumn{5}{r}{\emph{continued on the next page}} \\
\endfoot
\bottomrule
\endlastfoot
% hidden: MILLET trained with its published recipe in our code (not reported)
% 1 & MILLET (InceptionTime + conjunctive), their recipe & 26/32/26 & 40/14/30 & 42/5/37 \\
1 & SEA-Net base (wide, $d=128$) + additive (MILLET) & 24/19/42 & 60/2/23 & 3/0/82 \\
2 & SEA-Net base (wide, $d=128$) + class-wise conjunctive & 27/19/39 & 60/1/24 & 3/0/82 \\
3 & MILLET (InceptionTime + conjunctive), our recipe & 13/25/46 & 56/2/26 & 3/0/81 \\
4 & SEA-Net base (wide, $d=128$) + conjunctive (MILLET) & 23/20/42 & 57/5/23 & 4/0/81 \\
5 & SEA-Net gated (max) + class-wise conjunctive & 20/17/48 & 61/0/24 & 4/0/81 \\
6 & SEA-Net input-gated + adaptive class-wise & 20/15/50 & 60/0/25 & 4/0/81 \\
7 & SEA-Net bottleneck + gated attention & 23/14/48 & 59/1/25 & 5/0/80 \\
8 & ResNet + conjunctive & 25/11/49 & 55/0/30 & 8/0/77 \\
9 & FCN + conjunctive & 21/12/52 & 55/1/29 & 5/3/77 \\
10 & SEA-Net reconstruction-residual + adaptive class-wise & 19/18/48 & 58/1/26 & 4/0/81 \\
11 & SEA-Net base (wide, $d=128$) + adaptive class-wise & 25/17/43 & 57/1/27 & 4/1/80 \\
12 & SEA-Net base (wide, $d=128$) + softmax conjunctive & 17/13/55 & 60/7/18 & 4/0/81 \\
13 & SEA-Net input-gated + Top-$k$ conjunctive & 23/16/46 & 60/1/24 & 3/1/81 \\
14 & SEA-Net gated (max) + Top-$k$ conjunctive & 22/16/47 & 59/0/26 & 5/1/79 \\
15 & SEA-Net bottleneck + softmax conjunctive & 22/17/46 & 58/1/26 & 3/0/82 \\
16 & SEA-Net bottleneck + Top-$k$ conjunctive & 20/15/50 & 58/1/26 & 4/0/81 \\
17 & SEA-Net base + Top-$k$ conjunctive & 20/11/54 & 60/1/24 & 5/0/80 \\
18 & SEA-Net spike/trend + Top-$k$ conjunctive & 19/23/43 & 56/1/28 & 6/0/79 \\
19 & SEA-Net gated (mean) + Top-$k$ conjunctive & 19/13/53 & 58/1/26 & 4/0/81 \\
20 & SEA-Net reconstruction-residual + Top-$k$ conjunctive & 19/19/47 & 59/0/26 & 5/0/80 \\
21 & SEA-Net gated (last) + gated attention & 17/15/53 & 58/0/27 & 3/0/82 \\
22 & SEA-Net spike/trend + gated attention & 15/14/56 & 58/1/26 & 4/0/81 \\
23 & SEA-Net gated (last) + adaptive class-wise & 18/18/49 & 57/0/28 & 4/1/80 \\
24 & SEA-Net gated (last) + Top-$k$ conjunctive & 17/17/51 & 59/0/26 & 3/1/81 \\
25 & SEA-Net spike/trend + adaptive class-wise & 24/12/49 & 57/0/28 & 4/0/81 \\
26 & SEA-Net gated + class-wise conjunctive & 16/15/54 & 55/1/29 & 3/0/82 \\
27 & SEA-Net gated (last) + class-wise conjunctive & 17/12/56 & 55/0/30 & 3/0/82 \\
\midrule
28 & MILLET conjunctive InceptionTime, published (mean of 5 seeds) & -- & -- & -- \\
\end{longtable}
\endgroup

\begingroup
\seatab\setlength{\tabcolsep}{2pt}\renewcommand{\arraystretch}{1.12}
\begin{longtable}{r L{4.6cm}C{0.9cm}C{0.9cm}C{0.9cm}C{0.9cm}C{0.9cm}C{0.9cm}C{0.9cm}C{0.9cm}}
\caption{Per-dataset accuracy on the 85 \ucr{} datasets that \millet{} publishes. \emph{MILLET pub.} is the published mean of 5 seeds of \millet{}; the next four columns are \seanet{} encoders with the Top-$k$ conjunctive head (the model of the paper is \emph{bottleneck}); \emph{MILLET (ours)}, \emph{ResNet} and \emph{FCN} are the baselines re-trained under our recipe, each with conjunctive pooling. The best value of each row is in bold (the published column included).}
\label{tab:per_dataset_acc} \\
\toprule
\textbf{\#} & \textbf{Dataset} & \rotatebox{90}{\textbf{MILLET pub.}} & \rotatebox{90}{\textbf{bottleneck}} & \rotatebox{90}{\textbf{gated (mean)}} & \rotatebox{90}{\textbf{input-gated}} & \rotatebox{90}{\textbf{spike/trend}} & \rotatebox{90}{\textbf{MILLET (ours)}} & \rotatebox{90}{\textbf{ResNet}} & \rotatebox{90}{\textbf{FCN}} \\
\midrule
\endfirsthead
\multicolumn{10}{l}{\emph{\tablename~\thetable{} (continued)}} \\
\toprule
\textbf{\#} & \textbf{Dataset} & \rotatebox{90}{\textbf{MILLET pub.}} & \rotatebox{90}{\textbf{bottleneck}} & \rotatebox{90}{\textbf{gated (mean)}} & \rotatebox{90}{\textbf{input-gated}} & \rotatebox{90}{\textbf{spike/trend}} & \rotatebox{90}{\textbf{MILLET (ours)}} & \rotatebox{90}{\textbf{ResNet}} & \rotatebox{90}{\textbf{FCN}} \\
\midrule
\endhead
\midrule
\multicolumn{10}{r}{\emph{continued on the next page}} \\
\endfoot
\bottomrule
\endlastfoot
1 & Adiac & \textbf{0.832} & 0.733 & 0.788 & 0.749 & 0.772 & 0.812 & 0.785 & 0.788 \\
2 & ArrowHead & 0.838 & 0.851 & 0.829 & 0.840 & 0.869 & 0.827 & \textbf{0.886} & 0.840 \\
3 & Beef & 0.847 & 0.822 & \textbf{0.867} & 0.833 & 0.767 & 0.800 & 0.800 & 0.733 \\
4 & BeetleFly & \textbf{0.920} & 0.867 & 0.750 & 0.850 & 0.900 & 0.883 & 0.850 & 0.700 \\
5 & BirdChicken & 0.940 & 0.850 & 0.800 & 0.950 & 0.850 & 0.983 & \textbf{1.000} & 0.900 \\
6 & Car & 0.920 & 0.867 & 0.883 & 0.917 & 0.883 & 0.911 & \textbf{0.933} & 0.883 \\
7 & CBF & 1.000 & 0.993 & 0.947 & 0.999 & \textbf{1.000} & 0.997 & 0.989 & 0.997 \\
8 & ChlorineConcentration & \textbf{0.858} & 0.681 & 0.700 & 0.598 & 0.616 & 0.803 & 0.590 & 0.780 \\
9 & CinCECGTorso & 0.824 & 0.732 & 0.749 & 0.704 & 0.768 & 0.844 & 0.860 & \textbf{0.897} \\
10 & Coffee & \textbf{1.000} & \textbf{1.000} & \textbf{1.000} & \textbf{1.000} & \textbf{1.000} & \textbf{1.000} & \textbf{1.000} & \textbf{1.000} \\
11 & Computers & 0.769 & 0.721 & 0.640 & 0.688 & 0.596 & 0.741 & 0.732 & \textbf{0.828} \\
12 & CricketX & 0.827 & 0.811 & 0.833 & 0.818 & \textbf{0.836} & 0.808 & 0.718 & 0.710 \\
13 & CricketY & 0.823 & 0.805 & 0.772 & \textbf{0.836} & 0.813 & 0.818 & 0.679 & 0.715 \\
14 & CricketZ & \textbf{0.857} & 0.816 & 0.800 & 0.836 & 0.823 & 0.817 & 0.774 & 0.764 \\
15 & DiatomSizeReduction & \textbf{0.972} & 0.966 & 0.935 & 0.954 & 0.971 & 0.948 & 0.951 & 0.922 \\
16 & DistalPhalanxOutlineAgeGroup & 0.715 & 0.755 & 0.770 & 0.770 & 0.770 & 0.734 & \textbf{0.784} & 0.734 \\
17 & DistalPhalanxOutlineCorrect & 0.770 & 0.775 & 0.768 & 0.750 & 0.772 & \textbf{0.777} & 0.768 & 0.732 \\
18 & DistalPhalanxTW & 0.660 & 0.703 & 0.698 & 0.698 & \textbf{0.712} & 0.700 & 0.698 & 0.705 \\
19 & Earthquakes & 0.725 & 0.748 & 0.748 & 0.748 & \textbf{0.777} & 0.755 & 0.741 & 0.748 \\
20 & ECG200 & 0.896 & 0.887 & 0.890 & \textbf{0.900} & 0.850 & 0.830 & 0.830 & 0.890 \\
21 & ECG5000 & 0.942 & 0.944 & 0.933 & \textbf{0.945} & 0.942 & 0.940 & 0.941 & 0.942 \\
22 & ECGFiveDays & \textbf{1.000} & \textbf{1.000} & \textbf{1.000} & \textbf{1.000} & \textbf{1.000} & \textbf{1.000} & 0.957 & 0.978 \\
23 & ElectricDevices & 0.712 & 0.719 & 0.696 & 0.740 & 0.712 & 0.708 & \textbf{0.750} & 0.729 \\
24 & FaceAll & \textbf{0.953} & 0.805 & 0.822 & 0.807 & 0.834 & 0.925 & 0.922 & 0.919 \\
25 & FaceFour & 0.959 & 0.962 & \textbf{0.977} & 0.920 & 0.955 & 0.958 & 0.920 & 0.932 \\
26 & FacesUCR & 0.960 & 0.946 & 0.960 & \textbf{0.961} & \textbf{0.961} & 0.955 & 0.949 & 0.910 \\
27 & FiftyWords & \textbf{0.835} & 0.784 & 0.809 & 0.811 & 0.789 & 0.807 & 0.692 & 0.637 \\
28 & Fish & \textbf{0.981} & 0.952 & 0.954 & 0.971 & 0.954 & 0.977 & 0.960 & 0.971 \\
29 & FordA & 0.954 & 0.962 & 0.967 & 0.963 & 0.961 & 0.957 & \textbf{0.968} & 0.921 \\
30 & FordB & 0.850 & 0.845 & 0.837 & 0.826 & \textbf{0.858} & 0.840 & 0.836 & 0.785 \\
31 & GunPoint & 0.985 & 0.987 & 0.987 & 0.993 & 0.993 & 0.987 & \textbf{1.000} & \textbf{1.000} \\
32 & Ham & 0.709 & 0.721 & 0.733 & 0.657 & 0.686 & 0.711 & 0.686 & \textbf{0.752} \\
33 & HandOutlines & 0.948 & 0.943 & \textbf{0.951} & 0.938 & 0.897 & 0.939 & 0.916 & 0.889 \\
34 & Haptics & \textbf{0.524} & 0.464 & 0.393 & 0.448 & 0.432 & 0.497 & 0.442 & 0.471 \\
35 & Herring & 0.578 & 0.641 & 0.641 & 0.656 & 0.594 & 0.583 & \textbf{0.672} & \textbf{0.672} \\
36 & InlineSkate & \textbf{0.491} & 0.276 & 0.293 & 0.304 & 0.309 & 0.418 & 0.364 & 0.293 \\
37 & InsectWingbeatSound & 0.615 & \textbf{0.625} & 0.610 & 0.607 & 0.616 & 0.597 & 0.444 & 0.337 \\
38 & ItalyPowerDemand & 0.951 & 0.965 & \textbf{0.969} & 0.966 & 0.967 & -- & 0.932 & 0.962 \\
39 & LargeKitchenAppliances & 0.901 & 0.898 & 0.904 & \textbf{0.920} & 0.901 & 0.904 & 0.912 & 0.893 \\
40 & Lightning2 & 0.856 & 0.803 & \textbf{0.885} & 0.770 & 0.853 & 0.803 & 0.738 & 0.672 \\
41 & Lightning7 & 0.827 & 0.772 & 0.822 & \textbf{0.836} & 0.781 & 0.804 & 0.822 & 0.794 \\
42 & Mallat & 0.969 & 0.765 & 0.683 & 0.768 & 0.963 & 0.960 & 0.934 & \textbf{0.969} \\
43 & Meat & \textbf{0.987} & 0.856 & 0.933 & 0.817 & 0.883 & 0.983 & 0.817 & 0.800 \\
44 & MedicalImages & \textbf{0.802} & 0.741 & 0.712 & 0.730 & 0.720 & 0.754 & 0.762 & 0.746 \\
45 & MiddlePhalanxOutlineAgeGroup & 0.522 & 0.574 & 0.610 & 0.591 & 0.623 & 0.606 & \textbf{0.656} & 0.649 \\
46 & MiddlePhalanxOutlineCorrect & 0.810 & 0.817 & 0.790 & \textbf{0.828} & \textbf{0.828} & 0.801 & 0.804 & 0.808 \\
47 & MiddlePhalanxTW & 0.510 & 0.574 & 0.558 & 0.506 & 0.571 & 0.589 & 0.571 & \textbf{0.591} \\
48 & MoteStrain & 0.902 & 0.864 & 0.881 & 0.865 & 0.891 & 0.904 & \textbf{0.906} & 0.893 \\
49 & NonInvasiveFetalECGThorax1 & \textbf{0.951} & 0.869 & 0.908 & 0.882 & 0.922 & 0.943 & 0.939 & 0.945 \\
50 & NonInvasiveFetalECGThorax2 & \textbf{0.953} & 0.840 & 0.930 & 0.903 & 0.928 & 0.938 & 0.927 & 0.951 \\
51 & OliveOil & \textbf{0.847} & 0.400 & 0.400 & 0.400 & 0.400 & 0.556 & 0.400 & 0.800 \\
52 & OSULeaf & 0.947 & 0.843 & 0.872 & 0.876 & 0.835 & 0.950 & \textbf{0.975} & 0.955 \\
53 & PhalangesOutlinesCorrect & \textbf{0.830} & 0.809 & 0.814 & 0.823 & 0.822 & 0.823 & 0.803 & 0.780 \\
54 & Phoneme & \textbf{0.301} & 0.222 & 0.208 & 0.196 & 0.207 & 0.268 & 0.240 & 0.260 \\
55 & Plane & \textbf{1.000} & \textbf{1.000} & \textbf{1.000} & \textbf{1.000} & \textbf{1.000} & \textbf{1.000} & \textbf{1.000} & \textbf{1.000} \\
56 & ProximalPhalanxOutlineAgeGroup & 0.854 & 0.842 & 0.829 & 0.820 & 0.859 & 0.839 & \textbf{0.868} & 0.844 \\
57 & ProximalPhalanxOutlineCorrect & \textbf{0.923} & 0.884 & 0.900 & 0.880 & 0.873 & 0.901 & 0.883 & 0.869 \\
58 & ProximalPhalanxTW & 0.760 & 0.780 & \textbf{0.800} & 0.785 & 0.795 & 0.768 & 0.795 & 0.795 \\
59 & RefrigerationDevices & 0.510 & 0.506 & 0.469 & \textbf{0.517} & \textbf{0.517} & 0.506 & 0.491 & 0.501 \\
60 & ScreenType & 0.576 & 0.484 & 0.491 & 0.539 & 0.424 & 0.541 & 0.560 & \textbf{0.637} \\
61 & ShapeletSim & 0.909 & \textbf{1.000} & \textbf{1.000} & \textbf{1.000} & 0.483 & 0.998 & \textbf{1.000} & \textbf{1.000} \\
62 & ShapesAll & \textbf{0.928} & 0.854 & 0.855 & 0.857 & 0.875 & 0.907 & 0.855 & 0.870 \\
63 & SmallKitchenAppliances & 0.782 & \textbf{0.808} & 0.803 & 0.800 & 0.773 & 0.772 & 0.797 & 0.800 \\
64 & SonyAIBORobotSurface1 & 0.934 & 0.836 & 0.787 & 0.887 & 0.809 & 0.946 & 0.977 & \textbf{0.987} \\
65 & SonyAIBORobotSurface2 & 0.827 & 0.871 & 0.854 & 0.864 & 0.877 & 0.816 & \textbf{0.959} & 0.932 \\
66 & StarLightCurves & \textbf{0.977} & 0.970 & 0.970 & 0.974 & 0.972 & 0.975 & 0.975 & 0.969 \\
67 & Strawberry & \textbf{0.975} & 0.964 & 0.960 & 0.954 & 0.946 & 0.960 & 0.941 & 0.965 \\
68 & SwedishLeaf & 0.967 & 0.949 & 0.933 & 0.947 & 0.941 & 0.967 & \textbf{0.973} & 0.971 \\
69 & Symbols & 0.958 & 0.948 & 0.944 & 0.957 & \textbf{0.959} & 0.926 & 0.922 & 0.931 \\
70 & SyntheticControl & 0.997 & 0.996 & 0.993 & 0.997 & 0.993 & 0.999 & \textbf{1.000} & 0.983 \\
71 & ToeSegmentation1 & \textbf{0.963} & 0.936 & 0.934 & 0.952 & 0.908 & 0.958 & 0.952 & 0.952 \\
72 & ToeSegmentation2 & 0.909 & 0.944 & 0.954 & 0.923 & 0.946 & 0.879 & \textbf{0.962} & 0.939 \\
73 & Trace & \textbf{1.000} & \textbf{1.000} & \textbf{1.000} & \textbf{1.000} & \textbf{1.000} & \textbf{1.000} & \textbf{1.000} & \textbf{1.000} \\
74 & TwoLeadECG & 0.999 & 0.995 & 0.972 & 0.990 & 0.972 & 0.999 & 0.998 & \textbf{1.000} \\
75 & TwoPatterns & \textbf{1.000} & \textbf{1.000} & \textbf{1.000} & \textbf{1.000} & \textbf{1.000} & \textbf{1.000} & 1.000 & 0.995 \\
76 & UWaveGestureLibraryAll & 0.945 & 0.954 & \textbf{0.965} & 0.959 & 0.958 & 0.945 & 0.895 & 0.913 \\
77 & UWaveGestureLibraryX & \textbf{0.810} & 0.804 & 0.805 & 0.789 & 0.806 & 0.788 & 0.786 & 0.792 \\
78 & UWaveGestureLibraryY & \textbf{0.753} & 0.728 & 0.731 & 0.719 & 0.751 & 0.728 & 0.723 & 0.702 \\
79 & UWaveGestureLibraryZ & 0.743 & 0.733 & 0.711 & \textbf{0.763} & 0.738 & 0.732 & 0.753 & 0.709 \\
80 & Wafer & 0.998 & 0.997 & 0.998 & \textbf{0.999} & 0.996 & 0.997 & 0.998 & 0.997 \\
81 & Wine & \textbf{0.819} & 0.500 & 0.500 & 0.500 & 0.593 & 0.500 & 0.500 & 0.500 \\
82 & WordSynonyms & 0.742 & 0.706 & \textbf{0.745} & 0.718 & 0.730 & 0.725 & 0.538 & 0.578 \\
83 & Worms & \textbf{0.766} & 0.697 & 0.610 & 0.701 & 0.714 & 0.697 & 0.740 & 0.727 \\
84 & WormsTwoClass & 0.706 & 0.697 & 0.701 & \textbf{0.766} & 0.740 & 0.714 & 0.753 & 0.714 \\
85 & Yoga & \textbf{0.903} & 0.872 & 0.884 & 0.893 & 0.873 & 0.848 & 0.720 & 0.749 \\
\midrule
 & \textbf{Mean} & 0.845 & 0.810 & 0.808 & 0.813 & 0.809 & 0.827 & 0.815 & 0.814 \\
\end{longtable}
\endgroup

\begingroup
\seatab\setlength{\tabcolsep}{2pt}\renewcommand{\arraystretch}{1.12}
\begin{longtable}{r L{4.6cm}C{0.9cm}C{0.9cm}C{0.9cm}C{0.9cm}C{0.9cm}C{0.9cm}C{0.9cm}C{0.9cm}}
\caption{Per-dataset AOPCR on the 85 \ucr{} datasets that \millet{} publishes. \emph{MILLET pub.} is the published mean of 5 seeds of \millet{}; the next four columns are \seanet{} encoders with the Top-$k$ conjunctive head (the model of the paper is \emph{bottleneck}); \emph{MILLET (ours)}, \emph{ResNet} and \emph{FCN} are the baselines re-trained under our recipe, each with conjunctive pooling. The best value of each row is in bold among the columns trained with our recipe; the published column ($^\dagger$) was trained with \millet{}'s recipe and its AOPCR is on another scale, so it is shown but not compared.}
\label{tab:per_dataset_aopcr} \\
\toprule
\textbf{\#} & \textbf{Dataset} & \rotatebox{90}{\textbf{MILLET pub.$^\dagger$}} & \rotatebox{90}{\textbf{bottleneck}} & \rotatebox{90}{\textbf{gated (mean)}} & \rotatebox{90}{\textbf{input-gated}} & \rotatebox{90}{\textbf{spike/trend}} & \rotatebox{90}{\textbf{MILLET (ours)}} & \rotatebox{90}{\textbf{ResNet}} & \rotatebox{90}{\textbf{FCN}} \\
\midrule
\endfirsthead
\multicolumn{10}{l}{\emph{\tablename~\thetable{} (continued)}} \\
\toprule
\textbf{\#} & \textbf{Dataset} & \rotatebox{90}{\textbf{MILLET pub.$^\dagger$}} & \rotatebox{90}{\textbf{bottleneck}} & \rotatebox{90}{\textbf{gated (mean)}} & \rotatebox{90}{\textbf{input-gated}} & \rotatebox{90}{\textbf{spike/trend}} & \rotatebox{90}{\textbf{MILLET (ours)}} & \rotatebox{90}{\textbf{ResNet}} & \rotatebox{90}{\textbf{FCN}} \\
\midrule
\endhead
\midrule
\multicolumn{10}{r}{\emph{continued on the next page}} \\
\endfoot
\bottomrule
\endlastfoot
1 & Adiac & 15.89 & -0.83 & -0.02 & -0.27 & -0.64 & 1.60 & -2.05 & \textbf{1.73} \\
2 & ArrowHead & 3.20 & 0.24 & 0.45 & \textbf{1.37} & 0.67 & 0.60 & 0.84 & 1.31 \\
3 & Beef & 5.49 & 1.33 & 1.95 & 0.80 & -0.30 & 1.34 & \textbf{2.07} & 1.63 \\
4 & BeetleFly & 1.24 & 0.58 & \textbf{2.35} & 1.18 & 1.59 & 0.24 & 2.29 & 1.35 \\
5 & BirdChicken & 3.37 & 1.96 & \textbf{2.77} & 1.30 & 1.58 & 0.96 & 1.15 & 1.36 \\
6 & Car & 6.62 & 0.40 & 1.81 & -0.03 & 2.80 & 1.02 & 2.55 & \textbf{3.66} \\
7 & CBF & 3.58 & \textbf{1.97} & 1.58 & 0.92 & 1.71 & 0.87 & 0.97 & 0.85 \\
8 & ChlorineConcentration & 2.14 & 0.49 & 0.31 & 0.80 & \textbf{1.79} & 0.26 & 0.20 & 1.13 \\
9 & CinCECGTorso & 4.77 & 2.31 & 2.43 & \textbf{2.50} & 0.01 & 1.48 & 1.80 & 1.97 \\
10 & Coffee & 3.49 & 0.60 & 0.22 & 1.19 & \textbf{1.69} & 0.89 & 1.05 & 0.64 \\
11 & Computers & 3.44 & \textbf{0.89} & 0.65 & 0.53 & 0.45 & 0.53 & 0.74 & 0.47 \\
12 & CricketX & 4.51 & \textbf{2.03} & 1.40 & 1.36 & 1.34 & 1.27 & 1.78 & 1.87 \\
13 & CricketY & 4.89 & 1.75 & 1.80 & 1.45 & 1.21 & 1.22 & \textbf{1.95} & 1.94 \\
14 & CricketZ & 4.77 & 1.83 & 1.14 & 1.28 & 1.71 & 1.36 & \textbf{2.13} & 1.93 \\
15 & DiatomSizeReduction & 1.03 & 2.41 & -0.02 & 0.92 & \textbf{2.44} & -0.15 & 0.67 & 0.99 \\
16 & DistalPhalanxOutlineAgeGroup & 2.43 & 0.44 & 0.17 & 0.48 & 0.02 & 0.66 & \textbf{0.88} & 0.81 \\
17 & DistalPhalanxOutlineCorrect & 0.64 & 0.52 & 0.54 & 0.27 & -0.29 & 0.30 & \textbf{0.86} & 0.73 \\
18 & DistalPhalanxTW & 3.12 & 1.05 & 0.92 & \textbf{1.61} & -0.71 & 0.71 & 0.71 & 0.85 \\
19 & Earthquakes & 3.49 & 0.20 & 0.11 & 0.15 & 0.48 & \textbf{0.73} & 0.35 & 0.13 \\
20 & ECG200 & 2.86 & 0.38 & 0.74 & 0.10 & \textbf{1.02} & 0.44 & 0.49 & 0.45 \\
21 & ECG5000 & 6.70 & 1.01 & 1.32 & 0.49 & 1.21 & 1.22 & 1.26 & \textbf{1.46} \\
22 & ECGFiveDays & 4.76 & 0.68 & -0.51 & -0.01 & 0.31 & 0.19 & 0.27 & \textbf{1.31} \\
23 & ElectricDevices & 4.64 & 0.87 & 1.06 & 0.75 & 1.19 & 0.72 & 1.22 & \textbf{1.39} \\
24 & FaceAll & 4.00 & 0.78 & 0.94 & 0.25 & 0.41 & 0.20 & 1.23 & \textbf{1.41} \\
25 & FaceFour & 6.47 & 1.23 & 0.62 & 1.13 & \textbf{1.59} & 0.47 & 1.21 & 0.97 \\
26 & FacesUCR & 3.18 & 1.07 & \textbf{1.20} & 1.06 & 0.90 & 0.03 & 0.95 & 0.93 \\
27 & FiftyWords & 3.31 & 2.19 & 1.78 & 1.73 & 1.77 & 0.84 & \textbf{2.39} & 2.38 \\
28 & Fish & 5.48 & 0.71 & 1.09 & 1.00 & 1.84 & 1.51 & 2.33 & \textbf{3.45} \\
29 & FordA & 2.71 & \textbf{0.87} & 0.73 & -0.11 & 0.70 & 0.28 & -0.31 & 0.75 \\
30 & FordB & 1.64 & 0.51 & 0.03 & 0.51 & \textbf{0.93} & 0.37 & 0.18 & 0.25 \\
31 & GunPoint & 4.01 & \textbf{1.05} & 0.49 & 0.70 & 0.70 & 0.91 & 0.48 & 0.74 \\
32 & Ham & 3.93 & 1.11 & 1.51 & 1.10 & \textbf{1.82} & 0.75 & 1.59 & 1.62 \\
33 & HandOutlines & 12.71 & 1.48 & 2.90 & 2.02 & \textbf{5.52} & 1.85 & 1.34 & 3.51 \\
34 & Haptics & 5.99 & 1.63 & 1.37 & 2.13 & 1.44 & 1.25 & 2.01 & \textbf{2.47} \\
35 & Herring & 4.61 & 1.44 & 1.80 & 0.95 & \textbf{5.24} & 2.70 & 3.18 & 2.34 \\
36 & InlineSkate & 8.16 & 1.52 & 1.88 & \textbf{2.70} & 0.80 & 2.05 & 1.98 & 1.75 \\
37 & InsectWingbeatSound & 2.73 & 1.18 & \textbf{1.43} & 0.69 & 1.02 & 0.61 & 0.82 & 0.97 \\
38 & ItalyPowerDemand & 4.16 & 0.43 & \textbf{0.92} & 0.42 & 0.30 & -- & 0.54 & 0.49 \\
39 & LargeKitchenAppliances & 9.51 & \textbf{2.20} & 1.22 & 1.70 & 1.63 & 1.68 & 1.80 & 1.57 \\
40 & Lightning2 & 5.38 & 0.98 & 1.09 & 1.62 & 1.00 & 0.96 & \textbf{1.67} & 1.45 \\
41 & Lightning7 & 4.25 & \textbf{1.86} & 1.16 & 1.38 & 1.16 & 1.30 & 1.48 & 1.36 \\
42 & Mallat & 6.74 & 1.87 & 2.28 & 2.56 & 1.54 & 1.76 & \textbf{2.99} & 2.52 \\
43 & Meat & 6.19 & 1.62 & 1.59 & 0.68 & 0.95 & \textbf{2.08} & 1.74 & 2.01 \\
44 & MedicalImages & 2.89 & 0.91 & 0.96 & 0.52 & 0.69 & 0.92 & 0.99 & \textbf{1.07} \\
45 & MiddlePhalanxOutlineAgeGroup & 0.99 & 0.35 & 0.35 & 0.25 & 0.41 & 0.86 & \textbf{1.25} & 0.78 \\
46 & MiddlePhalanxOutlineCorrect & 2.45 & 0.72 & 0.58 & -0.18 & -0.36 & 0.13 & 2.69 & \textbf{2.84} \\
47 & MiddlePhalanxTW & 3.40 & 0.66 & 0.62 & 0.66 & 0.05 & 0.57 & \textbf{1.04} & 0.47 \\
48 & MoteStrain & 3.56 & 0.52 & 1.39 & 0.75 & \textbf{2.12} & 0.55 & 0.46 & 0.58 \\
49 & NonInvasiveFetalECGThorax1 & -2.26 & 1.73 & 1.46 & 1.29 & 1.23 & 0.30 & \textbf{8.51} & 5.18 \\
50 & NonInvasiveFetalECGThorax2 & 2.99 & 1.73 & 1.41 & 1.23 & 1.00 & 0.50 & \textbf{3.77} & 2.93 \\
51 & OliveOil & -3.63 & 0.12 & 0.08 & 0.12 & 0.37 & \textbf{3.79} & 0.37 & -1.05 \\
52 & OSULeaf & 6.53 & 1.26 & 0.73 & 1.16 & 0.58 & 0.84 & \textbf{2.56} & 1.90 \\
53 & PhalangesOutlinesCorrect & 2.04 & 0.04 & 0.45 & 0.36 & 0.31 & 0.33 & -0.32 & \textbf{0.48} \\
54 & Phoneme & 4.78 & \textbf{1.71} & 1.16 & 1.16 & 1.26 & 0.50 & 0.64 & 0.53 \\
55 & Plane & 3.39 & \textbf{1.13} & 1.05 & 1.07 & 1.07 & 0.31 & 0.26 & 0.46 \\
56 & ProximalPhalanxOutlineAgeGroup & 5.65 & 0.07 & -0.25 & 0.56 & \textbf{1.05} & 1.04 & 0.92 & 0.11 \\
57 & ProximalPhalanxOutlineCorrect & -3.54 & 0.08 & 0.31 & -1.67 & -0.12 & \textbf{0.69} & -1.35 & 0.68 \\
58 & ProximalPhalanxTW & 5.70 & -0.10 & -0.88 & -1.46 & \textbf{0.57} & -2.17 & -0.64 & -0.86 \\
59 & RefrigerationDevices & 4.25 & 0.32 & 0.28 & \textbf{1.41} & 0.28 & 0.63 & 0.86 & 0.71 \\
60 & ScreenType & 4.22 & 0.83 & 0.37 & 0.53 & 0.56 & 0.69 & 1.07 & \textbf{1.15} \\
61 & ShapeletSim & 3.43 & 1.92 & \textbf{2.46} & 1.15 & 0.23 & 0.74 & 1.18 & 0.85 \\
62 & ShapesAll & 9.56 & 2.36 & 2.23 & 2.28 & 1.44 & 1.64 & \textbf{3.82} & 3.15 \\
63 & SmallKitchenAppliances & 3.93 & 0.90 & \textbf{1.09} & 0.50 & 0.32 & 0.50 & 0.90 & 1.04 \\
64 & SonyAIBORobotSurface1 & 3.26 & 0.58 & -0.48 & 0.07 & 0.22 & 0.31 & \textbf{0.63} & 0.50 \\
65 & SonyAIBORobotSurface2 & 2.96 & 0.10 & -0.13 & 0.40 & \textbf{0.45} & 0.17 & 0.35 & 0.15 \\
66 & StarLightCurves & 8.42 & 2.65 & 1.69 & 1.89 & \textbf{3.22} & 1.01 & 2.83 & 1.86 \\
67 & Strawberry & 6.22 & 0.61 & 0.16 & 0.78 & 0.75 & 0.51 & 1.25 & \textbf{2.58} \\
68 & SwedishLeaf & 4.16 & \textbf{1.32} & 1.17 & 0.98 & 0.84 & 0.47 & 0.89 & 0.94 \\
69 & Symbols & 2.56 & 1.29 & 1.17 & 1.87 & \textbf{2.04} & 1.20 & 1.31 & 1.71 \\
70 & SyntheticControl & 3.30 & \textbf{0.95} & 0.40 & 0.57 & 0.74 & 0.66 & 0.75 & 0.54 \\
71 & ToeSegmentation1 & 2.18 & 0.53 & 0.24 & 0.65 & 0.63 & 0.33 & 0.64 & \textbf{0.86} \\
72 & ToeSegmentation2 & 3.21 & 0.91 & 0.42 & 0.83 & 0.31 & 0.41 & \textbf{1.14} & 0.62 \\
73 & Trace & 4.92 & 2.05 & 1.40 & 1.82 & \textbf{2.46} & 1.76 & 2.24 & 2.21 \\
74 & TwoLeadECG & 4.76 & \textbf{1.49} & 0.94 & 1.02 & 0.12 & -0.22 & 0.83 & 0.29 \\
75 & TwoPatterns & 12.16 & 2.42 & \textbf{3.07} & 2.22 & 1.89 & 1.35 & 2.88 & 1.29 \\
76 & UWaveGestureLibraryAll & 8.08 & 1.95 & 2.03 & 2.43 & 1.46 & 1.27 & 2.45 & \textbf{2.96} \\
77 & UWaveGestureLibraryX & 4.30 & 1.21 & 1.70 & 1.09 & 0.81 & 0.85 & 1.78 & \textbf{1.99} \\
78 & UWaveGestureLibraryY & 4.10 & 1.23 & 1.55 & 1.12 & 0.93 & 0.87 & \textbf{1.66} & 1.61 \\
79 & UWaveGestureLibraryZ & 4.65 & 1.44 & 1.62 & 1.13 & 1.22 & 0.92 & \textbf{1.70} & 1.57 \\
80 & Wafer & 1.14 & 0.37 & 0.50 & \textbf{0.84} & 0.27 & -0.27 & 0.75 & 0.77 \\
81 & Wine & 26.96 & 0.61 & 2.27 & 1.12 & \textbf{3.31} & 0.03 & 0.05 & 0.03 \\
82 & WordSynonyms & 3.04 & 1.49 & 1.28 & 1.33 & 0.94 & 0.69 & 2.09 & \textbf{2.20} \\
83 & Worms & 3.11 & 1.22 & 1.46 & 1.89 & 1.59 & 0.89 & 1.63 & \textbf{2.15} \\
84 & WormsTwoClass & 4.30 & 0.95 & 1.19 & 1.01 & 1.13 & 0.90 & 1.29 & \textbf{1.74} \\
85 & Yoga & 6.62 & 1.28 & 1.72 & 1.80 & \textbf{3.22} & 0.64 & 1.14 & 2.62 \\
\midrule
 & \textbf{Mean} & 4.55 & 1.09 & 1.06 & 0.96 & 1.11 & 0.81 & 1.33 & 1.38 \\
\end{longtable}
\endgroup

% ---------------------------------------------------------------------
\subsection{Relation to the Published \millet{} Results}
\label{app:published_all}
\label{app:published}

The main comparisons of the paper use \millet{} retrained with our own budget, and not its published numbers. \Cref{tab:published} sets our retrained baseline next to the published numbers of \millet{} \citep{early2024millet}, with one row per way of obtaining the number; the \webtraffic{} columns are means over the runs and the \ucr{} columns are means over the $85$ published datasets. \citet{early2024millet} report $0.924$ over five seeds on \webtraffic{}, and our retrained baseline reaches $0.888$ with the same architecture. The two differ in the training recipe (\Cref{app:baselines}).

\begin{table}[!htbp]
\caption{The \millet{} baseline (InceptionTime $+$ conjunctive pooling): the published numbers of \citet{early2024millet} and our retrained baseline. The \webtraffic{} columns are means ($\pm$ std) over the runs in both results folders, and the \ucr{} columns are means over the 85 published datasets (the number in brackets if fewer were finished). Only the last row is used in the main comparisons. $^\dagger$: the published AOPCR comes from \millet{}'s own recipe and is on another scale, so AOPCR must only be compared within one recipe.}
\label{tab:published}
\centering
\seatab\setlength{\tabcolsep}{3pt}
\begin{tabular}{r L{4.3cm}C{0.75cm}C{1.3cm}C{1.55cm}C{1.4cm}C{1.35cm}C{1.2cm}}
\toprule
& & & \multicolumn{3}{c}{\textbf{WebTraffic}} & \multicolumn{2}{c}{\textbf{UCR-85 mean}} \\
\cmidrule(lr){4-6}\cmidrule(lr){7-8}
\textbf{\#} & \textbf{How the number was obtained} & \textbf{Runs} & \textbf{Acc.} & \textbf{AOPCR} & \textbf{NDCG@$n$} & \textbf{Acc.} & \textbf{AOPCR} \\
\midrule
1 & Published, mean of 5 seeds & 5 & 0.924 & 12.768$^\dagger$ & 0.674 & 0.8445 & 4.553$^\dagger$ \\
2 & Published, 5-model ensemble & 1 & 0.940 & 11.100$^\dagger$ & 0.694 & -- & -- \\
% hidden: MILLET trained with its published recipe in our code (not reported)
% 3 & Re-trained here, \emph{their} recipe (1500 epochs, no label smoothing, no validation split) & 2 & $0.918$ \newline $\pm 0.002$ & $12.060$ \newline $\pm 1.208$$^\dagger$ & $0.643$ \newline $\pm 0.019$ & 0.8434 (84) & 3.926$^\dagger$ \\
3 & \textbf{Re-trained here, our recipe (400 epochs, label smoothing, early stopping): the baseline of every comparison} & 4 & $0.888$ \newline $\pm 0.007$ & $2.457$ \newline $\pm 0.655$ & $0.681$ \newline $\pm 0.013$ & 0.8274 (84) & 0.812 \\
\bottomrule
\end{tabular}
\end{table}

The same table also shows the problem of \Cref{sec:res:aopcr}. The published \webtraffic{} AOPCR of \millet{} is $12.768$, against $2.457$ for our retrained baseline, although the two have a similar \ndcg{} ($0.674$ and $0.681$). A factor of about five on a metric that is supposed to measure the quality of an explanation therefore does not reflect a better ranking of the time points. This is why the AOPCR column of \Cref{tab:web_main} is only read across its own rows, and never against an AOPCR reported in another paper, including the published $4.55$ of \millet{}.

\end{document}